\pdfoutput=1
\documentclass[11pt]{article}

\usepackage[T1]{fontenc}
\usepackage{iftex}
\ifPDFTeX\usepackage[utf8]{inputenc}\fi
\usepackage[letterpaper,margin=1.15in,headheight=14pt]{geometry}
\usepackage{amsmath,amsthm,mathtools}
\usepackage[type1]{libertine}
\usepackage[libertine]{newtxmath}
\usepackage[type1,scale=0.95]{sourcesanspro}
\usepackage{bm}
\usepackage{pifont}
\usepackage{microtype}
\usepackage{booktabs,tabularx,array,multirow,longtable}
\usepackage{ragged2e}
\usepackage{xcolor}
\definecolor{accent}{HTML}{1F4E8C}
\definecolor{accentlight}{HTML}{EAF0F8}
\definecolor{tolblue}{HTML}{4477AA}
\definecolor{tolgreen}{HTML}{228833}
\definecolor{tolorange}{HTML}{EE7733}
\definecolor{tolgrey}{HTML}{8C8C8C}
\usepackage{caption}
\usepackage{graphicx}
\usepackage{tikz}
\usetikzlibrary{arrows.meta,calc,positioning,shapes.geometric,fit}
\usepackage[edges]{forest}
\usepackage{enumitem}
\usepackage{float}
\usepackage[section]{placeins}
\usepackage[round,sort]{natbib}
\defcitealias{anthropic2025november}{\textit{Circuits Updates}, 2025}
\defcitealias{anthropic2026may}{\textit{Circuits Updates}, 2026}

\usepackage[explicit]{titlesec}
\titleformat{\section}{\Large\sffamily\bfseries\color{accent}}{\thesection}{0.75em}{#1}
\titleformat{\subsection}{\large\sffamily\bfseries\color{accent}}{\thesubsection}{0.75em}{#1}
\titleformat{\subsubsection}{\normalsize\sffamily\bfseries\color{accent}}{\thesubsubsection}{0.75em}{#1}
\titleformat{\paragraph}[runin]{\normalsize\sffamily\bfseries}{}{0pt}{#1}
\titlespacing*{\paragraph}{0pt}{1.0ex plus .5ex}{0.8em}
\usepackage{tocloft}

\usepackage{fancyhdr}
\fancypagestyle{plain}{\fancyhf{}\fancyfoot[C]{\small\sffamily\thepage}}

\usepackage[colorlinks=true,linkcolor=accent,citecolor=tolgreen!75!black,urlcolor=tolblue,
  linktoc=page,pagebackref=true]{hyperref}
\renewcommand*{\backrefalt}[4]{\ifcase #1 \or{\footnotesize\color{black!55}Cited on page~#2.}%
  \else{\footnotesize\color{black!55}Cited on pages~#2.}\fi}
\usepackage[capitalise,nameinlink,noabbrev]{cleveref}

\setlist{nosep,leftmargin=*}
\newtheoremstyle{accentdef}{4pt}{4pt}{\normalfont}{}{\sffamily\bfseries\color{accent}}{.}{0.5em}{}
\newtheoremstyle{accentplain}{4pt}{4pt}{\itshape}{}{\sffamily\bfseries\color{tolgreen!70!black}}{.}{0.5em}{}
\theoremstyle{accentdef}
\newtheorem{definition}{Definition}[section]
\newtheorem{example}[definition]{Example}
\theoremstyle{accentplain}
\newtheorem{result}[definition]{Result}

\theoremstyle{remark}

\usepackage{tcolorbox}
\tcbuselibrary{breakable,skins}
\tcbset{callout/.style={enhanced,frame hidden,sharp corners,boxrule=0pt,
  borderline west={2.6pt}{0pt}{#1},colback=#1!7!white,left=9pt,right=8pt,top=4pt,bottom=4pt}}
\tcolorboxenvironment{definition}{callout=tolblue}
\tcolorboxenvironment{example}{callout=tolgrey,breakable}
\tcolorboxenvironment{result}{callout=tolgreen}
\tcbset{summarybox/.style={enhanced,colback=accentlight,colframe=accent,boxrule=0.5pt,
  arc=2pt,left=8pt,right=8pt,top=5pt,bottom=5pt,colbacktitle=accent,coltitle=white,
  fonttitle=\sffamily\bfseries,toptitle=2pt,bottomtitle=2pt}}

\newcommand{\R}{\mathbb{R}}
\newcommand{\E}{\mathbb{E}}
\newcommand{\Prob}{\mathbb{P}}

\newcommand{\cL}{\mathcal{L}}
\newcommand{\norm}[1]{\left\lVert #1\right\rVert}
\newcommand{\ip}[2]{\left\langle #1,#2\right\rangle}
\newcommand{\Sup}{\mathrm{S}}                 %
\newcommand{\Rlin}{R_{\mathrm{lin}}}          %

\newcommand{\doesnot}[1]{\par\smallskip\noindent{\sffamily\bfseries\small\color{black!60}Does not establish.}~#1\par\smallskip}

\newcommand{\status}[1]{}

\newcommand{\takeaway}[2]{%
  \par\medskip
  \begin{tcolorbox}[enhanced,colback=tolorange!7!white,colframe=tolorange!85!black,
    boxrule=0.5pt,arc=2pt,left=8pt,right=8pt,top=5pt,bottom=5pt,colbacktitle=tolorange!85!black,
    coltitle=white,fonttitle=\sffamily\bfseries,title={Takeaway Message of Section~\ref*{#1}}]
  #2
  \end{tcolorbox}
  \par\medskip}

\newcommand{\routestations}{2/Definitions \& measures/sec:definitions,3/Geometry \& capacity/sec:geometry,
  4/Learning \& access/sec:access,5/Computation/sec:computation,6/Implications/sec:relations,
  7/Recovery methods/sec:recovery,8/Evaluation/sec:protocol,9/Failure modes/sec:failures,
  10/Applications/sec:applications,11/Open questions/sec:open}
\newcommand{\routemap}{%
  \begin{tikzpicture}[x=0.0985\linewidth]
  \shade[left color=tolblue,right color=tolgreen] (4.62,-2pt) rectangle (5.38,2pt);
  \draw[tolblue,line width=4pt,line cap=round] (0.5,0) -- (4.62,0);
  \draw[tolgreen,line width=4pt] (5.38,0) -- (9.5,0);
  \draw[tolgreen,line width=4pt,-{Stealth[length=9pt,width=11pt]}] (9.5,0) -- (9.98,0);
  \foreach \n/\lab/\tgt [count=\i from 0] in \routestations{
    \ifnum\i<5 \def\c{tolblue}\else\def\c{tolgreen}\fi
    \node[circle,fill=white,draw=\c,line width=1.8pt,minimum size=17pt,inner sep=0pt,
      font=\fontsize{8}{8}\selectfont\sffamily\bfseries,text=\c!80!black] at (\i+0.5,0)
      {\hyperref[\tgt]{\textcolor{\c!80!black}{\n}}};
    \node[font=\fontsize{7.6}{9}\selectfont\sffamily,text=black!75,anchor=north,align=center,
      text width=0.095\linewidth] at (\i+0.5,-13pt) {\hyperref[\tgt]{\textcolor{black!75}{\lab}}};}
  \draw[tolblue!70,line width=0.6pt] (0.3,17pt) -- (0.3,21pt) -- (4.7,21pt) -- (4.7,17pt);
  \draw[tolgreen!70,line width=0.6pt] (5.3,17pt) -- (5.3,21pt) -- (9.7,21pt) -- (9.7,17pt);
  \node[fill=white,inner xsep=4pt,font=\scriptsize\sffamily\bfseries,text=tolblue!85!black] at (2.5,21pt) {THEORY\ \ \mdseries organized by the decoder class};
  \node[fill=white,inner xsep=4pt,font=\scriptsize\sffamily\bfseries,text=tolgreen!75!black] at (7.5,21pt) {PRACTICE\ \ \mdseries graded by the strength of claims};
  \end{tikzpicture}}
\newcommand{\roadmap}{%
  \begin{tikzpicture}[x=0.1\linewidth]
  \shade[left color=tolblue,right color=tolgreen] (4.62,-1pt) rectangle (5.38,1pt);
  \draw[tolblue!80,line width=2pt,line cap=round] (0.5,0) -- (4.62,0);
  \draw[tolgreen!80,line width=2pt,line cap=round] (5.38,0) -- (9.5,0);
  \foreach \n/\lab/\tgt [count=\i from 0] in \routestations{
    \ifnum\i<5 \def\c{tolblue}\else\def\c{tolgreen}\fi
    \ifnum\n=\value{section}
      \node[circle,fill=\c,draw=\c,minimum size=13pt,inner sep=0pt,
        font=\fontsize{7}{7}\selectfont\sffamily\bfseries,text=white] at (\i+0.5,0) {\n};
      \node[font=\fontsize{6.8}{8}\selectfont\sffamily\bfseries,text=\c!80!black,anchor=north] at (\i+0.5,-8pt) {\lab};
    \else
      \node[circle,fill=white,draw=\c!80,line width=1pt,minimum size=9pt,inner sep=0pt,
        font=\fontsize{5.6}{5.6}\selectfont\sffamily,text=\c!80!black] at (\i+0.5,0)
        {\hyperref[\tgt]{\textcolor{\c!80!black}{\n}}};
    \fi}
  \end{tikzpicture}\par\smallskip}

\newcommand{\kw}[1]{\tikz[baseline=(k.base)]\node[rounded corners=2.5pt,fill=accent!8,text=accent,
  inner xsep=5pt,inner ysep=2.5pt,font=\small\sffamily](k){#1};}

\newcommand{\aff}[1]{\raisebox{0.55ex}{\scriptsize #1}}

\newcommand{\treelab}{\hyphenpenalty=10000\relax\raggedright}

\begin{document}

\hypersetup{
  pdftitle={Feature Superposition in Neural Networks: From Theory to Practice},
  pdfauthor={Dai Shi, Xiaoyu Li, Andi Han, Jose Miguel Hernandez-Lobato}
}

\thispagestyle{plain}
\vspace*{-1.0cm}
{\centering\sffamily\bfseries\fontsize{25}{30}\selectfont\color{accent}Feature Superposition in Neural Networks\par}
\vspace{7pt}
{\centering\sffamily\bfseries\fontsize{20}{24}\selectfont\color{black!70}From
  \textcolor{tolblue!90!black}{Theory} to \textcolor{tolgreen!80!black}{Practice}\par}
\vspace{14pt}
{\centering\sffamily\bfseries Dai Shi\aff{1}\,{\color{accent}\ding{41}}\quad Xiaoyu Li\aff{2}\quad Andi Han\aff{3}\quad
  Jos\'e Miguel Hern\'andez-Lobato\aff{1}\par}
\vspace{5pt}
{\centering\small\color{black!60}\aff{1}University of Cambridge\quad \aff{2}University of New South Wales\quad
  \aff{3}University of Sydney\quad {\color{accent}\ding{41}}\,\href{mailto:ds2213@cam.ac.uk}{ds2213@cam.ac.uk}\par}
\vspace{22pt}
\noindent{\sffamily\bfseries\small\color{accent}ABSTRACT}\par\vspace{5pt}
\noindent Superposition refers to neural networks representing more features than they have dimensions.
It was proposed to explain polysemantic neurons, which respond to several features, and also
other phenomena, such as neural scaling laws and adversarial vulnerability. A feature stored in
superposition is useful to the extent that its value can be read back despite interference from
the other features. The theory of superposition therefore mainly asks how wide a layer must be for
the features to be read back, and by which decoder. In theoretical models, a single linear map that
reads back every combination of nonzero features needs a width nearly quadratic in their number,
while sparse optimization can succeed with a width linear in that number. In practice, one may hope
to interpret a trained network by recovering its features in the same way. However, its features
are usually unknown, and thus they are estimated from its activations or weights by different
recovery methods, such as sparse autoencoders and transcoders. In this work, we organize the theory
of superposition by the decoder class and its practice by the strength of the claims about a
feature supported by each method's results. On the theory side, we review the lower bounds on the
overlap between feature directions, the conditions under which trained toy models share dimensions
among features, and the separation between storing features and computing new features. On the
practice side, we categorize recovery methods into four families, from probes for a named feature
to decompositions of the trained weights. We then compare their evaluations, failure modes, and
applications, and grade the evidence on one scale of claims about a feature, from its presence in
the hidden representation to an explanation of the network's computation. Finally, we assess
previously stated open problems and identify remaining theoretical and empirical questions about
superposition in trained networks. We hope our work can pave the way for a deeper understanding of
superposition and more reliable methods for interpreting neural networks.

\par\vspace{12pt}
\noindent{\small\sffamily\color{black!55}Keywords}\enspace\kw{Superposition}\ \kw{Mechanistic Interpretability}\
  \kw{Sparse Autoencoders}\ \kw{Compressed Sensing}\par
\vspace{6pt}
\noindent{\small\sffamily\color{black!55}Project page}\enspace{\small\sffamily
  \href{https://eethanshi.github.io/awesome-superposition/}{eethanshi.github.io/awesome-superposition}}\par
\vspace{26pt}
\noindent\routemap\par
\vspace{6pt}
{\centering\footnotesize\sffamily\color{black!60}Structure of the survey: Sections 2 to 6 review the theory and Sections 7
  to 11 its practice.\par}
\vfill
{\color{black!20}\hrule}\vspace{10pt}
{\centering\includegraphics[height=0.6cm]{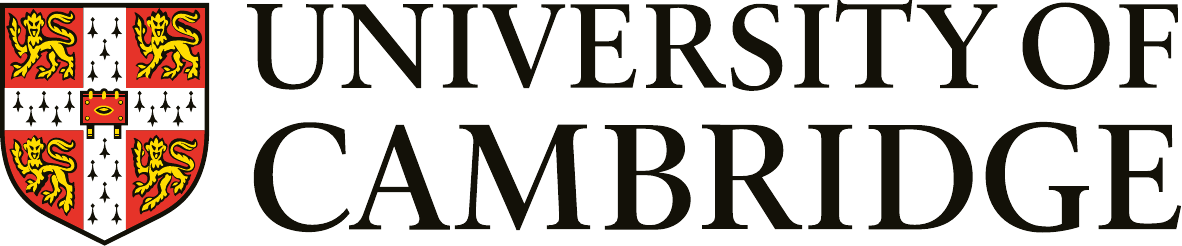}\hspace{0.55cm}%
  \includegraphics[height=0.85cm]{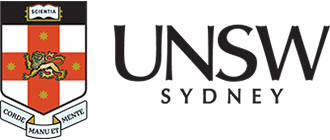}\hspace{0.55cm}%
  \includegraphics[height=0.75cm]{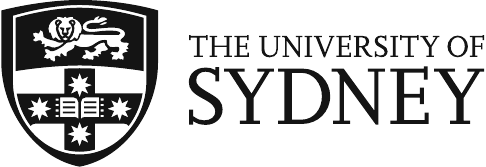}\par}
\clearpage
\tableofcontents
\clearpage

\section{Introduction}
\label{sec:intro}

Superposition refers to neural networks representing more features than they have dimensions
\citep{elhage2022toy}. It was proposed to explain \emph{polysemantic} neurons, which respond to
several features \citep{elhage2022toy,bricken2023monosemanticity}, and also other phenomena, such as
neural scaling laws \citep{liu2025scaling} and adversarial vulnerability
\citep{elhage2022toy,gorton2025adversarial,stevinson2026interference}.

In the toy models of \citet{elhage2022toy}, the hidden activation vector is a sum of feature
directions, each scaled by the value of its feature. Since a space holds at most as many mutually
orthogonal directions as its dimension, some feature directions overlap when features outnumber
dimensions. This overlap causes \emph{interference}, i.e., a readout along the direction of one feature
also picks up the overlapping features that are nonzero on the same input. When the features are
sparse, a nonlinear decoder suppresses the rare and small interference at the cost of shrinking the
nonzero values, and the trained toy models represent more features than their width. Beyond these toy
models, a feature stored in superposition is useful to the extent that its value can be read back despite
this interference. The theory of superposition therefore mainly asks how wide a layer must be for the
features to be read back, and by which decoder.

Sparse coding and compressed sensing study the same question for a sparse vector observed through
fewer linear measurements than its length \citep{olshausen1996,donoho2003optimally}. Already for two
features stored in one dimension, the answer depends on the class of decoders allowed
(\cref{ex:antipodal}). As the features become sparser, a ReLU decoder reads both back with a
vanishing fraction of unexplained variance, while every linear decoder leaves at least half of the
variance of each feature unexplained. More generally, a single linear map that reads back every
combination of nonzero features needs a width nearly quadratic in their number
\citep{garg2026howmany}. By contrast, sparse optimization with a suitable encoder can succeed with a
width linear in that number, up to a logarithmic factor \citep{candes2006stable}. Besides the decoder
class, these guarantees depend on the distribution of the feature values and on the permitted
reconstruction error. Two bounds that appear to conflict often hold for different settings of these
three choices.

In practice, one may hope to interpret a trained network by recovering its features in the same way
\citep{sharkey2025open,bereska2024mireview}. However, the features of a trained network are usually
unknown, and thus they are estimated from its activations or weights by different recovery methods,
such as sparse autoencoders \citep{bricken2023monosemanticity,cunningham2024sae} and transcoders
\citep{dunefsky2024transcoders}. The resulting estimates are usually evaluated indirectly, for example
by how well a description in words predicts when a feature is active, or by how an edit along its
direction changes the output. The two evaluations support claims of different strength about a
feature, since a feature described well in words need not be used by the network. Moreover, a high
score can arise with directions that barely move from their random initialization, as sparse
autoencoders constrained in this way score close to fully trained sparse autoencoders on several such
evaluations \citep{korznikov2026sanity}.

Motivated by these two difficulties, it is natural to \textbf{organize the theory of superposition by
the decoder class, and its practice by the strength of the claims about a feature supported by each
method's results}. Since polysemantic neurons also arise when orthogonal feature directions are
rotated away from the neuron axes, we adopt a definition stated on the feature directions rather than
on the neurons \citep{prieto2026correlations}. Under this definition, features are in superposition if their
directions overlap and a single decoder from a given class reads each of them back to a given
tolerance. We measure the degree of superposition by the number of features read back per dimension
for a fixed decoder class, distribution of the feature values, and permitted error. On the theory
side, we review the lower bounds on the overlap between feature directions, the conditions under
which trained toy models share dimensions among features, and the width needed by each decoder
class. As later layers also transform the stored values into new features, we then review the
separation between storing features and computing new features. For specified families of sparse
Boolean tasks, a fixed-depth network able to compute every task in a family needs more width than a
layer storing the outputs of those tasks \citep{adler2024complexity}. We also examine the evidence
that superposition explains the phenomena above, from polysemantic neurons to adversarial
vulnerability, and the assumptions behind the recovery of features from a trained network.

On the practice side, we categorize recovery methods into four families, from probes for a named
feature to decompositions of the trained weights. Supervised probes test whether a property named in
advance can be read from the activations, while sparse autoencoders and other sparse dictionaries
reconstruct the activations as sparse combinations of directions learned from unlabeled data.
Transcoders and other functional replacement methods extend this reconstruction from the activations
to the computation of a sublayer, and parameter decomposition methods split the trained weights
themselves into components. We grade the results of all four families on one scale of claims about a
feature, from its presence in the hidden representation to an explanation of the network's
computation. A successful probe supports a claim near the lower end of this scale, while a claim near
the upper end requires an intervention on the feature and a measurement of the resulting change in
behavior. We then compare the evaluations of these methods by the property of a recovered feature
tested by each evaluation, since a claim about a recovered feature needs a reference for the
claimed property. A sparse autoencoder can still score well on these
evaluations while splitting one feature of the network into several latents, and we therefore
collect the documented failure modes of these methods. Recovered features nevertheless support useful detection
and intervention in particular settings, and we review such applications, from monitoring a
concept to steering and editing a network. Finally, we assess previously stated open problems and identify
remaining theoretical and empirical questions about superposition in trained networks.

\paragraph{Related Surveys.}
Among broad reviews of neural-network interpretability, \citet{rauker2023transparent} classify
methods by the part of the network explained by each method, from weights and neurons to subnetworks
and latent representations. Within mechanistic interpretability, \citet{bereska2024mireview} present
superposition as one of the hypotheses about the representation of features and review methods for
causally dissecting model behavior. \citet{ranaldi2025mi} reviews the same field for generative
language models, from the neurons, attention heads, and circuits that store knowledge to the training
dynamics of memorization and generalization. \citet{sharkey2025open} group the open problems of the
field into its methods, their application to specific goals, and socio-technical questions, and the
problems of its methods include limitations of sparse dictionary learning. At the method level, the
dedicated survey by \citet{shu2025saesurvey} examines sparse autoencoders, covering their
architectures and training as well as feature explanations, evaluation, and applications. For
superposition itself, \citet{klindt2026unifying} combine identifiability theory, compressed-sensing
guarantees for sparse coding, and quantitative interpretability metrics into a three-step framework.
Complementing these works, our survey organizes the theory of superposition by the decoder class and
grades the results of four families of recovery methods on one scale of claims about a feature.
Sparse autoencoders form one of these families, and we retain an overview of their architectures in
\cref{sec:rec-dictionaries} and refer to \citet{shu2025saesurvey} for a detailed catalogue.

\paragraph{Paper Search and Selection Criteria.}
\emph{(1) Search terms.} The literature search used a \href{https://arxiv.org}{preprint
repository and its API}, a \href{https://aclanthology.org}{computational linguistics publication
archive}, and an \href{https://transformer-circuits.pub}{online collection of interpretability
research reports}, and we traced citations forward and backward from the retrieved works. The queries
combined superposition, polysemanticity, monosemanticity, sparse autoencoder, dictionary learning,
linear representation hypothesis, feature geometry, computation in superposition, and feature
capacity. Direct automated access to a \href{https://openreview.net}{review platform} was
unavailable during the search, so for works available there alone, publication status was obtained
from preprint metadata and conference listings.

\emph{(2) Selection criteria.} We include studies of representations with more features than
dimensions, together with methods that produce or evaluate estimates of a trained network's
features. Negative results concerning these representations or methods are included under the
same criteria. The literature comprises journal and conference papers, preprints, and
institutional research reports. Results from research reports are attributed in the text to their authors or report titles
(\cref{sec:prot-evidence}).

\emph{(3) Exclusions.} We exclude quantum superposition and quantum machine learning, the
Kolmogorov superposition theorem and Kolmogorov-Arnold networks, and superposition coding in
communications. We also exclude parameter, model, and task-vector superposition in model merging,
which combines trained models within one set of weights rather than features within one network.

\paragraph{Organization.}
\Cref{fig:taxonomy} shows the structure of the survey, together with representative works for each
section. \Cref{sec:definitions} defines superposition by overlapping feature directions and their recovery by
a decoder of a given class. It also introduces the packing degree, the superposition benefit, and the
six levels of claims about a feature. The theory part then reviews the lower bounds on the overlap
between feature directions and how sparsity and nonlinear decoding change the effect of this overlap
on recovery (\cref{sec:geometry}). \Cref{sec:access} states the conditions under which trained toy
models share dimensions, the width needed by each decoder class, and the assumptions needed to
identify the feature directions themselves. \Cref{sec:computation} reviews the separation between
storing features and computing new features, together with explicit constructions that compute with
shared neurons. \Cref{sec:relations} then examines the evidence for superposition as an
explanation of observed behavior and states the assumptions behind feature recovery. The practice
part organizes recovery methods into four families (\cref{sec:recovery}) and compares the
evaluations of their outputs and the computational cost of these methods (\cref{sec:protocol}).
\Cref{sec:failures} collects their documented
failures, and \cref{sec:applications} reviews the interventions and applications built on their
outputs in light of these failures. \Cref{sec:open} assesses previously stated open problems and
develops the remaining questions, and \cref{sec:conclusion} states the conclusions supported by the
theoretical and empirical evidence. Each technical section opens with a summary of its results and
closes with a takeaway of the points used in later sections.

\begin{figure}[t]
\centering
\resizebox{\linewidth}{!}{%
\begin{forest}
  forked edges,
  for tree={grow'=0, draw, rounded corners=2pt, font=\sffamily\small,
    anchor=west, child anchor=west, parent anchor=east, edge={draw=black!40, line width=0.6pt},
    l sep=5mm, s sep=1.6mm, fork sep=2.5mm, inner xsep=4pt, inner ysep=2.6pt, line width=0.5pt},
  where level=0{rotate=90, anchor=center, parent anchor=south, fill=accent, draw=accent,
    text=white, font=\sffamily\bfseries\normalsize, minimum width=7cm}{},
  where level=1{rotate=90, anchor=center, child anchor=north, parent anchor=south,
    font=\sffamily\bfseries\small, minimum width=2.6cm}{},
  where level=2{text width=4.4cm, font=\sffamily\bfseries\small\treelab}{},
  where level=3{text width=9.4cm, font=\sffamily\small\treelab, fill=white}{},
[{Feature Superposition in Neural Networks}
  [{Theory}, fill=tolblue!25, draw=tolblue
    [{Definitions and measures (\cref{sec:definitions})}, fill=tolblue!10, draw=tolblue
      [{Overlap and recovery \citep{prieto2026correlations}; packing degree; superposition benefit; six claim levels}, draw=tolblue!60]]
    [{Geometry and capacity (\cref{sec:geometry})}, fill=tolblue!10, draw=tolblue
      [{Total-overlap lower bound \citep{welch1974,benedetto2003frames}; sparsity and nonlinear decoding \citep{elhage2022toy}; counting bounds}, draw=tolblue!60]]
    [{Learning, access, and identifiability (\cref{sec:access})}, fill=tolblue!10, draw=tolblue
      [{Phase diagrams \citep{elhage2022toy}; fourth-moment threshold \citep{scherlis2022capacity}; incidental polysemanticity \citep{lecomte2024incidental}; linear versus sparse width \citep{garg2026howmany,candes2006stable}; identifiability}, draw=tolblue!60]]
    [{Computation (\cref{sec:computation})}, fill=tolblue!10, draw=tolblue
      [{Storing versus computing \citep{adler2024complexity}; explicit constructions \citep{hanni2024computation}; compressed computation}, draw=tolblue!60]]
    [{Implications (\cref{sec:relations})}, fill=tolblue!10, draw=tolblue
      [{Boundaries with polysemanticity; scaling laws \citep{liu2025scaling}; adversarial vulnerability \citep{gorton2025adversarial,stevinson2026interference}; assumptions behind recovery}, draw=tolblue!60]]]
  [{Practice}, fill=tolgreen!25, draw=tolgreen
    [{Recovery methods (\cref{sec:recovery})}, fill=tolgreen!10, draw=tolgreen
      [{Probes; sparse autoencoders \citep{bricken2023monosemanticity,cunningham2024sae}; transcoders \citep{dunefsky2024transcoders}; parameter decomposition \citep{braun2025apd,bushnaq2025spd}}, draw=tolgreen!60]]
    [{Evaluation (\cref{sec:protocol})}, fill=tolgreen!10, draw=tolgreen
      [{Synthetic ground truth; benchmarks \citep{karvonen2025saebench}; negative controls \citep{korznikov2026sanity}; metric audits \citep{chanin2026reliable}}, draw=tolgreen!60]]
    [{Failure modes (\cref{sec:failures})}, fill=tolgreen!10, draw=tolgreen
      [{Splitting, absorption, and hedging \citep{chanin2025absorption,chanin2025hedging}; seed instability; interpretability illusions; dark matter \citep{engels2025darkmatter}}, draw=tolgreen!60]]
    [{Applications (\cref{sec:applications})}, fill=tolgreen!10, draw=tolgreen
      [{Monitoring; steering \citep{wu2025axbench}; unlearning and editing; circuits \citep{ameisen2025circuit}}, draw=tolgreen!60]]
    [{Open questions (\cref{sec:open})}, fill=tolgreen!10, draw=tolgreen
      [{Stated open problems \citep{elhage2022toy,sharkey2025open}; theoretical and empirical questions}, draw=tolgreen!60]]]]
\end{forest}}
\caption{Structure of the survey. The theory sections (blue) organize results by the decoder class,
and the practice sections (green) grade the results of recovery methods on one scale of claims about
a feature. Each leaf lists representative works reviewed in that section.}
\label{fig:taxonomy}
\end{figure}

\section{Background and Definitions}
\label{sec:definitions}

A claim that a network represents more features than dimensions requires a criterion for deciding
which features are recoverable. Overlapping directions can carry feature values that a linear
decoder cannot reconstruct accurately, while orthogonal directions can produce polysemantic
neurons after a rotation. These cases require separate tests of geometry and recovery.

To test geometry and recovery separately, we use the additive feature model studied in the
report by \citet{elhage2022toy} and formalized by \citet{prieto2026correlations}. The model starts
with given input features and their joint distribution, then uses their values as coefficients
in a weighted sum of feature directions in a lower-dimensional hidden representation. A decoder attempts to
reconstruct the input feature values from that hidden vector. This formulation connects
superposition to sparse coding and compressed sensing \citep{olshausen1996,donoho2003optimally},
while making the recovery target explicit.

The feature distribution, encoder, and recovery criterion make the distinction between
superposition and coordinate-dependent polysemanticity precise. A two-feature example then
shows why decoder choice matters to recovery. Beyond deciding whether the criterion holds, quantitative
measures compare how many features are recoverable and whether overlap improves reconstruction.
Claims about trained networks require further distinctions between evidence for a feature and
the different quantities described as capacity.

\begin{tcolorbox}[summarybox,title={Summary: What Superposition Is}]
\roadmap
\begin{itemize}
\item \textbf{Geometry and recovery (\cref{def:superposition}).} The definition requires
  overlapping feature directions and a single decoder that recovers each feature at a stated
  tolerance.
\item \textbf{Coordinate dependence.} A rotation can make neurons polysemantic while preserving
  orthogonality between feature directions.
\item \textbf{Decoder dependence.} One dimension carries two sparse features at $R^2\to1$
  under a ReLU decoder, against a best linear score of $R^2=1/2$ (\cref{ex:antipodal}).
\item \textbf{Two measures (\cref{def:packing,def:benefit}).} The packing degree counts
  recoverable features per dimension; the benefit compares reconstruction with the best
  orthogonal representation at the same width.
\item \textbf{Evidence and capacity.} Six claim levels distinguish recovery from identification,
  substitution, and causal explanation. Seven capacity quantities measure different resources
  (\cref{tab:capacity}).

\end{itemize}
\end{tcolorbox}

\subsection{Setup and Notation}
\label{sec:def-objects}

We follow the notation of \citet{prieto2026correlations}, which refines the original toy-model
setting of \citet{elhage2022toy}. The setting consists of three objects: the features, the encoder
that carries them, and the decoder that reads them back.

\begin{enumerate}
\item \textbf{Features.} There are $d$ features with values $f = [f_1, \dots, f_d]^\top \in \R^d$,
drawn from a distribution $\mathcal D_f$. A common choice makes them \emph{sparse}, in which
case each feature is nonzero with a small probability $p$. When features are properties of data,
such as ``is written in French'' or ``mentions a dog,'' they arise as
$f(x) = [\rho_1(x), \dots, \rho_d(x)]^\top$ for interpretable maps $\rho_j$ on inputs $x$.
\item \textbf{Encoder.} A linear encoder $W \in \R^{m \times d}$ with $m < d$ maps features to a
hidden vector
\begin{equation}
\label{eq:code}
  h \;=\; W f \;=\; \sum_{i=1}^{d} f_i\, w_i \;\in\; \R^m ,
\end{equation}
where the columns $w_1, \dots, w_d \in \R^m$ are the \emph{feature directions}. Following the
sparse-coding literature, we refer to this assignment of features to directions as the \emph{code}.
\item \textbf{Decoder.} Any map $\psi : \R^m \to \R^d$ that reconstructs $f$ from $h$. Its quality
on feature $i$ is measured by the per-feature coefficient of determination
\begin{equation}
\label{eq:R2}
  R^2_i(W, \psi)
  \;=\;
  1 - \frac{\E\bigl[(f_i - \psi(W f)_i)^2\bigr]}{\operatorname{Var}[f_i]} .
\end{equation}
The expectation is over $\mathcal D_f$, and the criterion assumes
$0 < \operatorname{Var}[f_i] < \infty$. A score $R^2_i = 1$ means perfect recovery, while
$R^2_i = 0$ matches the constant predictor $\E[f_i]$.
\end{enumerate}

The recovery criterion also depends on which maps the decoder can use. We distinguish four
families:
\begin{equation}
\label{eq:decoderclasses}
  \Psi_{\mathrm{lin}}, \qquad \Psi_{\mathrm{lin}+\sigma}, \qquad
  \Psi_{\mathrm{sparse}}, \qquad \Psi_{\mathrm{all}} .
\end{equation}
These families comprise affine maps $\psi(h)=V^\top h+c$; affine maps followed by a specified
elementwise nonlinearity; sparse-optimization decoders such as basis pursuit; and all measurable
maps. The inference map of a ReLU sparse autoencoder belongs to the second family.

These names describe decoding procedures rather than a chain of set inclusions. A fixed ReLU
output excludes affine maps with negative outputs, and basis pursuit is defined by a different
optimization problem. The identity choice for $\sigma$ recovers the affine family when that
choice is allowed. We compare the families under their stated activation, input, and accuracy
conditions in \cref{sec:access}.

The tied ReLU autoencoder in the report by \citet{elhage2022toy} fixes one such decoding
procedure. Related toy models vary the loss, feature law, or activation function
\citep{scherlis2022capacity,lecomte2024incidental,prieto2026correlations}. In the original model,
the encoder is $W$, and the decoder is $W^\top$ followed by a bias and a ReLU. Training minimizes
\begin{equation}
\label{eq:reluae}
  \cL_{\mathrm{ReLU\text{-}AE}}(f, W, b)
  \;=\;
  \bigl\lVert\, f - \mathrm{ReLU}(W^\top W f + b) \,\bigr\rVert_2^2 .
\end{equation}
Two built-in choices of \cref{eq:reluae} carry quantitative consequences, which \cref{sec:geometry}
develops. First, the decoder directions are the encoder's own columns rather than parameters of
their own, which makes the readout \emph{tied}. Second, the decoder class is
$\Psi_{\mathrm{lin}+\sigma}$, the second class of \cref{eq:decoderclasses}.

\Cref{tab:notation} collects the notation used below. In particular, $d$ counts features and $m$
denotes hidden width; these symbols have different meanings in some of the cited sources.

\begin{table}[tbp]
\centering\small
\renewcommand{\arraystretch}{1.25}
\caption{Notation used throughout the survey, following \citet{prieto2026correlations}.}
\label{tab:notation}
\begin{tabularx}{\textwidth}{@{}l X@{}}
\toprule
Symbol & Meaning \\
\midrule
$d$ & number of features \\
$f = [f_1, \dots, f_d]^\top \sim \mathcal D_f$ & feature values and their distribution \\
$p = \Prob(f_i \ne 0)$;\; $k$ & activation probability; hard sparsity level ($\norm{f}_0 \le k$) \\
$\rho_j : \mathcal X \to \R$ & interpretable property of the data inducing feature $j$ \\
$m$ & hidden width ($m < d$) \\
$W = [\,w_1, \dots, w_d\,] \in \R^{m \times d}$ & linear encoder; column $w_i$ is the direction of feature $i$ \\
$h = W f \in \R^m$;\; $\xi$ & hidden representation; residual (unmodeled structure or noise) \\
$\psi : \R^m \to \R^d$ & decoder; linear decoders written $\psi(h) = V^\top h + c$ \\
$\Psi_{\mathrm{lin}},\ \Psi_{\mathrm{lin}+\sigma},\ \Psi_{\mathrm{sparse}},\ \Psi_{\mathrm{all}}$ & decoder families, \cref{eq:decoderclasses} \\
$R^2_i(W, \psi)$;\; $\varepsilon$ & per-feature recovery quality, \cref{eq:R2}; tolerance \\
$G = W^\top W$;\; $M = V^\top W$ & Gram matrix (feature geometry); matrix governing the linear readout \\
$\Rlin(W, V)$ & risk of the linear readout, \cref{eq:riskidentity} \\
$C_i = \norm{w_i}^4 / \sum_j \ip{w_i}{w_j}^2$ & feature capacity, \cref{sec:geo-scherlis} \\
$\kappa_\varepsilon(W; \Psi)$;\; $\Sup(W)$ & packing degree (\cref{def:packing}); superposition benefit (\cref{def:benefit}) \\
\bottomrule
\end{tabularx}

\end{table}

\subsection{Definitions}
\label{sec:def-superposition}

The distinction between overlapping and rotated orthogonal representations requires a definition
in terms of feature directions. We adopt the geometric and recovery conditions of
\citet{prieto2026correlations}, with recovery measured by \cref{eq:R2}.

\begin{definition}[Superposition \citep{prieto2026correlations}]
\label{def:superposition}
Fix a tolerance $\varepsilon > 0$. A set of features $F \subseteq [d]$ is represented \emph{in
superposition} by the encoder $W$ if:
\begin{enumerate}
\item \textup{(Interference)} for every $i \in F$ there is some $j \in F$, $j \ne i$, with
  $\ip{w_i}{w_j} \ne 0$;
\item \textup{(Recoverability)} there is a single decoder $\psi : \R^m \to \R^d$ with
  $R^2_i(W, \psi) \ge 1 - \varepsilon$ for every $i \in F$.
\end{enumerate}
\end{definition}

\begin{figure}[t]
\centering
\includegraphics[width=\textwidth]{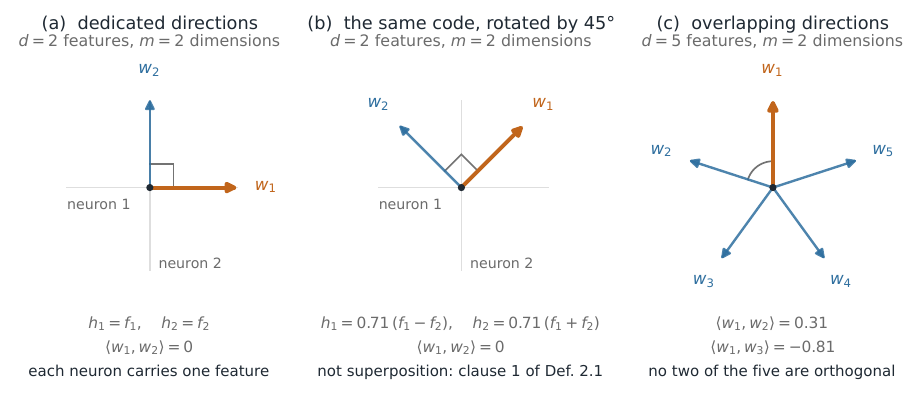}
\caption{Feature directions in a two-dimensional hidden space and their relation to neuron
responses. \textbf{(a)} Two features occupy the two neuron axes, so the encoding $h=Wf$ gives
each neuron one feature value. \textbf{(b)} A $45^\circ$ rotation makes each neuron respond to
both features while preserving the orthogonality of their directions and all encoded information.
The rotated code therefore still fails clause 1 of \cref{def:superposition}.
\textbf{(c)} Five directions share the same two-dimensional space; the displayed nonzero
inner products $\ip{w_i}{w_j}$ show the overlap that is unavoidable when features outnumber
dimensions.}
\label{fig:intuition}
\end{figure}

The two clauses distinguish geometric overlap from recovery. \Cref{fig:intuition} shows why
the geometric clause concerns feature directions rather than neuron responses. Clause 1
excludes codes that merely \emph{appear} shared. For example, consider two features stored along
orthogonal directions ($W = I_2$, \cref{fig:intuition}a) and rotate that code by $45^\circ$
(\cref{fig:intuition}b). Every neuron then responds to both features, but the directions remain
orthogonal, and rotating back restores the original code with no information lost. Polysemantic
neuron responses are therefore a property of the coordinate system, whereas the inner products
$\ip{w_i}{w_j}$ on which \cref{def:superposition} is stated are left unchanged by a rotation. Under
clause 1, such codes are not in superposition, however polysemantic their neurons appear.

Clause 2 excludes collections of directions whose feature values cannot be recovered at the
chosen tolerance. It requires one decoder to meet the criterion for every feature in $F$, with
errors averaged over the same distribution. This condition differs from exact recovery on every
possible input. The geometric definition can also hold for a subset with $|F|\le m$; the packing
degree below distinguishes the case with more recoverable features than dimensions.

The recoverability clause permits any decoder. The next definition distinguishes features
according to whether a linear decoder can attain the tolerance. Failure of linear recovery
leaves nonlinear procedures to be evaluated under their own assumptions.

\begin{definition}[Linear and non-linear superposition \citep{prieto2026correlations}]
\label{def:linsup}
Let $F$ be a set of features in superposition. A feature $i \in F$ is in \emph{linear
superposition} if some linear decoder $\psi_{\mathrm{lin}} \in \Psi_{\mathrm{lin}}$ achieves
$R^2_i \ge 1 - \varepsilon$; it is in \emph{non-linear superposition} if some decoder in
$\Psi_{\mathrm{all}}$ achieves $R^2_i \ge 1 - \varepsilon$ but no linear decoder achieves it.
\end{definition}

Both definitions assume that the feature directions, the columns of $W$, are already specified.
In toy models, these directions are explicit parameters whose values can be inspected after
training. In a trained network with unknown features, applying the definitions
first requires a hypothesis relating input properties to hidden activations, followed by an
estimate of the corresponding directions. The linear representation hypothesis supplies this
relation by treating high-level concepts as directions in the hidden space.

\begin{definition}[Linear representation hypothesis \citep{park2024linear, prieto2026correlations}]
\label{def:lrh}
A hidden representation $h$ satisfies the \emph{linear representation hypothesis} (LRH) with
respect to properties $\{\rho_j\}_{j=1}^{d}$ if there are directions
$w_1, \dots, w_d \in \R^m$ with
\[
  h(x) \;\approx\; \sum_{j=1}^{d} \rho_j(x)\, w_j
  \qquad \text{for } x \sim \mathcal D_x .
\]
\end{definition}

If the LRH assigns more than $m$ nonzero directions to a layer, at least one pair is
nonorthogonal. This count does not ensure that every feature overlaps another, as clause 1
requires for the chosen set $F$, or that clause 2 holds. Recovery remains a separate condition.
Likewise, a downstream layer may use a feature without reconstructing its value to the specified
tolerance. We distinguish this causal use from recovery throughout the empirical sections.

The directions are specified in the definitions but must be estimated when studying a trained
network. \Cref{sec:relations} develops the assumptions behind this transfer, and
\cref{sec:recovery} reviews methods for estimating the directions.

\subsection{A Worked Example}
\label{sec:def-example}

We now check \cref{def:superposition,def:linsup} on a code small enough to compute in closed form.

\begin{figure}[t]
\centering
\includegraphics[width=\textwidth]{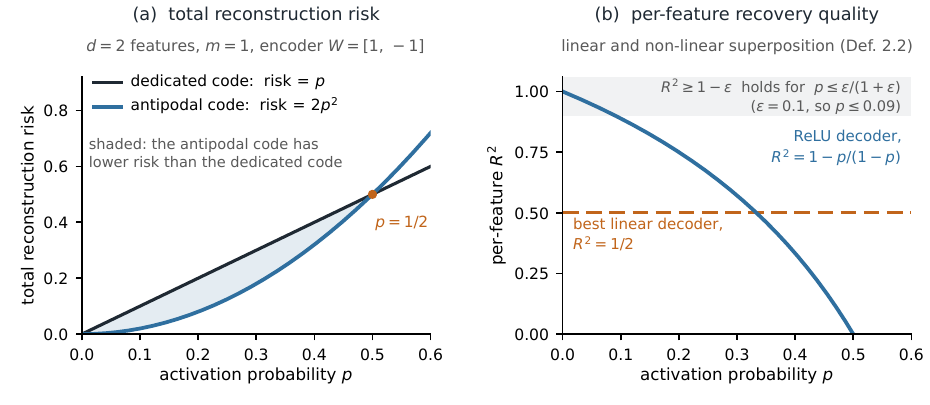}
\caption{Reconstruction and recovery in the one-dimensional codes of \cref{ex:antipodal}.
\textbf{(a)} For two $\mathrm{Bernoulli}(p)$ features, the dedicated code drops one feature and
incurs risk $p$, whereas the antipodal code keeps both features and incurs $2p^2$.
The antipodal error occurs when the two features fire together, an event of probability $p^2$
on which the ReLU decoder errs on each feature. \textbf{(b)} Despite this error, the ReLU decoder achieves
$R^2 = 1 - \frac{p}{1-p} \to 1$ as features become sparse, while the best linear decoder is capped
at $R^2 = 1/2$ for every $p$. For $\varepsilon < \tfrac12$ and small $p$, these features are
therefore in non-linear superposition by \cref{def:linsup}.}
\label{fig:antipodal}
\end{figure}

\begin{example}[Two features in one dimension]
\label{ex:antipodal}
Let $d = 2$, $m = 1$, with $f_1, f_2$ independent $\mathrm{Bernoulli}(p)$ with $0<p<1$. Use the antipodal encoder
$W = [\,1, -1\,]$, so $h = f_1 - f_2$ and $\ip{w_1}{w_2} = -1$: clause 1 of
\cref{def:superposition} holds. We compare two decoders for this code, and one baseline code that
forgoes superposition (\cref{fig:antipodal}):
\begin{enumerate}
\item \emph{ReLU decoder}, $\psi(h) = (\mathrm{ReLU}(h), \mathrm{ReLU}(-h))$. It is exact on the
patterns $(0,0)$, $(1,0)$, and $(0,1)$, and it errs only when both features fire, which happens
with probability $p^2$. On that event $h = 0$, so the decoder returns $(0,0)$ and reads both
features as absent. The risk of each feature is therefore $p^2$. With
$\operatorname{Var}[f_i] = p(1-p)$,
\[
  R^2_i \;=\; 1 - \frac{p^2}{p(1-p)} \;=\; 1 - \frac{p}{1-p} \;\longrightarrow\; 1
  \quad (p \to 0),
\]
so for any tolerance $\varepsilon$, once $p \le \varepsilon/(1+\varepsilon)$ the set
$F = \{1, 2\}$ is in superposition: two usable features in one dimension.
\item \emph{Best linear decoder.} The best affine prediction of $f_1$ from the scalar $h$
achieves exactly $R^2 = \operatorname{corr}(f_1, h)^2 = \tfrac12$ for every $p$. The best affine
prediction of $f_2$ achieves the same value. Linear readout is therefore capped at
$R^2 = \tfrac12$ regardless of sparsity. For $\varepsilon < \tfrac12$ and
$p \le \varepsilon/(1+\varepsilon)$, these features are in \emph{non-linear} superposition
(\cref{def:linsup}). The nonlinearity of the decoder is essential to the code.
\item \emph{Dedicated baseline.} Store $f_1$ exactly and decode $f_2$ as $0$: the total risk is
$p$. The antipodal code with the ReLU decoder errs on both features whenever both fire, so its
total risk is $2p^2$. The antipodal code therefore has lower risk than the dedicated baseline
exactly when $p < \tfrac12$.
\end{enumerate}
\end{example}

The example shows why the input distribution and decoder belong in a recovery claim. The ReLU
decoder removes cross-talk on single-feature inputs and incurs error when the features
co-activate. A linear probe retains a fixed normalized error even as co-activation becomes rare.
A low linear-probe score therefore need not imply that a feature is absent from the
representation. \Cref{sec:geometry} examines related geometries with more feature directions.

\subsection{Measures of Superposition}
\label{sec:def-degrees}

The definition determines whether a chosen feature set is in superposition, but it does not
quantify how many features are recoverable or whether overlap improves reconstruction. We
introduce two measures in this survey to answer these separate questions.

\begin{definition}[Packing degree]
\label{def:packing}
For tolerance $\varepsilon$ and decoder class $\Psi$, the \emph{packing degree} of $W$ is
\[
  \kappa_\varepsilon(W; \Psi)
  \;=\;
  \frac{1}{m}\, \max\Bigl\{ |F| : F \text{ satisfies both clauses of \cref{def:superposition}
  with some } \psi \in \Psi \Bigr\} ,
\]
which is the number of simultaneously usable features per dimension (with $\kappa_\varepsilon = 0$
when no feature set satisfies both clauses).
\end{definition}

A value $\kappa_\varepsilon > 1$ states that the code sustains more readable features than its
width, which no orthogonal code can achieve. For example, in \cref{ex:antipodal},
$\kappa_\varepsilon = 2$ under $\Psi_{\mathrm{all}}$ for small $p$, while under
$\Psi_{\mathrm{lin}}$ at any $\varepsilon < \tfrac12$ no feature set satisfies both clauses. The
degree of superposition of a code thus depends on the decoder class reading it, and
\cref{sec:access} quantifies that dependence.
\begin{definition}[Superposition benefit]
\label{def:benefit}
Fix $\mathcal D_f$ and a decoder class $\Psi$, and let
$R^\star(W) = \inf_{\psi \in \Psi} \E \lVert f - \psi(W f) \rVert_2^2$. The \emph{superposition
benefit} of $W$ is
\[
  \Sup(W)
  \;=\;
  \inf_{\substack{W' \colon\, W'^\top W' \ \mathrm{diagonal}}} R^\star(W')
  \;-\;
  R^\star(W) .
\]
\end{definition}

The comparison uses encoders $W'$ with the same dimensions as $W$. A diagonal Gram matrix then
permits at most $m$ nonzero columns, so the baseline optimizes over orthogonal representations
at the same width. A positive $\Sup$ means that $W$ improves on this baseline under the chosen
distribution and decoder family.

The difference $p-2p^2$ in \cref{ex:antipodal} compares two specified reconstruction procedures.
It is not automatically the optimized benefit in \cref{def:benefit}. For example, a decoder that
predicts the mean of a dropped Bernoulli feature improves on the zero prediction used in that
example. This distinction also matters for phase diagrams, which report solutions found by
training rather than the two infima defining $\Sup$ (\cref{sec:acc-when}).

\subsection{Strength of Claims about Features}
\label{sec:def-ladder}

The measures above concern a specified representation. Empirical claims about features can
instead concern recovery, the identity of a direction, or its role in a computation. We use six
claim levels to distinguish the evidence needed as an explanation becomes more detailed.
These levels describe evidential commitments rather than a universal chain of logical
implications.

\begin{description}[leftmargin=1.6em, style=sameline, itemsep=1pt]
\item[L1 (representation).] $f_i$ is statistically encoded in $h$: some decoder improves on a
  specified predictor that does not use $h$.
\item[L2 (accessibility).] $R^2_i \ge 1 - \varepsilon$ within a stated decoder class. This is the
  recoverability clause of \cref{def:superposition}; \cref{def:linsup} is the L2 distinction
  between $\Psi_{\mathrm{lin}}$ and $\Psi_{\mathrm{all}}$.
\item[L3 (identification).] The directions $w_i$ themselves are recovered from samples of $h$, up
  to permutation, sign, and scale.
\item[L4 (functional substitution).] Replacing $h$ by its reconstruction preserves the downstream
  model's behavior.
\item[L5 (causal deployment).] Intervening on the feature (ablation, steering) moves behavior
  through the claimed pathway.
\item[L6 (mechanistic faithfulness).] The feature-level account predicts internal and behavioral
  counterfactuals it was not fitted to.
\end{description}

The distinctions prevent a positive result from answering a different question without further
evidence. Feature values can be decodable while the generating dictionary remains
non-identifiable (L2 without L3, \cref{sec:acc-ident}). A reconstruction-preserving substitute
need not reproduce the original computation (L4 without L6, \cref{sec:failures}). Identification
and intervention tests also address different targets: recovering a direction does not by
itself establish how the network uses that direction.

\subsection{Notions of Capacity}
\label{sec:def-capacity}

Evidence about an individual feature does not determine how many features a layer can support.
The word \emph{capacity} covers several different counts and geometric statistics in this
literature. \Cref{tab:capacity} separates seven quantities by the resource or recovery target
that each constrains.

Feature capacity and the Welch bound concern overlaps between encoder columns. Recovery widths
also depend on the allowed input patterns and decoding procedure, while information-theoretic
counts need a precision or noise model. Computational bounds add a specified family of functions
and assumptions about the resources counted. These distinctions organize the next three sections,
beginning with the relation between feature overlaps and reconstruction error.

\begin{table}[tbp]
\centering\small
\renewcommand{\arraystretch}{1.3}
\caption{Seven quantities used in discussions of capacity, in the notation of
\cref{tab:notation}. Their input assumptions, recovery criteria, and resource counts differ.}
\label{tab:capacity}
\begin{tabularx}{\textwidth}{@{}>{\raggedright\arraybackslash}p{0.22\textwidth} >{\raggedright\arraybackslash}p{0.36\textwidth} X@{}}
\toprule
Quantity & Definition and source & Constrains \\
\midrule
Overcompleteness ratio $d/m$ &
Ratio of features to dimensions. &
Counts candidate directions; does not by itself specify recovery or information content. \\
Feature capacity $C_i$; bound $\sum_i C_i \le m$ \status{PROVED} &
$C_i = \norm{w_i}^4 / \sum_j \ip{w_i}{w_j}^2 \in (0,1]$; $\sum_i C_i \le \operatorname{rank} W$
\citep{scherlis2022capacity}; identical to the feature dimensionality of \citet{elhage2022toy}
(\cref{sec:geo-scherlis}). &
Geometric allocation for nonzero columns; does not bound task loss or causal use. \\
Welch bound \status{PROVED} &
$\max_{i \ne j} |\ip{w_i}{w_j}| \ge \sqrt{\smash[b]{\tfrac{d-m}{m(d-1)}}}$ \citep{welch1974}. &
Worst-case pairwise overlap for unit-norm columns; related total-overlap bounds appear in
\cref{sec:geo-budget}. \\
Linear-access width \status{PROVED} &
Least $m$ such that one linear map recovers every $k$-sparse $f \in [-1,1]^d$ to
$\ell_\infty$-error $\varepsilon$: $O(\varepsilon^{-2} k^2 \log d)$ and
$\Omega_\varepsilon\!\bigl(\tfrac{k^2}{\log k} \log \tfrac{d}{k}\bigr)$ in the regime of \cref{res:garg} \citep{garg2026howmany}. &
The cost of simultaneous uniform \emph{linear} readout: quadratic in $k$. \\
Sparse-recovery width \status{PROVED} &
$m = O(k \log(d/k))$ suffices with a decoder in $\Psi_{\mathrm{sparse}}$ and a random or RIP
dictionary \citep{candes2006stable, donoho2003optimally}. &
Linear in $k$; the gap to the row above is the decoder class, not a contradiction. \\
Information-theoretic capacity \status{PROVED} &
With coordinate range $R$ and resolution $\delta$: $\log_2 N \le m \log_2(1 + 2R/\delta)$
distinguishable messages. &
Why finite width alone constrains nothing without a precision model (\cref{sec:geo-precision}). \\
Computational capacity \status{PROVED} &
Realizing the Boolean family in \cref{res:separation} imposes
a $d'=O(m^2/\log m)$ expressivity limit under its depth and parameter assumptions; representing
$d'$ Boolean features uses $O(\log d')$ dimensions \citep{adler2024complexity}. &
Resources for realizing a function family; distinct constructions and quantifiers are compared
in \cref{sec:computation}. \\
\bottomrule
\end{tabularx}

\end{table}

\takeaway{sec:definitions}{To compare recovery claims, fix the feature distribution, accuracy
criterion, and decoder family. A low linear-probe score can coexist with accurate nonlinear
recovery, so the score does not establish feature absence. To interpret a recovered direction,
distinguish its predictability from identification and causal use. Capacity comparisons also
require the same resource count: overlapping directions, recoverable feature values, and
computable functions answer different questions.}

\section{Geometry, Interference, and Capacity}
\label{sec:geometry}

The definitions separate feature overlap from successful recovery. For a specified encoder and
readout, the contribution of overlap to reconstruction error depends on the feature covariance.
Unit-norm feature directions obey a lower bound on total squared overlap, but sparsity and nonlinear
decoding can change how this overlap affects recovery. We first examine this distinction for linear
readouts, then compare the effects of feature statistics and decoding procedures. The resulting
conditions constrain possible toy representations, leaving a further question about which geometries
training actually reaches.

\begin{tcolorbox}[summarybox,title={Summary: Lower Bounds on Interference}]
\roadmap
\begin{itemize}
\item \textbf{Geometric overlap (\cref{res:budget}).} For $d$ unit-norm directions in
  $m<d$ dimensions, total squared overlap is at least $d(d-m)/m$.
\item \textbf{Reconstruction risk (\cref{eq:riskidentity}).} Feature covariance and the
  readout determine how this geometry affects linear coefficient reconstruction.
\item \textbf{Tied and untied readouts (\cref{res:untied}).} At a unit-norm tight frame,
  tied linear risk is $d/m$ times optimal untied risk for isotropic noiseless features.
\item \textbf{Nonlinear recovery.} Sparse nonnegative inputs permit exact recovery in some
  overlapping constructions, while other support patterns cause collisions.
\item \textbf{Different counts.} Finite-precision message counts, pairwise direction packing,
  and uniform sparse recovery impose different constraints on width.

\end{itemize}
\end{tcolorbox}

\subsection{Interference and Linear Reconstruction Risk}
\label{sec:geo-risk}

Let $W\in\R^{m\times d}$ have feature directions $w_i$, as in \cref{tab:notation}. To allow
unmodeled structure or noise, we extend \cref{eq:code} to $h=Wf+\xi$; the original model has
$\xi=0$. A linear readout uses directions $V=[v_1,\ldots,v_d]\in\R^{m\times d}$ and predicts
$V^\top h$. Its prediction for feature $i$ separates into signal, cross-talk, and residual terms:

\begin{equation}
\label{eq:interference}
  v_i^\top h
  \;=\;
  \underbrace{\ip{v_i}{w_i}\, f_i}_{\text{signal}}
  \;+\;
  \underbrace{\textstyle\sum_{j \ne i} \ip{v_i}{w_j}\, f_j}_{\text{cross-talk}}
  \;+\;
  \underbrace{v_i^\top \xi}_{\text{residual}} .
\end{equation}
Each cross-talk contribution depends on an overlap between a readout direction $v_i$ and a
feature direction $w_j$. The product $M=V^\top W$ collects these overlaps. By contrast, the
Gram matrix $G=W^\top W$ records overlaps between feature directions themselves. The two
matrices coincide for a tied readout, $V=W$, but an independently chosen readout can respond
differently to the same representation.

The feature distribution determines how often these contributions occur and how they combine.
To express their average effect on reconstruction, suppose that the feature vector $f$ and the
residual $\xi$ are independent and each has mean zero, with covariance matrices $\Sigma_f$ and
$\Sigma_\xi$, respectively. The expected squared error is

\begin{equation}
\label{eq:riskidentity}
  \Rlin(W, V)
  \;\coloneqq\;
  \E\,\bigl\lVert V^\top h - f \bigr\rVert_2^2
  \;=\;
  \operatorname{tr}\!\bigl( (M - I)\, \Sigma_f\, (M - I)^\top \bigr)
  \;+\;
  \operatorname{tr}\!\bigl( V^\top \Sigma_\xi V \bigr) .
\end{equation}
The first term measures errors in reconstructing feature coefficients; the second measures the
effect of the additive residual on the readout. Although $M$ has rank at most $m<d$, the risk
also depends on $\Sigma_f$. An isotropic feature law assigns variance to every feature-space
direction, whereas a singular covariance can restrict variation to a recoverable subspace.

Average risk also differs from error on a particular support pattern. In
\cref{ex:antipodal}, a nonlinear decoder succeeds on single-feature inputs and fails on a
co-activation. The next example develops this distinction for three directions in a plane,
a geometry observed in trained toy models in the report by \citet{elhage2022toy}.

\begin{example}[Three features in two dimensions]
\label{ex:simplex}
Let $d = 3$, $m = 2$, and place unit directions $w_1, w_2, w_3 \in \R^2$ at mutual angles $120^\circ$,
so $\ip{w_i}{w_j} = -\tfrac12$ for $i \ne j$. Take nonnegative features, $f \ge 0$, with the tied
nonlinear decoder $\psi(h) = \mathrm{ReLU}(W^\top h)$. If exactly one feature is active, $f = t e_i$
with $t \ge 0$, then $W^\top h = t\,(e_i - \tfrac12 e_j - \tfrac12 e_k)$, and the ReLU deletes the
negative cross-talk. Recovery is then \emph{exact} for three features in two dimensions, at every
amplitude. But if all three fire equally, $f = t\,(1,1,1)$, then $h = t\,(w_1 + w_2 + w_3) = 0$, so
the fully active pattern collides with the empty one and no decoder in any class can separate them.
\end{example}

\subsection{Lower Bounds on Interference}
\label{sec:geo-budget}

The examples use particular feature arrangements. To compare them with other arrangements,
we need a bound that holds for all unit-norm columns. Classical frame theory provides such a
bound through the eigenvalues of $W^\top W$ \citep{welch1974,benedetto2003frames}. The same
bound gives a reconstruction-risk statement after the feature covariance and readout are fixed.

\begin{result}[Interference lower bound; tied readout]
\label{res:budget}
\status{PROVED (classical)}
Let $W \in \R^{m \times d}$ have unit-norm columns, $d > m$. Then
\begin{equation}
\label{eq:budget}
  \sum_{i \ne j} \ip{w_i}{w_j}^2 \;\ge\; \frac{d\,(d-m)}{m},
\end{equation}
with equality if and only if $W$ is a unit-norm tight frame, $W W^\top = \tfrac{d}{m}\, I_m$.
Consequently, by \cref{eq:riskidentity} with tied readout $V = W$, isotropic features
$\Sigma_f = \sigma^2 I$, and $\xi = 0$,
\begin{equation}
\label{eq:tiedfloor}
  \Rlin(W, W) \;=\; \sigma^2\, \norm{W^\top W - I}_F^2 \;\ge\; \sigma^2\, \frac{d\,(d-m)}{m},
\end{equation}
and the worst pairwise overlap obeys the Welch bound
$\max_{i \ne j} |\ip{w_i}{w_j}| \ge \sqrt{(d-m)/(m(d-1))}$.
\end{result}

The normalization is essential to \cref{eq:budget}. Unit-norm columns give
$\operatorname{tr}(W^\top W)=d$, while the Gram matrix has at most $m$ nonzero eigenvalues.
Cauchy--Schwarz then gives $\|W^\top W\|_F^2\ge d^2/m$. Subtracting the diagonal contribution
$d$ yields the total-overlap bound. Tight frames attain this minimum, although their individual
pairwise overlaps need not be equal.

The risk bound in \cref{eq:tiedfloor} additionally uses isotropic features and the tied linear
readout. An untied readout can reduce the error by changing $V^\top W$ while leaving
$W^\top W$ unchanged. The relevant limit is then the rank of $V^\top W$.

\begin{result}[Lower bound for arbitrary linear readouts]
\label{res:untied}
\status{PROVED (classical, via best low-rank approximation)}
For \emph{every} pair $W, V \in \R^{m \times d}$ in the isotropic noiseless setting of
\cref{res:budget} ($\Sigma_f = \sigma^2 I$, $\xi = 0$), the matrix $M = V^\top W$ has rank at most
$m$. The best rank-$m$ Frobenius approximation of $I_d$ leaves residual $d - m$; hence
\begin{equation}
\label{eq:untiedfloor}
  \Rlin(W, V) \;\ge\; \sigma^2\, (d - m)
  \qquad \text{for all } W, V .
\end{equation}
A unit-norm tight frame attains \emph{both} lower bounds simultaneously: with
$V_\star = W (W^\top W)^{+}$, the matrix $V_\star^\top W$ is the orthogonal projection onto the
row space of $W$, giving $\Rlin(W, V_\star) = \sigma^2 (d - m)$, while
$\Rlin(W, W) = \sigma^2\, d(d-m)/m$. The ratio of the two bounds is the overcompleteness ratio $d/m$:
\begin{equation}
\label{eq:priceoftying}
  \frac{\Rlin(W, W)}{\Rlin(W, V_\star)} \;=\; \frac{d}{m}
  \qquad \text{at any unit-norm tight frame.}
\end{equation}
\end{result}

\doesnot{These bounds concern linear reconstruction of isotropic feature coefficients at one
specified site. They do not directly bound a downstream task loss or a nonlinear decoder's
error. The optimal untied map projects onto an $m$-dimensional subspace of feature space,
leaving error in the remaining directions. Applying the result to several layers requires
specifying the joint representation and its width.}

\begin{figure}[t]
\centering
\includegraphics[width=\textwidth]{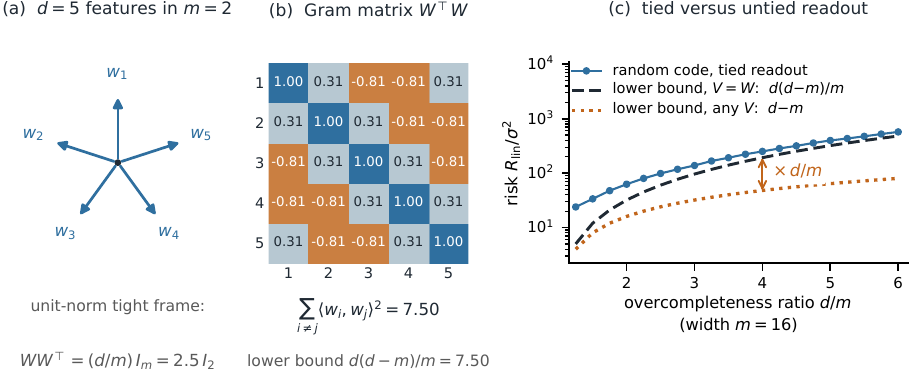}
\caption{Interference and linear-readout risk for overcomplete feature codes.
\textbf{(a)} Arranging five features as a regular pentagon in two dimensions gives a unit-norm
tight frame satisfying $W W^\top = (d/m)\, I_m$. \textbf{(b)} The off-diagonal entries of its Gram
matrix give the pairwise interferences, whose squared sum meets the lower bound
$d(d-m)/m = 7.5$ of \cref{res:budget}. \textbf{(c)} At width $m = 16$, the risk curves compare
tied and untied linear readouts across overcompleteness ratios $d/m$, with risk measured in units of $\sigma^2$.
Random codes read with the tied
readout $V = W$ sit above the tied-readout lower bound $d(d-m)/m$. The optimal untied
readout for a full-row-rank encoder attains $d-m$. The ratio $d/m$ is attained by the two
risks at a unit-norm tight frame, rather than by every random encoder (\cref{res:untied}).}
\label{fig:budget}
\end{figure}

For a unit-norm tight frame with $d>m$, \cref{eq:priceoftying} quantifies the improvement from
optimizing a linear readout under the isotropic noiseless feature law. Other encoders can have
a different ratio, and the equation does not describe the nonlinear ReLU loss in
\cref{eq:reluae}. At $d=m$, an orthonormal basis gives zero risk for both readouts, so the
risk ratio is undefined.

Sparse inputs lead to a different comparison. For unit-norm columns and a tied readout,
coherence $\mu=\max_{i\ne j}|\langle w_i,w_j\rangle|$ gives a sufficient cross-talk bound
of $k\mu$ when at most $k$ feature values are nonzero and each lies in $[-1,1]$. Requiring $k\mu\le\varepsilon$ provides one way to
certify uniform accuracy. When $d\ge2m$, the Welch bound limits this certificate to widths
of order at least $k^2/\varepsilon^2$. This argument bounds the usefulness of the coherence
certificate; it does not establish a lower bound for every linear recovery procedure.
\Cref{sec:acc-linear} presents the general linear-access result of \citet{garg2026howmany}.

\subsection{Mechanisms that Reduce the Realized Cost}
\label{sec:geo-escapes}

The preceding risk bounds fix both a linear decoder and a feature law. The toy-model literature
studies how reconstruction changes when these choices vary. We begin with sparse activation,
whose effect depends on whether the decoder suppresses cross-talk on inactive features.

\paragraph{Sparsity and Co-Activation.}
In a linear readout, one active feature can produce error on several inactive coordinates.
Sparsity reduces how often this error occurs, but does not by itself require two features to be
active. A suitable nonlinear decoder can remove some of these single-feature errors. In the
antipodal construction, the remaining error requires co-activation, which occurs with probability
$p^2$ for independent Bernoulli features. The difference between $p$ and $p^2$ therefore
depends on both the support law and the decoding geometry.

Other toy models express the statistical condition through moments of the feature law.
In the quadratic model of \citet{scherlis2022capacity}, a fourth-moment threshold determines
whether fractional allocation of dimensions improves the objective. Sparsity can raise this
moment ratio, but the result concerns the specified quadratic loss rather than every sparse
feature distribution. \Cref{sec:acc-when} develops the allocation result and its learning scope.
\status{PROVED in the toy model}

\paragraph{Nonlinear Readout.}
For nonnegative features, a ReLU removes negative pre-activations that would otherwise count as
errors on inactive coordinates. Antipodal pairs and simplices exploit this sign structure,
as \cref{ex:antipodal,ex:simplex} show. A negative bias can also remove small positive cross-talk,
although the same bias shrinks active signals. The nonlinear decoder trades these effects
against one another under the feature distribution.

This recovery remains conditional on the input patterns. The simplex decoder is exact on
single-feature inputs, but an equal co-activation of all three features produces the same hidden
state as the empty input. Increasing the common amplitude makes its reconstruction error
arbitrarily large. Nonlinearity can improve sparse recovery without removing collisions from
the representation.
\status{PROVED in small cases; NUMERICALLY-OBSERVED at scale}

\paragraph{Untied Readout and Feature Importance.}
Optimizing the linear readout is a separate way to reduce error. In the isotropic noiseless
setting, \cref{res:untied} gives the best attainable rank-limited risk and the exact improvement
at unit-norm tight frames. This optimization changes the decoder while preserving the encoder.

Importance weights instead change which reconstruction errors training penalizes. In the
quadratic toy model of \citet{scherlis2022capacity}, sufficiently important features receive
dedicated dimensions, weakly weighted features are dropped, and an intermediate group can share
dimensions. A common marginal-loss condition determines the allocation. This result explains
how a fixed-width representation can favor some features over others; its precise threshold
and assumptions are discussed in \cref{sec:acc-when}.
\status{PROVED in the toy model}

\paragraph{Feature Correlation.}
Beyond marginal sparsity and importance, correlations change which features are nonzero on
the same input and which feature combinations the decoder must reconstruct. Features that rarely co-activate can share opposite directions with
little realized interference. Frequently co-active features may instead require separation or
may support reconstruction through positively aligned directions. The outcome depends on the
feature law and training objective, rather than on correlation alone.

The report by \citet{elhage2022toy} describes changes in toy-model geometry under
feature correlation. \citet{prieto2026correlations} further study cases in which correlated
features contribute constructively to one another's reconstruction. Under bottleneck pressure
or weight decay, their models produce clusters and cyclic arrangements resembling structures
reported in language models. A recovery method that seeks separate directions for correlated
properties may nevertheless regard their merging as a failure. \Cref{sec:failures} returns to
this dependence on the chosen feature target.
\status{NUMERICALLY-OBSERVED}

\subsection{Feature Capacity}
\label{sec:geo-scherlis}

The lower bound of \cref{sec:geo-budget} sums squared overlaps across the whole code. To describe
how those overlaps are distributed among features, the toy-model literature uses a per-feature
statistic comparing each column's squared self-overlap with its total squared overlap against
all columns.

\begin{result}[Feature capacity bound \citep{scherlis2022capacity}]
\label{res:scherlis}
\status{PROVED}
For any finite $W$ with nonzero columns, define
$C_i = \norm{w_i}^4 / \sum_j \ip{w_i}{w_j}^2$. Then $C_i \in (0, 1]$, with $C_i = 1$ exactly when
$w_i$ is orthogonal to every other column, and
\[
  \sum_{i=1}^{d} C_i \;\le\; \operatorname{rank}(W) \;\le\; m .
\]
\end{result}

$C_i$ compares the squared self-overlap of $w_i$ with its total squared overlap against all
columns. It equals one for an isolated orthogonal direction and decreases when other directions
overlap it. The interpretation as a fractional dimension is geometric; its sum is bounded by
the rank. At $d = m$ with orthogonal columns every $C_i$ equals $1$ and
the sum equals $m$, so the code that gives every feature a dimension of its own attains the bound.
When the bound is saturated more generally, the columns form tight frames within orthogonal
blocks \citep{scherlis2022capacity}, which is the abstract version of the arrangements that
\cref{sec:geo-shapes} reports from trained toy models.

\citet{elhage2022toy} call the same quantity \emph{feature dimensionality}, defined by
\[
D_i=\frac{\norm{w_i}^2}{\sum_j\ip{\widehat w_i}{w_j}^2},
\qquad \widehat w_i=\frac{w_i}{\norm{w_i}}.
\]
Multiplying the numerator and the denominator of $D_i$ by $\norm{w_i}^2$ gives
\[
  D_i \;=\; \frac{\norm{w_i}^4}{\sum_j \ip{w_i}{w_j}^2} \;=\; C_i .
\]
The central quantities of \citet{scherlis2022capacity} and \citet{elhage2022toy} are therefore one
functional. The fractions at which toy-model features cluster, $\tfrac12$ for antipodal pairs,
$\tfrac23$ for triangles, $\tfrac34$ for tetrahedra, and $\tfrac25$ for pentagons, are the values of
$C_i$ on those configurations.

\doesnot{Feature capacity is a geometric statistic of the encoder. It does not specify a
decoder, reconstruction accuracy, causal use, or Shannon capacity. The optimal-allocation and
fourth-moment results of \citet{scherlis2022capacity} additionally require their independent-feature
quadratic toy model.}

\subsection{Finite Precision and Counting Bounds}
\label{sec:geo-precision}

The geometric bounds above constrain overlaps between feature directions; a reconstruction
bound additionally specifies the feature law and decoder. A count of distinguishable messages
requires a different set of assumptions, concerning precision or noise. With unrestricted real
precision, finite dimension alone does not provide a finite message count. If each coordinate
instead lies in $[-R,R]$ and its distinguishable values are separated by at least $\delta$, at most
$(1+2R/\delta)^m$ distinguishable coordinate combinations are available. Recovering $N$ distinct
messages then requires $\log_2N\le m\log_2(1+2R/\delta)$.
\status{PROVED (counting)}

Pairwise direction packing measures a different quantity. For fixed coherence tolerance,
exponentially many nearly orthogonal unit directions can fit in $\R^m$, as constructions based
on the Johnson--Lindenstrauss lemma show \citep{johnson1984extensions}. Pairwise separation
does not by itself guarantee recovery of sums of those directions.

Recovery depends on which sums are allowed. For fixed sparsity and error tolerance,
\citet{garg2026howmany} permit exponentially many candidate coordinates with one fixed linear
readout, at width $O_\varepsilon(k^2\log d)$. This sparse-input guarantee differs from the
isotropic coefficient-risk bounds of \cref{sec:geo-budget}. The comparisons in
\cref{tab:capacity} therefore require the input support, amplitude range, accuracy criterion,
and decoder to be specified along with the number of directions.

\subsection{Observed Geometries}
\label{sec:geo-shapes}

The geometric bounds constrain possible encoders but do not predict which arrangement training
selects. The report by \citet{elhage2022toy} observes antipodal pairs, triangles,
pentagons, tetrahedra, and square antiprisms in trained toy models. Several arrangements occur
in mutually orthogonal blocks and have the fractional capacities listed above. Their resemblance
to solutions of Thomson-type packing problems motivates a geometric interpretation, although
the report treats the detailed polytope taxonomy as tentative.
\status{NUMERICALLY-OBSERVED}

The arrangements change with the model. The quadratic objective of
\citet{scherlis2022capacity} favors a large superposed block in cases where the ReLU model
forms smaller polytopes. Correlated data can instead produce clusters and cycles
\citep{prieto2026correlations}. These differences show why an observed shape needs to be
interpreted together with the training objective and feature law.

Evidence from trained language models also includes structures beyond individual linear
directions. Some quantities vary along curved manifolds \citep{gurnee2026manifolds}, and some
studied concept representations span multiple dimensions \citep{engels2025notall}. These
observations limit the scope of a one-direction-per-feature description without determining how
prevalent each alternative is. The next section asks when training produces a superposed
representation and what guarantees permit its features to be recovered.

\takeaway{sec:geometry}{A geometric overlap bound becomes a reconstruction bound after the
feature law and readout are fixed. Comparisons between tied linear, untied linear, and nonlinear
decoders should retain these assumptions. Sparse recovery also requires the allowed support
patterns, since well-separated directions can still produce collisions. Learned geometry adds
a further question: the loss and feature statistics determine which of the admissible
arrangements training favors.}

\section{Learning, Access, and Identifiability}
\label{sec:access}

The frame bound in \cref{sec:geometry} limits how far unit-norm directions can be separated. A trained network must still choose which features to represent and how to arrange them. Once that arrangement is fixed, a decoder must recover feature values from their shared coordinates. An observer who lacks the encoder faces a further problem: recovering the directions themselves. These three questions require different assumptions, even when they concern the same activation vector.

\begin{tcolorbox}[summarybox,title={Summary: Training, Access, and Identification}]
\roadmap
\begin{itemize}
\item Toy models link dimension sharing to sparsity, importance, and fourth moments under specified objectives (\cref{sec:acc-when}).
\item Initialization and regularization can produce polysemantic neurons even when every feature could have its own dimension (\cref{sec:acc-incidental}).
\item Uniform recovery by one linear map requires nearly quadratic dependence on the number of active features; sparse optimization can reduce that dependence under suitable encoder conditions (\cref{sec:acc-linear,sec:acc-gap}).
\item Recovering feature values through a known encoder and identifying an unknown encoder are separate problems (\cref{sec:acc-ident}).
\item A geometric argument predicts $1/m$ loss scaling in a toy-model regime. Fits to language models support an empirical connection whose mechanism remains to be established (\cref{sec:acc-scaling}).
\end{itemize}
\end{tcolorbox}

\subsection{Emergence of Superposition in Training}
\label{sec:acc-when}

\paragraph{Phase diagrams.}
\citet{elhage2022toy} train the tied ReLU autoencoder of \cref{eq:reluae} while varying feature sparsity and importance. The trained solutions allocate a dedicated dimension to some features, represent others along shared directions, and drop the rest. Changing the feature distribution changes these allocations because the loss balances the cost of dropping a feature against the interference created by retaining it.

The phase diagrams describe solutions reached by training. They do not directly evaluate the optimized comparison $\Sup$ in \cref{def:benefit}, which requires optimizing both the unrestricted and orthogonal model classes. The analytical calculations in \citet{elhage2022toy} compare selected candidate solutions in small cases. For the bias-free model with two features in one hidden dimension ($d=2$, $m=1$), Tom McGrath's accompanying comment reports an exact expected-loss calculation and an additional ``confused feature'' regime. At low sparsity, when both features are important, their weights can align near $w_1=w_2=1/\sqrt{2}$ even without correlated inputs \citep[Comments and Replications]{elhage2022toy}. The observed boundaries therefore need to be read with their feature law, activation function, and objective. \status{NUMERICALLY-OBSERVED}

\paragraph{A fourth-moment threshold.}
A quadratic toy model permits a more explicit account of allocation. \citet{scherlis2022capacity} study independent, centered features with unit variance and show how their fourth moments affect the optimal capacities of \cref{res:scherlis}. Above the Gaussian kurtosis $\E[f_i^4]/\E[f_i^2]^2=3$, an intermediate range of feature importances can receive fractional capacity. Below that threshold, the optimal allocation is all-or-nothing in the analyzed regime. Fractional capacity describes dimensions shared among features, so the result identifies conditions under which sharing minimizes this model's objective. \status{PROVED in the toy model}

This calculation concerns an optimum; the phase diagrams also depend on whether training reaches one. It explains one role of sparsity without making sparsity a universal criterion. Sparse distributions can have large normalized fourth moments, but the relevant moment, centering assumptions, and reconstruction objective must be specified before transferring the threshold to another model.

\subsection{Incidental Polysemanticity}
\label{sec:acc-incidental}

Dimension sharing can explain why a neuron responds to several features, but such responses also arise when width is ample. \citet{lecomte2024incidental} study models with $m\ge d$, where every feature could receive an orthogonal direction. Random initialization can place several features in the basin of the same neuron, and sparsity-inducing regularization or noise can preserve that arrangement. Their results establish this mechanism in stylized settings and observe it in trained models. \status{PROVED in stylized settings; NUMERICALLY-OBSERVED}

Polysemantic responses alone therefore leave the source of mixing unresolved. The geometric definition in \cref{def:superposition} additionally asks for nonorthogonal feature directions and recoverability under a stated decoder. Demonstrating a recoverable count above $m$ establishes the stronger claim that the representation accommodates more features than its width. A collection of neuron labels establishes neither count nor geometry, and spare width alone does not determine the optimized benefit $\Sup$.

\subsection{Linear Access}
\label{sec:acc-linear}

To quantify recovery independently of which arrangement training selects, consider designing an encoder together with one fixed linear decoder. The encoder maps each admissible feature vector to the hidden representation, and the decoder must reconstruct that vector without adapting its weights to the active features. Define $m_{\mathrm{lin}}(d,k,\varepsilon)$ as the least width for which there exist $W$ and $V$ satisfying $\norm{V^\top Wf-f}_\infty\le\varepsilon$ for every $k$-sparse $f\in[-1,1]^d$. This is a uniform approximation problem over a specified input set, rather than an accuracy guarantee for an arbitrary trained encoder.

\begin{result}[Width required for uniform linear access \citep{garg2026howmany}]
\label{res:garg}
\status{PROVED}
For every fixed $\varepsilon > 0$,
\[
  m_{\mathrm{lin}}(d, k, \varepsilon) \;=\; O\!\bigl(\varepsilon^{-2}\, k^2 \log d\bigr),
\]
and, for fixed $0<\varepsilon<\tfrac12$, $2\le k=O(\log d)$, and sufficiently large $d$,
\[
  m_{\mathrm{lin}}(d, k, \varepsilon)
  \;=\;
  \Omega_\varepsilon\!\Bigl(\tfrac{k^2}{\log k}\, \log\tfrac{d}{k}\Bigr).
\]
\end{result}

At fixed error tolerance, the bounds give nearly quadratic dependence on the number $k$ of simultaneously active features, up to logarithmic factors. At fixed $k$, one linear map can still read exponentially many candidate features as width grows. The obstruction concerns simultaneous activity and uniform error, rather than an unconditional polynomial limit on the number of linearly readable features.

Coherence provides a sufficient construction criterion in the tied case discussed in \cref{sec:geo-budget}. \citet{garg2026howmany} establish a lower bound that also covers independently chosen encoders and decoders. The lower bound is essential: a limitation of a coherence certificate by itself would not rule out a better construction.

\doesnot{\Cref{res:garg} concerns one fixed linear map, uniform $\ell_\infty$ error, and every bounded $k$-sparse sign pattern. Average accuracy under a particular feature distribution can behave differently. The lower bound requires its stated error regime and logarithmic upper scale for $k$, and does not apply to a decoder with additional nonlinear computation.}

\subsection{Comparison of Decoder Classes}
\label{sec:acc-gap}

The preceding width bounds restrict recovery to one fixed linear map. A sparse-optimization decoder instead uses the sparsity of each input by solving for its active coefficients. Under suitable restricted-isometry conditions on $W$, basis pursuit recovers every $k$-sparse $f\in\R^d$ exactly from the hidden vector $h=Wf$ using
\[
  m=O\!\bigl(k\log(d/k)\bigr)
\]
measurements; suitable random matrices achieve these conditions \citep{candes2006stable,donoho2003optimally}. Compared with \cref{res:garg}, its dependence on $k$ improves from nearly quadratic to linear, with different logarithmic factors and an exact-recovery target. At $k=1$, both displayed upper bounds are logarithmic in $d$ at fixed $\varepsilon$.

These guarantees require properties of the encoder as well as the decoder. Angular separation alone does not suffice: the simplex in \cref{ex:simplex} has optimally spread directions but satisfies $w_1+w_2+w_3=0$. If both the all-active and empty patterns are admissible, they produce the same observation and defeat every decoder. Conversely, the limitations of a coherence-based proof do not limit all recovery methods. When $d\gg m$, the Welch bound restricts certificates based on $k\mu\lesssim1$ to sparsities of order $\sqrt m$; stronger properties such as restricted isometry can certify larger sparsities.

\Cref{tab:access} separates these statements by their recovery target. Transferring a sparse-recovery guarantee to a trained network requires checking the needed encoder property. Randomness is one way to obtain that property, not a necessary description of every successful learned dictionary.

\begin{table}[tbp]
\centering\small
\renewcommand{\arraystretch}{1.35}
\caption{Recovery guarantees for different decoder families. The first and third rows compare uniform sparse recovery with different error criteria and computational procedures. The remaining rows state model-dependent limits rather than intermediate width bounds.}
\label{tab:access}
\begin{tabularx}{\textwidth}{@{}>{\raggedright\arraybackslash}p{0.26\textwidth} >{\raggedright\arraybackslash}p{0.22\textwidth} X@{}}
\toprule
Decoder class & Guarantee considered & Conditions and limits \\
\midrule
One linear map, $\Psi_{\mathrm{lin}}$ &
$m=\widetilde\Theta_\varepsilon(k^2)$ \citep{garg2026howmany} &
Uniform $\ell_\infty$ approximation on $k$-sparse inputs in $[-1,1]^d$. The logarithmic dependence on $d$ and $k$ is explicit in \cref{res:garg}; its lower bound requires fixed $0<\varepsilon<\tfrac12$, $2\le k=O(\log d)$, and sufficiently large $d$. \\
Linear map with pointwise nonlinearity, $\Psi_{\mathrm{lin}+\sigma}$ &
Depends on the activation and input law &
ReLU improves recovery in \cref{ex:antipodal}; that example does not give a width bound for every signed sparse input. \\
Sparse optimization, $\Psi_{\mathrm{sparse}}$ &
$m=O(k\log(d/k))$ \citep{candes2006stable} &
Uniform exact recovery under suitable restricted-isometry conditions. Coherence alone gives a more restrictive sufficient certificate. \\
Unrestricted decoder, $\Psi_{\mathrm{all}}$ &
Depends on the information retained by the encoder &
Admissible patterns mapped to the same observation cannot be distinguished, regardless of decoder complexity. \\
\bottomrule
\end{tabularx}

\end{table}

\subsection{Identifiability}
\label{sec:acc-ident}

The preceding results assume a known or jointly designed encoder. Dictionary learning starts with observations and seeks the feature directions themselves. Recovering those directions, up to the symmetries allowed by the model, requires distinguishing among different latent explanations of the same data. Accurate reconstruction alone cannot make that distinction.

Positive identification results distinguish the latent directions through explicit assumptions or additional access to the system. For example, \citet{liu2026extracting} recover the directions $v_i$ in a sum of ridge functions, $\sum_i\sigma_i(\ip{v_i}{x})$, using noisy query access under their nondegeneracy conditions. The directions can be overcomplete and the response functions need not share a prescribed form. This query model lets the algorithm choose inputs $x$ and observe noisy function values, whereas an activation dictionary usually receives sampled hidden states. Within that activation setting, \citet{chen2025taming} give feature-recovery guarantees for an SAE variant under an activation generative model. Each result establishes identification for its observation model and assumptions.

Negative results explain why a reconstruction objective need not identify the intended features. In the regimes analyzed by \citet{cui2025limits}, closed-form SAE optima recover ground-truth features only under stringent sparsity conditions. \citet{tang2026unified} formulate sparse dictionary learning, including SAE, transcoder, and crosscoder objectives, as a piecewise biconvex problem and analyze its solution structure. Their characterization exposes non-identifiability and objective-induced pathologies, including mechanisms relevant to absorption and dead latents. These results locate failures within specified models; they do not establish that every trained dictionary must exhibit each failure. \status{PROVED as stated in the cited preprints}

Applying a positive theorem to model activations requires its assumptions to hold, or a robustness result that covers their violation. A recognizable latent label does not supply that evidence. Known-feature benchmarks can test recovery directly, while unknown-feature evaluations must assess the available proxies and their limitations. \Cref{sec:recovery,sec:protocol} develop this distinction for practical dictionaries.

\subsection{Superposition and Neural Scaling Laws}
\label{sec:acc-scaling}

The recovery and identification results above ask which feature values and directions can be recovered. A separate question concerns how the loss of the representation selected by training changes with width. \citet{liu2025scaling} argue that loss falls as $1/m$ in a strong-superposition regime of their toy model. Their geometric argument assumes approximately isotropic feature directions, whose squared overlaps have mean $1/m$, and relates those overlaps to reconstruction loss. Simulations support the predicted exponent in the studied regime. When activation probabilities instead follow a sufficiently steep power law $p_i\propto i^{-\alpha}$, the observed geometry changes and the reported exponent approaches $2(\alpha-1)$. The $1/m$ prediction therefore depends on the distributional regime as well as the geometric approximation.

To examine whether this width prediction extends beyond the toy model, the authors compare it with open-weight language models and report a fitted width exponent of $0.91\pm0.04$. They interpret tokens as atomic features and examine row norms and interference in the language-model head. Measured token frequencies have an exponent near $1$, within the regime they compare with the toy model. These measurements support a particular interpretation of the head; they do not identify the mechanism of loss scaling throughout the network. A related question varies training time instead of width: \citet{chen2026powerlaw} derive power-law learning behavior from a superposition bottleneck within their model. The width and training-time results concern different limits and retain their respective assumptions.
\status{Geometric argument and NUMERICALLY-OBSERVED toy-model scaling; REPORTED-EMPIRICAL language-model fits}

An empirical match in exponents motivates testing whether interference predicts how loss changes when width or feature statistics change. Such tests would connect the proposed mechanism to the scaling observation. The reconstruction models considered here ask the network to preserve existing feature values. \Cref{sec:computation} asks what additional resources are needed to compute new features from those shared representations.

\takeaway{sec:access}{A recovery claim needs both an input model and a readout procedure. Uniform sparse recovery can be possible even when a particular linear readout fails, while successful reconstruction can leave the underlying directions unidentified. These distinctions guide the later method comparisons: an activation dictionary must be evaluated for the feature information it recovers, the assumptions that make its directions meaningful, and the behavior those directions can explain.}

\section{Computation in Superposition}
\label{sec:computation}

Recovering an input feature shows that its value survived compression. Computing a new feature requires a layer to transform the compressed values while controlling interference at its output. The distinction matters even for simple targets: a code that preserves sparse Boolean inputs does not by itself implement their pairwise ANDs. We examine the resources needed for such transformations, the conditions under which their outputs can feed further layers, and the evidence that training finds these mechanisms.

\begin{tcolorbox}[summarybox,title={Summary: Computation in Superposition}]
\roadmap
\begin{itemize}
\item For specified families of sparse Boolean tasks, parameter-description lower bounds separate representing outputs from computing the requested instance (\cref{sec:comp-separation}).
\item Explicit ReLU constructions compute pairwise ANDs with shared neurons. Their widths depend on the input encoding, sparsity, and composition requirements (\cref{sec:comp-constructions}).
\item Approximate versus exact readout alone does not explain the differences between published bounds (\cref{tab:compbounds}).
\item Small trained models provide mechanism evidence, but their solutions depend on the architecture and loss, and decodability alone does not establish downstream use (\cref{sec:comp-dispute,sec:comp-gap}).
\item Delayed recall extends representation sharing across feature ages. It preserves past values rather than computing new Boolean features (\cref{sec:comp-temporal}).
\end{itemize}
\end{tcolorbox}

\subsection{From Recovery to Computation}
\label{sec:comp-questions}

The absolute-value experiment in \citet{elhage2022toy} gives a small trained example of the additional demand. An untied network with $m<d$ hidden dimensions learns to output $|f_i|$ for sparse signed features. A separate circuit would use two ReLUs per feature, one for each sign. The trained sparse solution shares neurons among features, and the authors trace how several neurons jointly compute one output.

One traced motif controls interference asymmetrically. A feature can excite an unrelated output, while another neuron inhibits that output where the unwanted positive contribution occurs. The output ReLU removes the resulting negative contribution. This observation connects the clipping mechanism of \cref{sec:geo-escapes} to a transformation of the input, although it establishes the mechanism only in the inspected toy setting. \status{NUMERICALLY-OBSERVED}

A general account must address three further questions. How does the required width grow with the target functions? Can the outputs remain compressed and support repeated composition? Under what objectives does training find the required circuits? The formal constructions below address the first two; the later training studies supply evidence for the third in specific models.

\subsection{The Representing-versus-Computing Separation}
\label{sec:comp-separation}

\citet{adler2024complexity} compare representation and computation for families of sparse Boolean tasks. The input is $f\in\{0,1\}^d$ with $\norm{f}_0\le k$. An instance of $\textsc{2-and}_{d,d'}$ specifies an ordered list of $d'$ pairwise AND outputs; another family specifies permutations of the inputs. Both input and output features are encoded in shared coordinates.

Choosing the instance is part of the computation problem. A universal architecture receives a parameter string that determines which outputs it must compute, and the lower bound counts the bits needed to specify that behavior. The upper bound constructs a ReLU network whose output is decoded by a fixed linear map followed by thresholding at $1/2$. This thresholded readout recovers the Boolean targets exactly.

\begin{result}[Separation of representing and computing \citep{adler2024complexity}]
\label{res:separation}
\status{PROVED}
\begin{enumerate}
\item \textup{(Representing.)} For a fixed bound on the number of simultaneously active features, storing which of $d'$ Boolean features are active, each
  recoverable to fixed tolerance by its own linear functional, requires only $m = O(\log d')$
  dimensions, by the Johnson--Lindenstrauss lemma \citep{johnson1984extensions}; equivalently, a
  width-$m$ layer can \emph{represent} up to $2^{O(m)}$ features.
\item \textup{(Computing, lower bound.)} Fix $d'\le\binom d2$ distinct input pairs in a set
$S$, and let $U=\{e_i+e_j:(i,j)\in S\}$ be the corresponding two-hot inputs. Consider the
family of all orderings of these $d'$ AND outputs. A single universal architecture, with its
behavior specified by a parameter string, must realize every member of this family correctly
on at least a $(1-\delta)$-fraction of $U$, for fixed $0\le\delta<1/2$.
This requires $\Omega(d'\log d')$ parameter bits in the worst case. The source establishes
the lower bound for almost all members of a constructed robust subset of the family, whose
members differ on more than a $2\delta$ fraction of $U$. An analogous argument applies to
permutations with one-hot inputs. For networks with a constant number of $m\times m$ layers
and $O(1)$ bits per parameter, the family-level requirement forces
$m=\Omega(\sqrt{d'\log d'})$.
\item \textup{(Computing, upper bound.)} An explicit ReLU network with
  $m = O(\sqrt{d'} \log d')$ neurons, $O(d' \log^2 d')$ parameters, and $O(1)$ layers computes
  every instance of $\textsc{2-and}_{d,d'}$ and every permutation \emph{exactly}, with inputs and
  outputs both in superposition, for $k\le2$; instances can be chained to arbitrary depth
  without error accumulation. This is within a $\sqrt{\log d'}$ factor of the lower bound.
\item \textup{(Capacity.)} Under this fixed-depth, fixed-precision accounting, a width-$m$
architecture that can realize the specified family must have $d'=O(m^2/\log m)$. This limits
family expressivity; individual specialized functions can have smaller implementations.
Sparse feature representation alone can accommodate exponentially many candidate features.
\end{enumerate}
\end{result}

\doesnot{The error fraction in item 2 is measured on $U$, not on every two-hot input.
When $d'\ll\binom d2$, an all-zero predictor can succeed on most of the latter inputs.
The almost-all statement concerns the constructed robust subset, rather than the full function
family. The bound does not constrain one specialized function. General sparsity costs the upper
construction a factor $O(k2^k\log d)$; the width lower bound also requires its depth and
precision assumptions.}

The separation shows that a short representation of an output vector need not be a short description of the function producing it. Within the specified families, parameter count is therefore a more informative computational resource than activation width alone. \citet{adler2024complexity} also interpret the result as a limit on compression: retaining arbitrary instance behavior requires enough parameter bits. Applying that interpretation to distillation or pruning would require showing that the model's task and parameter accounting meet the lower bound's assumptions.

\paragraph{Comparing the Published Widths.}
For comparison with this family-level requirement, the all-pairs task of \citet{hanni2024computation} fixes the collection of $d'=\binom d2$ ANDs. Their construction with superposed inputs has width $\widetilde O(\varepsilon^{-2}\sqrt d)$, which is smaller than the width lower bound above when expressed in $d'$. The task quantifiers differ: a fixed all-pairs computation is not an arbitrary ordered instance chosen through a parameter string. A comparison must also count encoding and readout resources, precision, sparsity, and depth consistently.

Readout exactness alone cannot reconcile these rates. A linear readout with uniform Boolean error $\varepsilon<1/2$ gives exact answers after thresholding at $1/2$. In particular, the fixed-sparsity all-pairs construction guarantees one model for all admissible inputs. The deep-circuit result has a different input-coverage guarantee. \Cref{tab:compbounds} records these distinctions without imposing a common ordering on the bounds.

For repeated composition, \citet{adler2024complexity} compare another regime of the concurrent construction at width $\widetilde\Theta(d'^{2/3})$ and polynomial depth with their own $O(\sqrt{d'}\log d')$ constant-depth construction.

\begin{table}[tbp]
\centering\small
\renewcommand{\arraystretch}{1.35}
\caption{Computation bounds with their task and recovery conditions. In the all-pairs rows, $d'=\binom d2$. In the circuit row, $n$ denotes the emulated circuit width, and the source provides a proof sketch with draft supporting material. The rows quantify over different tasks and resources; their widths do not form a hierarchy of readout strength.}
\label{tab:compbounds}
\begin{tabularx}{\textwidth}{@{}>{\raggedright\arraybackslash}p{0.25\textwidth} >{\raggedright\arraybackslash}p{0.30\textwidth} >{\raggedright\arraybackslash}p{0.17\textwidth} >{\RaggedRight\arraybackslash}X@{}}
\toprule
Result & Readout and error model & Sparsity & Width (neurons) \\
\midrule
Representation only \citep{johnson1984extensions} &
Fixed linear readouts recover sparse feature values to fixed tolerance &
$k=O(1)$ & $m=O(\log d')$ \\
All pairwise ANDs, basis-aligned inputs \citep{hanni2024computation} &
One linear readout with uniform $\ell_\infty$ error $\varepsilon$; single layer &
Fixed $k$ & $m=O(\varepsilon^{-2}\log^8d)$ \\
All pairwise ANDs, superposed inputs \citep{hanni2024computation} &
Same uniform guarantee, with a constructed input encoding &
Fixed $k$ & $m=\widetilde O(\varepsilon^{-2}\sqrt d)$ \\
Sparse Boolean circuits \citep{hanni2024computation} &
Approximation with interleaved correction; all but a negligible fraction of admissible inputs &
$k$ bounds activity throughout the circuit &
$m=\widetilde O(n^{2/3}k^2)$ for circuit width $n$ and polynomial depth \\
Ordered AND instances or permutations, upper bound \citep{adler2024complexity} &
Exact thresholded recovery; shared input and output codes; constant depth per instance, chainable &
$k\le2$ & $m=O(\sqrt{d'}\log d')$ \\
Same families, lower bound \citep{adler2024complexity} &
Family-level requirement supported by a robust subset; error fraction $\delta<1/2$ is measured on the specified set $U$ &
$k=2$ & $\Omega(d'\log d')$ parameter bits; width follows under fixed-depth, fixed-precision accounting \\
\bottomrule
\end{tabularx}

\end{table}

\subsection{Constructive Results}
\label{sec:comp-constructions}

\paragraph{Universal AND.}
The all-pairs construction makes its neuron sharing explicit. \citet{hanni2024computation} connect each ReLU neuron to a random subset of Boolean inputs and give it bias $-1$. The neuron activates when at least two connected inputs are active. A readout for a particular pair averages neurons connected to both members of that pair.

Other active inputs can also make those neurons fire. Sparsity limits these coincidences, and concentration over the sampled connections bounds the remaining readout error. Each output AND is consequently represented across many neurons. The construction uses knowledge of the weights to choose its readouts, rather than assigning one neuron to each gate.

\begin{result}[Universal AND \citep{hanni2024computation}]
\label{res:uand}
\status{PROVED}
Fix a hard sparsity $k$ and tolerance $\varepsilon$. For large $d$ there exist single-layer ReLU
networks whose post-activations make every pairwise AND of $d$ Boolean features readable by one
linear map to $\ell_\infty$ error $\varepsilon$, uniformly over all $k$-sparse inputs:
with basis-aligned inputs, at width $m = O(\varepsilon^{-2} \log^8 d)$; with inputs
themselves presented in superposition, at width $m = \widetilde O(\varepsilon^{-2} \sqrt d)$.
\end{result}

\doesnot{\Cref{res:uand} establishes existence for fixed sparsity through designed encodings, weights, and readouts. It gives no training guarantee or guarantee for a probe learned from finite data. The inputs are Boolean and the bounds are asymptotic: for scale, $\log_2^8d$ is about $2.5\times10^{10}$ at $d=10^6$. The result therefore needs explicit finite-width analysis before yielding a practical resource estimate.}

Basis-aligned inputs expose each input bit directly to the layer. Superposed inputs require the layer to work through an approximate input recovery map, which changes the error calculation and width requirement. This distinction concerns the representation supplied to the network, even though both constructions produce the same set of output functions.

\paragraph{Error Correction and Depth.}
Repeated computation introduces another demand because readout errors can accumulate. \citet{hanni2024computation} first show why a weaker representation criterion is insufficient: two weakly linearly represented features can have an AND that no MLP layer makes linearly separable. Their positive construction instead controls approximation error and alternates computation with a ReLU correction layer. A random ternary code and a piecewise-linear rounding operation recover a cleaner encoding of the Boolean state.

The source states that a circuit of width $n$, polynomial depth $L$, and activity bounded by $k$ can be emulated at width $\widetilde O(n^{2/3}k^2)$ and depth $2L$, outside a negligible fraction of admissible inputs. At fixed $k$, this corresponds to circuit width $\widetilde O(m^{3/2})$ for network width $m$. The supplied argument is a proof sketch with draft supporting material. Its separate informal estimate of roughly $\widetilde O(m^2)$ computable features is not a theorem and should not be identified with the family-dependent capacity in \cref{res:separation}. \status{Source-reported proof sketch; informal estimate is CONJECTURE}

\citet{adler2024complexity} construct chainable layers through a different routing scheme. They divide inputs by how many outputs each influences. Low-influence inputs use dedicated superposed output channels, while high-influence inputs are handled through a shared input code. Exact recovery and re-encoding allow the constructed instances to compose without accumulated error. The resource bound applies to that complete construction, rather than measuring an isolated cost of exact thresholding.

\subsection{The Compressed-Computation Dispute}
\label{sec:comp-dispute}

A construction shows what weights can implement; a trained model requires evidence for the mechanism it actually learned. A sequence of studies on compressed elementwise ReLU computation illustrates the difference. The papers share a senior author, and the later studies revisit the architecture and loss used in the initial experiment.

\paragraph{The Initial Loss Advantage.}
\citet{braun2025apd} train a network with $m=50$ hidden ReLU neurons, a residual connection, and a fixed random embedding to output $\mathrm{ReLU}(f_i)$ for $d=100$ sparse signed features. Its loss improves on a baseline that serves 50 features perfectly and drops the rest. The authors interpret the improvement as evidence that the network computes more nonlinear feature functions than it has neurons.

\paragraph{Separating Linear Mixing From the Target.}
To examine what the loss advantage establishes, \citet{bhagat2026compressed} show that the residual and embedding make the task algebraically equivalent to fitting $y=\mathrm{ReLU}(f)+Mf$ with a plain one-layer MLP. The matrix $M$ records the embedding's Gram deviation from the identity. The target therefore contains a linear mixing component in addition to the intended elementwise nonlinearities. \status{PROVED}

Their experiments test whether this component accounts for the advantage. At $M=0$, the advantage disappears, including after fine-tuning a model originally trained with nonzero mixing. In a sweep of mixing scales $\sigma\in[0,0.08]$, the advantage peaks near $0.03$. The trained read and write directions also concentrate in the top-50 eigenspace of $M$. These observations connect the loss improvement to the mixing term. They are single-run curves without reported seeds or error bars, and a hand-built factorization baseline reproduces the qualitative profile without matching the full trained loss. The authors leave the complete mechanism unresolved. \status{REPORTED-EMPIRICAL}

\paragraph{Changing the Loss.}
To test how the objective affects computation on an unmixed target, \citet{ferreira2026l4} remove the residual and train on the uncontaminated elementwise ReLU target. Under squared error, the model again serves half the features and drops the rest, with per-feature error coefficient of variation $0.999$. Higher loss exponents produce more evenly distributed errors: across exponents $2.5$ through $8$, the reported coefficients of variation are $0.028$ to $0.046$, with variation at most $0.003$ across ten seeds.

The authors trace the fourth-power-loss solution to a sparse binary code. Each feature uses 5 to 7 of the 50 neurons, and each neuron serves 10 to 12 features. The decoder closely matches a scaled pseudoinverse of the encoder, with Frobenius cosine $0.996$. Transplanting one feature's activation pattern to another feature's code neurons moves the top decoded output to the second feature in every tested pair. A three-scalar weight approximation recovers the trained loss within a factor $1.13$; the naive, emulated-bias, and random-network baselines have losses $28$, $26$, and $56$ times the trained loss. \status{REPORTED-EMPIRICAL}

These interventions support the proposed code mechanism within the tested model. The resemblance to the sparse binary construction of \cref{res:uand} concerns a motif: the trained task uses continuous ReLUs, one layer, and no correction layers. Its outputs are also attenuated, with reported mean slope about $0.81$. The motif and loss advantage leave the uniform Boolean guarantee of the construction unestablished.

\paragraph{What the Comparison Establishes.}
Within this experiment, changing the loss changes how error is distributed across features and which solution training reaches. Reconstruction under squared error already produces sharing in the representational toy models of \cref{sec:acc-when}, whereas this computational task favors the naive solution under the same exponent. The comparison motivates studying the task, architecture, and objective jointly. It does not establish a universal separation between the training pressures needed for representation and computation.

The mixing control also shows why a baseline loss comparison needs a mechanism check. Per-feature errors, intervention tests, and an explicit account of the learned weights reveal distinctions hidden by aggregate loss. Further tasks and objectives are needed to determine how broadly the observed computational code is learned.

\subsection{The Existence-versus-Use Gap}
\label{sec:comp-gap}

The formal constructions provide no guarantee that gradient training finds their weights. The absolute-value model and the compressed-ReLU studies supply trained examples under particular objectives. \citet{adler2025combinatorial} provide another: in two-layer MLPs trained on sparse pairwise-AND formulas, they identify output-channel routing motifs from signs of the learned weights. Together these studies motivate a learning theory, while leaving its task and optimization conditions unresolved.

A readable computed feature also need not be used by the model's subsequent computation. \citet{hanni2024computation} show that suitable random one-layer MLPs already make all pairwise ANDs readable by a map constructed from the weight signs, with error $\widetilde O(1/\sqrt m)$. The corresponding width dependence is $\widetilde\Omega(\varepsilon^{-2})$ in their regime. Thus a successful AND probe can detect functions available in random activations. Establishing that a trained model relies on those functions requires evidence about its downstream behavior. \status{PROVED as stated in the source}

This distinction motivates the intervention tests in \cref{sec:protocol} and the learning questions in \cref{sec:open}. A useful explanation must connect the representation, the transformation producing it, and the subsequent computation that depends on it. A resource bound or a probe score establishes only the part of that connection it measures.

\subsection{Temporal Superposition}
\label{sec:comp-temporal}

Recurrence introduces a related resource question even when the task requires only recall. In a $\tau$-delay task, the target output is zero for $t\le\tau$ and reproduces the input at time $t-\tau$ for $t>\tau$. The state must preserve both feature identity and age until the requested output is due. \citet{sharma2026temporal} compare linear recurrent networks, linear recurrences with ReLU readout, and fully nonlinear RNNs on this task, with much of the geometric analysis at width $m=2$.

Starting from zero state, a linear recurrence gives
\[
 h_t=\sum_{s=0}^{t-1}\sum_i f_{i,t-s}w_i^{(s)},
\]
where $h_t$ is the recurrent hidden state, $f_{i,t-s}$ is the value of feature $i$ at time $t-s$, and $w_i^{(s)}$ is its input direction after $s$ applications of the recurrence. A model retaining all $d$ features until delay $\tau$ must account for the $d(\tau+1)$ feature-age pairs currently awaiting output. This is a count of the targets that must remain available; a trained model may instead drop some features. The delay task extends representation across time rather than computing new features in the sense of \cref{sec:comp-separation}.

\begin{result}[Loss decomposition for linear recurrent codes \citep{sharma2026temporal}]
\label{res:temporal}
\status{PROVED in the cited source}
For a single input feature, let $f_t=B_tU_t$, with $B_t\sim\mathrm{Bernoulli}(p)$,
identically distributed values $U_t$ of finite second moment, and all gates and values mutually
independent. A linear recurrence starts from zero and uses a bias-free linear readout $v$.
For the $\tau$-delay task, the source's Equation~(5) decomposes the expected sum of squared
errors into task, mean-correction, projection-interference, and composition-interference terms.
Appendix~B.1, Equations~(24) and~(25), gives the complete indexed identity.
Writing $\mu=\E[U_t]$ and $\nu=\E[U_t^2]$, projection interference sums
$(v^\top w^{(s)})^2$ over wrong ages $s\ne\tau$ with coefficient $p\nu$;
composition interference sums products of readout projections over ordered pairs of distinct
ages with coefficient
$p^2\mu^2$. At fixed weights, horizon, and value distribution, the latter has higher order in
$p$ as $p\to0$. When the input values have nonzero mean,
negative cross-products reduce this composition term, favoring opposite readout projections
as discussed by the source and related to antipodal placement in
\cref{sec:geo-escapes}.
\end{result}

\doesnot{\Cref{res:temporal} summarizes the scalar-input identity. The source gives the vector-input extension under independence across both features and time in Appendix~G.1, Equations~(84) and~(85). Correlated sequences introduce additional covariance terms. The ReLU-readout expression is a sparse-limit approximation for nonnegative values, and nonlinear recurrent geometries are studied separately.}

In the ReLU-readout approximation, a wrong-age direction with $v^\top w^{(s)}\le0$ contributes zero projection interference. This favors placing waiting ages in the half-space opposite the readout while aligning the due age with it. The statement concerns the $p\to0$ approximation; co-active inputs can matter at finite sparsity. Trained linear recurrences approximate this arrangement with spiraling age directions. Across the reported sparsity sweep, the best models' angles move from roughly $90^\circ$ to $270^\circ$. Nonlinear RNNs can additionally suppress old states through their recurrent ReLU. \status{Sparse-limit analysis and NUMERICALLY-OBSERVED geometry}

When more features compete for the same two-dimensional state, the studied solutions tend to maintain a feature across all required ages or drop it. The reported counts are five, three, two, and one maintained feature at delays $0,1,2$, and at least $3$, respectively. At zero delay the geometry recovers the static pentagon of \cref{sec:geo-shapes}. These are selected small-model solutions: across the reported sweeps, roughly $8\%$ to $54\%$ of trained models exhibit the analyzed geometries. Their prevalence and stability at larger widths remain separate empirical questions.

The feedforward and recurrent studies both show why a feature count must specify what the network is asked to do with each feature. \Cref{sec:relations} uses these distinctions to examine which broader observations superposition can explain and which assumptions practical recovery methods inherit.

\takeaway{sec:computation}{Computational guarantees depend on the target family and resource accounting, while trained mechanisms require evidence about the weights and their effects. The all-pairs constructions, the learned toy circuits, and the delay-task geometries answer different questions. In practical models, decoding a function is a starting point for testing its role; it does not establish that subsequent layers use it or that training found the mechanism of a formal construction.}

\section{Implications of Superposition}
\label{sec:relations}

The preceding sections establish how feature geometry, decoder choice, and the target computation
constrain recovery from shared representations. Applying these results to trained networks requires
distinguishing superposition from neighboring concepts and specifying the resource or behavior
being explained. We examine how these requirements limit explanations of scaling, adversarial
vulnerability, and representation similarity. A claim about resource savings needs the same care,
since narrower activations leave precision, parameter count, and arithmetic to be assessed separately.
Recovering the unknown features adds assumptions about their format, sparsity, and identifiability,
which determine the scope of the methods in \cref{sec:recovery}.

\begin{tcolorbox}[summarybox,title={Summary: Implications of Superposition}]
\roadmap
\begin{itemize}
\item \textbf{Related concepts measure different properties.} Polysemanticity concerns neuron
  responses; the linear representation hypothesis concerns feature format. Both leave above-width
  feature recoverability unestablished.
\item \textbf{Empirical explanations require additional evidence.} Scaling fits concern the
  language-model head, adversarial controls test specified bottlenecks, and similarity metrics
  can respond to geometry despite shared feature content.
\item \textbf{Efficiency depends on the resource.} Recovering sparse features in a narrow
  activation space leaves precision, readout parameters, and the cost of new computations to be
  assessed separately.
\item \textbf{Broader benefits remain conditional.} A teacher-student model shows a larger fitted loss-decay exponent,
  while general improvements in generalization, sample efficiency, or robustness remain
  unestablished.
\item \textbf{Feature recovery adds assumptions.} Sparse dictionaries target linear,
  one-dimensional, layer-local features, with sufficient sparsity and identifiability for the
  chosen recovery method.
\end{itemize}
\end{tcolorbox}

\subsection{Boundaries with Neighboring Concepts}
\label{sec:rel-boundaries}

\paragraph{Polysemanticity}
A polysemantic neuron responds to several unrelated input properties. Superposition can produce
such responses when multiple feature directions contribute to the same neuron, as demonstrated
in the toy-model report by \citet{elhage2022toy}. The converse does not follow. A rotation
of mutually orthogonal feature directions can make neurons polysemantic while preserving the
orthogonality of the directions. Polysemantic units also arise in models with spare capacity
(\cref{sec:acc-incidental}), without pressure to represent more features than dimensions.

Interpreting neuron responses introduces a further uncertainty. \citet{bolukbasi2021illusion}
find coherent patterns among top-activating examples for random directions about as often as for
trained neurons. \status{REPORTED-EMPIRICAL} Such patterns alone therefore give weak evidence
for a particular representational mechanism. \citet{bereska2024mireview} discuss basis rotation,
training-noise redundancy, and capacity-driven superposition as competing explanations, adopting
the last as a working hypothesis. Toy models establish a capacity trade-off associated with
superposition, but the relative contributions of these explanations to polysemanticity in a
trained model have not been measured.

\paragraph{Mixed Selectivity}
Where polysemanticity describes responses to several unrelated properties, mixed selectivity concerns neurons responding to combinations of task variables. Its proposed
advantage concerns the dimension of population activity across task conditions, rather than the
number of recoverable features in a fixed-width representation. \citet{rigotti2013mixed} show
how nonlinear mixing increases this activity dimension and supports linear classification.
Their prefrontal-neuron measurements reach the maximum estimated dimension after enlarging the
population by resampling; a simulated pure-selectivity population with matched noise remains
below that maximum.
\status{REPORTED-EMPIRICAL; the maximum is attained only on the enlarged population}

Recovering the constituent stimuli is a different task from classifying combinations of them.
\citet{orhan2015multiple} study this recovery through unbiased estimators and optimal linear
estimators. In their averaging-type mixing model, nonnegative symmetric mixing weights reduce
recovery accuracy under additive noise and a large-population approximation. With independent
neurons, the cost is characterized by overlap between population derivative profiles for the two
stimuli. Their numerical comparisons find a larger degradation from mixing than from halving
neural gain or doubling the Fano factor.
\status{PROVED (monotone degradation) under additive noise, large populations, and nonnegative symmetric mixing weights; the derivative-profile characterization additionally assumes independent neurons; the magnitude comparison is NUMERICALLY-OBSERVED}

The derivative-profile overlap resembles the cross-talk term in \cref{eq:interference}, but
the models impose different conditions. These population-coding analyses assign every task
variable a value on each trial and mix tuning curves over stimulus space. The sparse toy models
instead mix feature directions in activation space and use thresholding to suppress interference
(\cref{sec:geo-escapes}). A negatively biased ReLU is outside the unbiased and linear estimator
classes studied by \citet{orhan2015multiple}, whose nonnegative mixing constraint also differs
from unrestricted feature directions.

These differences allow increased activity dimension and reduced constituent-recovery accuracy
to coexist. \citet{orhan2015multiple} suggest in their discussion that mixing may suit a task
requiring a function of the stimuli rather than the stimuli separately. This is their
interpretation, not a derived recovery benefit. The appropriate comparison with superposition
therefore specifies the feature statistics, permitted readout, and target computation, rather
than treating every neuron with multiple responses as the same phenomenon.

\paragraph{The Linear Representation Hypothesis}
Polysemanticity and mixed selectivity describe neuron responses, whereas the linear representation hypothesis (LRH) concerns the format in which a concept is represented.
Under \cref{def:lrh}, a direction carries the concept and a coefficient gives its intensity
\citep{park2024linear}. Superposition in this survey presupposes that format for the features
under study, then asks whether overlapping directions permit their recovery. The packing degree
separately measures how many such features are recoverable per dimension. The claims are
compatible but are not logically independent.

The LRH permits $d\le m$ concepts with mutually orthogonal directions, so it does not imply
nonorthogonality or above-width recoverability. A nonlinear concept falls outside the assumed
format rather than furnishing a converse example of superposition without the LRH. Accordingly,
evidence from probes or steering can support a linear representation without establishing
superposition. \citet{sharkey2025open} distinguish the weak claim that some concepts are linear
from the strong claim that every concept is linear, reporting counterexamples to the latter in
some models. Superposition claims about specified linear features need the weak claim, rather
than a statement covering every concept in a network.

\paragraph{Classical Sparse Coding}
For representations in this linear format, sparse coding models each datum as a sparse combination of dictionary atoms
\citep{olshausen1996}. Superposition applies this representation to network activations, with
the columns of $W$ carrying features and downstream layers recovering or computing with them.
The interpretation of a fitted dictionary consequently changes. Besides reconstructing data,
a dictionary fitted to activations is used to estimate an unknown feature representation inside
a trained model. This additional identification claim requires the recovery conditions examined
in \cref{sec:acc-ident}.

An earlier application already connects sparse coding to internal representations.
\citet{arora2018senses} prove under a generative text model that a polysemous word embedding is
approximately a frequency-weighted sum of sense vectors, which they recover by sparse coding.
The decomposition concerns raw word vectors; estimating a sense vector from its contexts also
requires a covariance-derived linear map, which the authors find differs from the identity in
practice. \status{PROVED under their generative model} The toy-model literature examined in
\cref{sec:geometry,sec:access,sec:computation} develops related questions about interference,
when sharing lowers risk, and the cost of computing new features from shared directions.

\subsection{Superposition as Explanation}
\label{sec:rel-explanation}

These distinctions separate an observed pattern from a proposed explanation based on shared
feature directions. The remaining explanations concern model loss, adversarial behavior, and
similarity between representations. Their evidence ranges from results under known feature
models to measurements made through proxies in trained networks.

\paragraph{Neural Scaling Laws}
The scaling analysis in \cref{sec:acc-scaling} connects a toy model to loss-width measurements
on open language models \citep{liu2025scaling}. Its $1/m$ prediction uses an isotropy ansatz
for learned directions in the strong-superposition regime and is supported numerically in the
toy model. \status{NUMERICALLY-OBSERVED in the toy model, derived under an isotropy ansatz}
The trained-model fit treats tokens as atomic features, setting $d$ to vocabulary size, and uses
the language-model head to identify the proposed regime. Intermediate layers are not measured
by that analysis. \status{REPORTED-EMPIRICAL on open models} The fit supports an interpretation
of head behavior under those modeling choices, while the toy-model derivation does not become
a theorem about the trained network.

\paragraph{Adversarial Vulnerability}
The overlap entering the loss-scaling argument also provides a potential target for adversarial perturbations. Sparse activity can make typical interference small even when feature directions overlap.
In the toy model of \cref{sec:geo-risk,sec:geo-escapes}, a ReLU removes negative contributions
and its negative bias suppresses small positive contributions. An adversarial perturbation can
instead activate many overlapping directions with contributions of the same sign at a target
readout. Their sum can exceed the bias despite each perturbation being small
\citep{gorton2025adversarial}.

The toy-model report observes this effect using analytically constructed perturbations
rather than gradient search \citep{elhage2022toy}. With perturbation norm bounded by one tenth
of average input norm, vulnerability increases by more than $3\times$ as superposition forms.
\status{NUMERICALLY-OBSERVED} The construction isolates a mechanism in known feature space.
To connect that mechanism to attacks in input space, \citet{stevinson2026interference} analyze
a linear bottleneck followed by a classification head, whose readout directions are directly
available from the weights.

\begin{result}[Optimal perturbations align with interference \citep{stevinson2026interference}]
\label{res:advinterference}
\status{PROVED}
Let inputs $x \in \R^{d_{\mathrm{in}}}$ be embedded linearly as $h = Ex$ with
$E \in \R^{m \times d_{\mathrm{in}}}$, and let the scores of $d$ classes be read out as
$z_i = \ip{w_i}{h}$. The readout directions $w_i\in\R^m$ define the class scores. Fix $\eta>0$ and write
$g=E^\top(w_a-w_b)$. Among input perturbations with $\norm{\delta}_2\le\eta$:
\begin{enumerate}
\item if $g\ne0$, the perturbation maximizing the margin $z_a-z_b$ is
\[
  \delta^\star
  \;=\;
  \eta\, \frac{E^\top (w_a - w_b)}{\norm{E^\top (w_a - w_b)}_2}\,,
\]
the pullback to input space of the difference of the two readout directions. If $g=0$,
every admissible perturbation gives the same margin;
\item two models whose embeddings and readout directions differ by a single orthogonal transform
of $\R^m$ have the same $g$ and the same maximizing perturbations. Thus the attack depends on
the geometry of the code rather than on the basis in which the weights are written.
\end{enumerate}
\end{result}

\doesnot{The result concerns a linear embedding, readout, and pairwise margin. Its attack formula
alone does not establish superposition as the source of vulnerability. The paper's deep-network
analysis measures correlations between input-space attack profiles and head interference,
rather than applying the proposition to recovered intermediate features. The authors interpret
their controlled evidence as supporting sufficiency in the studied capacity-constrained settings,
while leaving necessity and models without a capacity constraint open.}

The synthetic experiments connect the optimal perturbation to interference between classes.
The trained embedding is observed to align each class's input directions with its readout
direction. Given that alignment, a coordinate's contribution to $\delta^\star$ is proportional
to the corresponding class direction's overlap with $w_a-w_b$. Directions associated with
classes other than the source and target can then participate in the attack.
\status{NUMERICALLY-OBSERVED premise; PROVED given it}

\citet{stevinson2026interference} test this interpretation with three synthetic controls.
Gradient attacks align with $\delta^\star$ at cosine $0.92$ to $0.97$. Transfer between
independently trained models tracks Gram-matrix similarity, rising from $5\%$ to $98\%$ as
input correlations constrain the geometry. At $m=d$, the learned directions become orthogonal
and successful attacks fall to zero in $1000$ attempts.
\status{NUMERICALLY-OBSERVED in the synthetic task} The last control changes the width relative
to the class count; it tests the proposed mechanism rather than supplying an orthogonal
representation for a fixed-width model with more features than dimensions.

Deep-network observations provide a more indirect comparison. On CIFAR-10 vision transformers,
\citet{stevinson2026interference} find stronger correlations between attack and interference
profiles as the bottleneck tightens. Corruption robustness follows clean accuracy, while
adversarial robustness remains low and approximately flat. \status{REPORTED-EMPIRICAL}
The different trends help distinguish adversarial behavior from a general loss of predictive
capacity, while retaining the head-based definition of features used in the measurement.

A broader interpretation is proposed by \citet{gorton2025adversarial}, whose workshop-track
preprint argues that superposition may be a primary cause of adversarial examples. In toy
models, their norm-based features-per-dimension proxy correlates with vulnerability at
$r\approx0.99$ and decreases under adversarial training. Their ResNet18 evidence uses a
different proxy: more robust models have lower SAE reconstruction error at fixed sparsity.
\status{NUMERICALLY-OBSERVED in toy models; REPORTED-EMPIRICAL via proxy} The untested confounds
of that recovery-based proxy limit the connection to feature count and interference.

\citet{bereska2025lossy} obtain a different response to adversarial training with an
entropy-based feature count. They exponentiate the Shannon entropy of activation magnitudes
across SAE latents and find that this effective count can increase as robustness improves.
\status{REPORTED-EMPIRICAL} Their interpretation depends on task complexity relative to network
capacity, rather than a universal reduction in feature count. The two proxy measurements
therefore leave a narrower conclusion intact: controlled examples link shared directions to
vulnerability, while the necessity or prevalence of that mechanism in general deep networks
remains unresolved.

\paragraph{Representation-Similarity Metrics}
Beyond explaining the behavior of one network, shared-direction geometry can affect comparisons between networks. Two networks may
carry the same feature vector through different encoders, $h=Wf$ and $h'=W'f$.
\citet{liu2026similarity} analyze representational similarity analysis, linear centered kernel
alignment, and linear regression on these activations. Under their asymptotic assumptions and
zero-mean, identity-covariance features, the metrics depend on $W^\top W$ and $W'^\top W'$.
Networks with identical feature content can consequently appear less similar than networks
sharing a subset of features. \status{PROVED under their assumptions}

This dependence motivates comparing recovered features instead of raw activations.
\citet{liu2026similarity} construct a supervised TopK autoencoder that inverts a
compressed-sensing problem made solvable by construction. Alignment is restored in its latents
while raw-activation alignment remains low. Unsupervised autoencoders and pretrained vision and
language autoencoders give the weaker empirical finding that latent alignment exceeds
raw-activation alignment. \status{REPORTED-EMPIRICAL} This comparison inherits the quality of
feature recovery. Higher latent alignment in a trained model does not by itself establish that
the recovered coordinates faithfully identify its features (\cref{sec:failures}).

\subsection{Superposition as Resource Allocation}
\label{sec:rel-allocation}

\paragraph{Activation Width and Precision}
Beyond these explanations of network behavior, efficiency claims require specifying which resource a representation saves. The recovery gains examined earlier concern the number of sparse features represented at a
specified activation width. Precision remains a separate constraint. A width-$m$ activation in
$[-R,R]^m$ read at granularity $\delta$ distinguishes at most $(1+2R/\delta)^m$ states under
the counting convention of \cref{sec:geo-precision}. Rearranging feature directions does not
increase that count.

Sparsity reduces the number of feature configurations that need to be distinguished. Among $d$
features with $k$ active, the support takes $\binom{d}{k}$ possible values, requiring on the
order of $k\log_2(d/k)$ bits to specify in the sparse regime. Active magnitudes require
additional precision. The compressed-sensing rate $m=O(k\log(d/k))$ in \cref{sec:acc-gap}
has the same combinatorial form for an appropriate dictionary and sparse-optimization decoder,
but counts dimensions rather than bits. This connection explains why sparse support matters
without equating activation width with unrestricted information capacity.

\paragraph{Allocation Across Features}
The capacity quantities in \cref{res:scherlis} describe how features share that width.
For nonzero columns, $\sum_i C_i\le\operatorname{rank}(W)\le m$; a dropped feature with
$w_i=0$ is outside this bound's stated hypothesis. In the quadratic toy model, fourth moments
above the Gaussian value allow an intermediate importance band with fractional capacity,
between dedicated features with $C_i=1$ and dropped features. Below the Gaussian value,
that intermediate band disappears \citep{scherlis2022capacity}.
\status{PROVED in the toy model} \Cref{sec:acc-when} develops this allocation result; its
implication here is that architecture width does not specify which features receive that width.

The entropy-based count of \citet{bereska2025lossy} attempts a related measurement in trained
networks, reporting effective features per neuron. Under adversarial training, their simple
tasks with ample capacity allow feature expansion, termed abundance, while complex tasks or
restricted capacity produce feature reduction, termed scarcity.
\status{REPORTED-EMPIRICAL} This operational count depends on fitted SAE latents and is
distinct from the toy-model capacities $C_i$. Interpreting it as allocation by the trained model
requires accounting for the recovery artifacts of \cref{sec:failures}.

\paragraph{Neurons and Arithmetic}
Activation width does not uniquely determine the arithmetic performed by an architecture.
The toy-model report observes that dense architectures evaluate every neuron, while
conditional computation can activate a subset \citep{elhage2022toy}. In the idealized limit
where cost counts active neurons alone, a polysemantic neuron could be replaced by dedicated
feature neurons at the same arithmetic cost. The report argues that eliminating interference
this way might make superposition less favorable, and predicts less superposition in
mixture-of-experts models. \status{CONJECTURE}

The report's preliminary comparison of loss reduction per activation frequency suggests
superposition asymptotically matching dedicated dimensions, but supplies neither a protocol nor
numerical results. We did not identify a direct empirical test of its mixture-of-experts
prediction. A related architectural intervention is studied by \citet{kong2025expand}, who
split neurons into sparser sub-neurons at fixed nonzero parameter count. On symbolic Boolean
tasks, this lowers polysemanticity metrics and improves accuracy, with the largest gains at
high polysemantic load. The same widening improves classifiers on CLIP embeddings,
convolutional networks, and deeper multilayer networks. \status{REPORTED-EMPIRICAL}

\citet{kong2025expand} attribute these gains to fewer feature collisions, while finding that
random splits approximate the improvement. Their intervention changes architecture and parameter
sparsity, rather than directly testing the mixture-of-experts prediction. Together, the proposal
and the measured intervention motivate distinguishing feature sparsity from the neuron sparsity
that determines which computations run on each input.

\paragraph{Width and Parameters}
A narrow representation can still require a large explicit readout. In \cref{res:garg}, width
$m=\widetilde\Theta_\varepsilon(k^2)$ permits one linear map to recover every $k$-sparse $f$
to $\ell_\infty$ error $\varepsilon$. The dependence on $d$ is logarithmic, supporting
$d=\exp(\Theta(m/k^2))$ feature readouts, but storing those readouts uses $md$ parameters.
Thus the representation width and the parameter count of the recovery map measure different
costs.

Computing new features introduces a further cost in the families of \cref{res:separation}.
A universal parameter-driven architecture that realizes every ordering of $d'$ selected AND
outputs to the specified accuracy on $U$ requires $\Omega(d'\log d')$ parameter bits in the
worst case. The same counting argument covers permutations. For the error-tolerant result,
$U$ and the robust subset of functions are those defined in \cref{res:separation}; neither
quantifier extends automatically to the full sparse-input or function family. Under a constant
number of $m\times m$ layers with $O(1)$ bits per parameter, the requirement implies
$m=\Omega(\sqrt{d'\log d'})$. It bounds family expressivity while leaving specialized
computations and other tasks to be analyzed separately.

\subsection{Scope of Further Benefit Claims}
\label{sec:rel-negative}

The resource distinctions above also limit broader efficiency claims. A lower reconstruction
risk at fixed width concerns the chosen feature law, decoder, and reference representation.
The optimized superposition benefit $\Sup$ in \cref{def:benefit} fixes those choices; an
illustrative fixed-decoder comparison or an observed training phase does not automatically
measure that optimized difference. Similarly, a loss advantage alone does not certify
computation in superposition, as the dispute in \cref{sec:comp-dispute} illustrates
\citep{bhagat2026compressed,ferreira2026l4}.

Training dynamics provide a separate, narrowly tested benefit. In a teacher-student model,
\citet{chen2026powerlaw} report a fitted power-law loss-decay exponent up to ten times the
sequential-learning value obtained without a superposition bottleneck. The bottleneck produces
a common fitted exponent across the tested settings, which their linear-regime analysis does
not explain.
\status{NUMERICALLY-OBSERVED in a teacher-student model; preprint} This observation concerns
training dynamics in the constructed model; comparable acceleration in trained language models
has not been established.

General improvements in generalization, sample efficiency, out-of-distribution behavior,
parameter efficiency, or inference-time arithmetic remain unsupported by a general result in
the surveyed literature. Robustness also depends on the setting and measurement, as the
adversarial studies above show. A benefit claim therefore needs its target performance measure
and constrained resource stated together with the feature distribution, tolerance, and decoder.
In a trained model, evaluating these choices starts with estimating the features themselves.

\subsection{Superposition as a Premise for Feature Recovery}
\label{sec:rel-foundation}

Recovering more than $m$ feature directions requires a decomposition with more than $m$
components. Conventional principal-component, factor, and independent-component analyses
return at most $m$ directions, motivating overcomplete dictionaries for superposed activations.
\citet{sharkey2025open} use this motivation within a decomposition, description, and validation
procedure, while \citet{shu2025saesurvey} review the resulting SAE methods and evaluations.
The additional dictionary width supplies candidate directions; it does not establish that they
are the trained model's features.

Establishing feature identity requires assumptions connecting the dictionary to its generating
representation. \citet{klindt2026unifying} relate this problem to sparse coding and
compressed-sensing inversion. They also note that the linear identifiability theorem they discuss
assumes equal latent and representation dimensions, excluding the overcomplete setting of
superposition. Recovery guarantees for the relevant feature law and inference method therefore
need to be checked separately, as in \cref{sec:acc-ident}.

These recovery conditions first require a choice of feature format: a sparse activation dictionary uses linear, one-dimensional atoms. A direction with a scalar
coefficient fits that format, while a curved feature need not. \citet{engels2025notall} report
day, month, and year representations with circular geometry in two-dimensional subspaces.
Patching these subspaces changes answers on two templatic tasks. This supplies causal-use evidence (L5 in
\cref{sec:def-ladder}) for the tested intervention, without identifying every underlying
feature. A circle can also arise from correlated one-dimensional features compressed into a
layer, leaving that interpretation contested (\cref{sec:geo-shapes}). The counting-task
report adds a curved low-dimensional representation that dictionary atoms discretize
\citep{gurnee2026manifolds}.

The remaining assumptions concern sparsity, identifiability, and layer coverage. A sparse
coefficient model needs a sparsity regime appropriate to its recovery conditions
\citep{cui2025limits}. Accurate recovered values do not by themselves identify the generating
directions (\cref{sec:acc-ident}), and a per-layer dictionary observes a feature's representation
at that layer rather than its possible development across layers. These requirements determine
which feature claims a fitted dictionary can support.

An SAE can still return a reconstruction when the target features violate these assumptions.
The resulting latents may split, absorb other properties, or vary across fits, among the
failures examined in \cref{sec:failures}. \citet{sharkey2025open} accordingly question whether
the assumptions behind sparse dictionary learning are fundamentally valid or pragmatically
useful. Testing them requires both a recovery method and an evaluation matched to its target.
We next examine how probes, dictionaries, functional replacements, and parameter decompositions
estimate features or computations from trained networks (\cref{sec:recovery}), before assessing
their evidence in \cref{sec:protocol}.

\takeaway{sec:relations}{Polysemantic responses and linear feature format leave superposition's
nonorthogonality and recoverability requirements untested. Shared directions can affect attacks and
similarity measurements, but trained-model conclusions depend on the measured features and
controls. Efficiency claims need a specified resource; narrow activations alone establish
neither cheap computation nor faster training. Recovering the unknown features adds assumptions
about their format, sparsity, identifiability, and layer coverage, which the practical methods
and their evaluations need to examine.}

\section{Recovering Superposed Features}
\label{sec:recovery}

The recovery assumptions in \cref{sec:rel-foundation} concern feature directions and values,
but these quantities are unknown in a trained network. Recovery methods estimate them from
activations or trained weights. We organize these methods by their target, since recovering a
specified property, describing an activation, and replacing a computation require different
objectives and evaluations.

Supervised probes test properties specified in advance, while unsupervised dictionaries estimate
directions whose sparse combinations reconstruct activations. Functional-replacement methods
extend this reconstruction objective to a sublayer's input-output function. Parameter-space
methods instead decompose trained weights into components that participate in different
computations. The activation site further limits each analysis, since a feature may develop
across layers. Comparing these method families therefore requires specifying both the target
and the states available for estimating it.

\begin{tcolorbox}[summarybox,title={Summary: Recovering Superposed Features}]
\roadmap
\begin{itemize}
\item \textbf{Supervised probes} test whether a named property is accessible from activations.
  Supervised dictionaries add task-specific causal tests when suitable counterfactuals are known.
\item \textbf{Sparse autoencoders} estimate a dictionary of directions without predefined labels.
  Their variants change amplitude estimation, sparsity control, hierarchical reconstruction, or
  the behavioral objective, with overlapping effects on the learned latents.
\item \textbf{Functional replacement} approximates sublayer computations with sparse surrogates.
  Attribution graphs retain reconstruction errors as nodes to account for omitted computation.
\item \textbf{Parameter decomposition} seeks sparsely used components of trained weights.
  Its empirical demonstrations remain smaller than those of activation dictionaries.
\item \textbf{Cross-layer methods} account for directions that persist or develop across layers.
  A dictionary fitted at one site measures the representation available at that site.

\end{itemize}
\end{tcolorbox}

\subsection{Supervised Probes and Mean-Difference Dictionaries}
\label{sec:rec-supervised}

\paragraph{Linear Probes}
A named property can be tested by predicting its value from the model's activations. Given labeled
pairs $(h(x),\rho_j(x))$, a linear probe fits an affine map from $h(x)$ to the chosen property
$\rho_j(x)$. This tests recoverability for that property within $\Psi_{\mathrm{lin}}$, as in
\cref{def:superposition}. The idealized case in which a linear map reads all features together
has the width requirements of \cref{sec:acc-linear}.

Successful prediction supports representation or accessibility (L1 or L2 in
\cref{sec:def-ladder}); causal use requires an intervention. This distinction matters because
randomly initialized networks already make broad families of input functions linearly readable
(\cref{sec:comp-gap}). A probe can therefore predict a property that the trained computation does
not use.

\paragraph{Sparse Probing}
A dense probe can combine correlates spread across all $m$ coordinates, leaving the property's
localization unresolved. To test localization in the neuron basis, \citet{gurnee2023haystack}
restrict the probe to at most $q$ coordinates and reduce $q$. Their experiments cover the Pythia
family and more than one hundred binary properties. In middle layers, individual neurons detect
contexts such as French text or program code, and ablations test their contribution to behavior.
Ablating one French-text neuron increases loss by $8\%$ in a 70M-parameter model and $0.2\%$ in a
6.9B-parameter model. The smaller ablation effect limits how much causal importance can be
inferred from single-neuron detection in the larger model.
\status{REPORTED-EMPIRICAL}

Early layers give a different form of localization. Individual neurons distinguish a target
compound word from bigrams sharing one of its tokens while also responding to unrelated n-grams.
For one compound, a sum of three neurons separates the target almost perfectly from the other
observed token combinations. The authors interpret this pattern as evidence of superposition in
a trained language model. \status{REPORTED-EMPIRICAL} They also distinguish this interpretation
from demonstrating more features than neurons, which their study does not attempt. The
co-occurrence test excludes a union of neuron-aligned features but leaves a conjunction of finer
features possible, so a recoverable count above the width remains untested.

\paragraph{Class-conditional Mean Dictionaries}
Prior circuit analysis can supply enough named properties to construct a dictionary rather than
one probe. For each value $v$ of a property $\rho_j$, \citet{makelov2025principled} use the offset
$\E[h\mid\rho_j=v]-\E[h]$ as its direction. An activation is then reconstructed by adding one
offset per attribute to the overall mean. On indirect-object identification (IOI) in GPT-2 small,
the task attributes are known from earlier circuit analysis, making their dictionary a supervised
reference for evaluating unsupervised recovery.

The reference passes three causal tests. Sufficiency tests whether reconstruction preserves task
behavior, while necessity tests whether removing the dictionary's component removes that
behavior relative to mean ablation. Sparse control tests whether editing two latents reproduces
the effect of patching a counterfactual activation. Task-specific TopK and Gated autoencoders
approach this supervised reference on the same tests, whereas autoencoders trained on the full
pretraining distribution perform worse on every test. \status{REPORTED-EMPIRICAL} The authors
attribute the gap to the closer match between IOI data and the task, but they also report a wider
hyperparameter search for the task-specific autoencoders. The more costly full-distribution runs
received less tuning, leaving that comparison partly confounded.

The supervised reference covers the attributes supplied in advance. The offsets measure differences in conditional means;
properties distinguished only by higher moments can be missed. Discovering features beyond that specified list requires
a different objective.

\subsection{Unsupervised Dictionary Learning}
\label{sec:rec-dictionaries}

\paragraph{The Sparse Autoencoder}
Unsupervised dictionary learning seeks directions without a predefined property list. A
superposed layer may carry $d>m$ features, so a decomposition limited to $m$ directions has too
few components to assign one to each feature. This motivates an overcomplete dictionary rather
than the at-most-$m$-component decompositions of principal-component, factor, or conventional
independent-component analysis (\cref{sec:rel-foundation}). A sparse autoencoder (SAE) learns
such a dictionary by reconstructing activations from a wider hidden layer with sparse activity
\citep{cunningham2024sae,bricken2023monosemanticity}. In our notation, it learns
\begin{equation}
\label{eq:sae}
  z(h) \;=\; \sigma\bigl(U h + b\bigr) \,\in\, \R^{\hat d},
  \qquad
  \hat h \;=\; D\, z(h) + c ,
\end{equation}
Here $h\in\R^m$ is an activation collected at the chosen layer and site. The learned map $U\in\R^{\hat d\times m}$ and bias $b\in\R^{\hat d}$ produce latent coefficients $z(h)$ through $\sigma$, while the unit-norm columns $D_i$ of $D\in\R^{m\times\hat d}$ supply reconstruction directions in the original activation space. The bias $c\in\R^m$ gives the reconstruction offset. Training combines $\lVert h-\hat h\rVert_2^2$ with a sparsity penalty or constraint
on $z$. The chosen dictionary width $\hat d\gg m$ is a hyperparameter, distinct from the unknown
number $d$ of features.

The two maps in \cref{eq:sae} have different relations to the theory. Dictionary columns $D_i$
are candidate estimates of feature directions $w_i$, whose identification (L3) requires the
conditions discussed in \cref{sec:acc-ident}. The coefficient map $h\mapsto z(h)$ performs
feature recovery and is therefore a decoder in the terminology of \cref{eq:decoderclasses}.
ReLU and JumpReLU place this map in $\Psi_{\mathrm{lin}+\sigma}$; TopK instead couples
coordinates by selecting the largest pre-activations and falls outside that coordinatewise
class \citep{gao2025scaling}.

The restricted coefficient map is intended to resemble operations available to the model.
\citet{rajamanoharan2024gated}, following the report by
\citet{bricken2023monosemanticity}, argue that a representation requiring an iterative sparse
solver might exceed what a linear map followed by a pointwise nonlinearity can use. They leave
this argument's empirical test open. We follow \citet{lieberum2024gemmascope} in calling the
coordinates $z_i$ \emph{latents}, since their identification with the model's features is what
recovery seeks to establish.

\paragraph{Objectives and Architectural Choices}
The ReLU encoder with an $\ell_1$ penalty produced interpretable latents in language models, but
its objective couples the decision to activate a latent with the magnitude used in reconstruction.
Subsequent designs change this coupling and the control of sparsity. Other designs require
reconstruction at several dictionary widths or optimize the effect on model behavior. These
choices interact, as the comparison in \cref{tab:saevariants} shows; the dedicated survey of
\citet{shu2025saesurvey} provides further architectural detail.

\paragraph[Gated Sparse Autoencoder (Gated SAE)]{Gated Sparse Autoencoder (Gated SAE) \citep{rajamanoharan2024gated}}
The $\ell_1$ penalty rewards smaller latent magnitudes even when their firing pattern stays
unchanged. Training can reduce that penalty by accepting additional reconstruction error, biasing
magnitudes downward. The Gated SAE separates the firing decision from magnitude estimation and
applies the penalty to the firing path. This removes aggregate reconstruction bias in the main
settings studied by \citet{rajamanoharan2024gated}.

The improvement extends beyond that bias correction. An ablation attributes a small part of the
fidelity gain to removing shrinkage alone, with most of the gain associated with better learned
directions. At many sites, compute-matched Gated SAEs use about half the $L_0$ at matched fidelity.
Their blinded human evaluation finds comparable interpretability to the baseline, so the
reconstruction improvement does not establish more interpretable latents.
\status{REPORTED-EMPIRICAL}

\paragraph[TopK Sparse Autoencoder with Auxiliary Loss (TopK)]{TopK Sparse Autoencoder with Auxiliary Loss (TopK) \citep{gao2025scaling}}
The firing-path penalty still controls sparsity indirectly. TopK replaces this penalty-based
selection with an order statistic, retaining the largest pre-activations on each example. The
active count becomes a direct training choice, while all coordinates participate in determining
which latents survive the threshold.

A separate training difficulty arises when latents stop firing and receive little opportunity to
learn. \citet{gao2025scaling} add an auxiliary loss that trains dead latents on the reconstruction
residual and initialize the encoder as the dictionary's transpose. Without mitigation, the dead
fraction reaches $90\%$ at large widths; the combined recipe holds it to $7\%$ at 16 million
latents on GPT-4 activations. \status{REPORTED-EMPIRICAL} This measures dictionary utilization,
while identifying the active latents with model features requires further evaluation.

\paragraph[BatchTopK Sparse Autoencoder (BatchTopK)]{BatchTopK Sparse Autoencoder (BatchTopK) \citep{bussmann2024batchtopk}}
A fixed active count allocates equal sparsity to examples that may contain different amounts of
structure. BatchTopK moves selection to the batch, controlling the average count while allowing
individual examples to use different numbers of latents. Its batch-level threshold retains the
order-statistic coupling of TopK but changes where the sparsity constraint applies.

\paragraph[JumpReLU Sparse Autoencoder (JumpReLU)]{JumpReLU Sparse Autoencoder (JumpReLU) \citep{rajamanoharan2024jumprelu}}
Coordinatewise thresholding permits variable activity without deriving each threshold from other
latents in an example or batch. JumpReLU uses
$\sigma_\theta(t)=\mathbf{1}[t>\theta]\,\mathrm{ReLU}(t)$, with a learned threshold for each
latent, and optimizes an $L_0$ penalty through straight-through estimators. A penalty coefficient
still tunes the resulting sparsity level \citep{rajamanoharan2024jumprelu}. This design also
connects to the Gated SAE: the weight-tied Gated encoder is a JumpReLU encoder with thresholds
determined by its biases and learned per-latent rescaling \citep{rajamanoharan2024gated}.
\status{PROVED}

\paragraph[Matryoshka Sparse Autoencoder (Matryoshka SAE)]{Matryoshka Sparse Autoencoder (Matryoshka SAE) \citep{bussmann2025matryoshka}}
Controlling sparsity also changes how related properties share latents. Representing a conjunction
with one latent costs fewer active latents than representing its two properties separately. A
specific latent can then replace a general latent in some contexts, producing absorption
\citep{chanin2025absorption}. Related effects split a general concept into narrower variants or
combine independently varying properties into one latent \citep{bussmann2025matryoshka}.

Matryoshka SAEs retain BatchTopK selection and train nested dictionary prefixes to reconstruct
independently. The early prefixes consequently receive a reconstruction objective that later
specialist latents cannot satisfy on their behalf. On Gemma-2-2B, this reduces first-letter
absorption from $0.49$ to $0.05$ at matched sparsity, with two percentage points less explained
variance \citep{bussmann2025matryoshka}. Matryoshka is also the exception to worsening absorption
with width in SAEBench \citep{karvonen2025saebench}. These improvements concern the tested
first-letter properties; the board-game experiments with known ground truth do not distinguish
Matryoshka from the other architectures \citep{bussmann2025matryoshka}.
\status{REPORTED-EMPIRICAL}

\paragraph[End-to-End Sparse Dictionary Learning (e2e SAE)]{End-to-End Sparse Dictionary Learning (e2e SAE) \citep{braun2024e2e}}
The preceding designs retain activation reconstruction as their target. Squared reconstruction
error weights a direction by its activation variance, which can differ from its effect on model
behavior. End-to-end dictionary learning instead replaces the activation at the chosen site by its SAE reconstruction and minimizes the KL divergence between the original model's output distribution and the distribution after that replacement, optionally with downstream
activation-reconstruction terms.

This behavioral objective changes which directions receive latents. In a case study, the local
SAE assigns $593$ latents to a cluster around a high-variance direction, while the end-to-end
variant with downstream reconstruction assigns two. Resample ablation of that direction after
the first token barely changes the output \citep{braun2024e2e}. The same study reports less than
$45\%$ of the baseline loss increase at matched $L_0$, but output-divergence training alone gives
different latents across seeds and different computation routes. Behavioral preservation thus
needs its own checks of latent stability and computational faithfulness, developed in
\cref{sec:protocol,sec:failures}.
\status{REPORTED-EMPIRICAL}

\begin{table}[tbp]
\centering\footnotesize
\renewcommand{\arraystretch}{1.25}
\caption{Sparse-autoencoder designs and their evaluated effects. The choices of coefficient
estimation, sparsity control, hierarchy, and training target can interact. Reported improvements
apply to the stated objectives and evaluations; identification of model features requires the
additional conditions of \cref{sec:acc-ident}.}
\label{tab:saevariants}
\begin{tabularx}{\textwidth}{@{}>{\raggedright\arraybackslash}p{0.135\textwidth}
  >{\raggedright\arraybackslash}p{0.215\textwidth}
  >{\raggedright\arraybackslash}p{0.17\textwidth}
  >{\raggedright\arraybackslash}p{0.21\textwidth} >{\RaggedRight\arraybackslash}X@{}}
\toprule
Variant & Change to \cref{eq:sae} & Difficulty addressed & Demonstrated effect & Scope \\
\midrule
ReLU, $\ell_1$ \citep{bricken2023monosemanticity, cunningham2024sae} &
$\sigma = \mathrm{ReLU}$; penalty $\norm{z}_1$ &
the original recipe &
interpretable latents on real models &
its own penalty produces the artifacts below \\
Gated \citep{rajamanoharan2024gated} &
separate firing and magnitude paths; $\ell_1$ on the firing path only &
shrinkage: the penalty also rewards scaling magnitudes down &
removes aggregate magnitude bias; improves the sparsity-fidelity trade-off in
compute-matched comparisons &
blinded human interpretability comparison is null; ranking against TopK is protocol-dependent \\
TopK with auxiliary loss \citep{gao2025scaling} &
keep the largest pre-activations; dead latents fit the residual &
dead latents (up to $90\%$ unmitigated); indirect sparsity control &
$7\%$ dead at $16$M latents; sparsity set directly &
couples latents through an order statistic; dense, principal-component-like solutions as the
active count approaches $m$ \\
JumpReLU \citep{rajamanoharan2024jumprelu, lieberum2024gemmascope} &
per-latent learned threshold; direct $L_0$ objective via straight-through estimators &
thresholding without batch coupling &
basis of the open Gemma Scope suite; a weight-tied Gated encoder has this form with bias-determined thresholds and per-latent rescaling &
the release itself reports no ground-truth or causal validation \\
BatchTopK; Matryoshka \citep{bussmann2024batchtopk, bussmann2025matryoshka} &
batch-level threshold; nested dictionary prefixes, each reconstructing alone &
absorption, splitting, composition &
reduces first-letter absorption at matched sparsity; avoids worsening absorption with width
in the SAEBench comparison &
the nesting is a researcher-imposed hierarchy prior; no differentiation on board-game models
with known ground truth \\
End-to-end \citep{braun2024e2e} &
reconstruction replaced by KL divergence of spliced outputs, optionally plus
downstream-activation error &
reconstruction rewards variance the model never uses &
reduces loss increase at matched $L_0$ and allocates fewer latents to a high-variance
direction whose resample ablation after the first token barely changes output &
the pure variant is seed-unstable and routes computation differently; latent magnitudes shrink \\
\bottomrule
\end{tabularx}

\end{table}

\paragraph{Identification and Scale}
Improved reconstruction and sparsity do not establish that a learned dictionary equals its
generating dictionary. The analyses in \cref{sec:acc-ident} derive recovery conditions for
specified objectives and distributions. Within their settings, \citet{tang2026unified} derive
absorption and dead latents from the geometry of the solution set, while \citet{cui2025limits}
find recovery restricted to a narrow sparsity regime. The former is a preprint and the latter is
refereed; their conditions do not supply a uniform identification guarantee for the architectures
in \cref{tab:saevariants}.

Empirical comparisons also depend on the training and evaluation protocol. Gated and TopK
comparisons differ in width matching, fidelity metric, and dead-latent mitigation, with conflicting
dead-latent observations \citep{rajamanoharan2024gated,gao2025scaling}. Gated and end-to-end
training each report roughly a factor-of-two sparsity improvement over a ReLU-$\ell_1$ baseline
through different modifications, but a matched comparison or a test of their combined gains is
absent \citep{rajamanoharan2024gated,braun2024e2e}. \Cref{sec:protocol} examines the controls,
metric reliability, and residual-error measurements needed to interpret these comparisons.

The scale of available dictionaries makes these evaluations practically consequential. Gemma
Scope supplies more than $400$ autoencoders across every layer and three sites per layer of Gemma
2 2B and 9B, plus selected layers of 27B \citep{lieberum2024gemmascope}. Other studies train 16
million latents on GPT-4 activations \citep{gao2025scaling}; a 34-million-latent
dictionary for Claude 3 Sonnet has roughly 12 million live latents
\citep{templeton2024scaling,gao2025scaling}. These widths specify fitted dictionaries, rather than
feature counts. The Gemma Scope release explicitly notes that many latents likely overlap and
that 30 million latents would not establish 30 million features.

\subsection{Functional-Replacement Methods}
\label{sec:rec-replacement}

\paragraph{From Reconstruction to Substitution}
Activation dictionaries estimate a representation, but tracing computation also requires the
transformations between representations. Through an MLP, latent-to-latent attribution depends
on the MLP nonlinearity and the input at which it is evaluated. A \emph{transcoder} approximates
the MLP's input-output function with a sparse model to make these connections simpler to compute
\citep{dunefsky2024transcoders}.

The transcoder changes the reconstruction target in \cref{eq:sae} from the input activation to
the sublayer's output. Its encoder row and dictionary column consequently occupy the input and
output spaces, respectively. This allows part of the direct residual-path attribution to be
computed from the surrogate weights, as expressed in the following result.

\begin{result}[Latent-to-latent attributions through transcoders factorize
\citep{dunefsky2024transcoders}]
\label{res:transcoderattr}
\status{PROVED}
Consider a model whose sublayers add their outputs to a shared residual stream, and replace each
MLP sublayer $g^{(\ell)}$ by a trained transcoder
$\mathrm{TC}^{(\ell)}(h) = D^{(\ell)}\,\mathrm{ReLU}\bigl(U^{(\ell)} h + b^{(\ell)}\bigr) +
c^{(\ell)}$. The decoder bias $c^{(\ell)}$ is constant and plays no role in what follows. Write
$z^{(\ell)}_i$ for the activation of latent $i$ at layer $\ell$, $D^{(\ell)}_i$ for its
dictionary column, and $u^{(\ell')}_{i'}$ for the encoder row of latent $i'$ at a later layer
$\ell' > \ell$. If latent $i'$ is active on the input at hand, the direct residual-path
contribution of latent $i$ to the pre-activation of latent $i'$ is
\[
  z^{(\ell)}_i \,\cdot\, \bigl\langle D^{(\ell)}_i,\, u^{(\ell')}_{i'} \bigr\rangle ,
\]
where the first factor depends on the input and the second is computable from the weights alone.
The factorization holds because the residual stream adds sublayer outputs linearly and the
transcoder is linear in its latents above threshold. For an autoencoder at the same site the
corresponding path passes through the MLP's own nonlinearity, and both factors depend on the
input.
\end{result}

\doesnot{The input-invariant factor is a property of the surrogate, and it equals a property of
the model only to the extent that $\mathrm{TC}^{(\ell)} \approx g^{(\ell)}$; no bound on the
approximation error is propagated into the attributions. The statement conditions on latent
$i'$ already being active and drops its threshold, so it does not establish that latent $i$
makes latent $i'$ fire. Attention patterns are treated as constants throughout, so how attention
was routed lies outside what the result covers.}

The factorization simplifies attribution within the surrogate, while its usefulness also depends
on approximation quality. Published sparsity-fidelity comparisons with same-site autoencoders
give different rankings.
\citet{dunefsky2024transcoders} report transcoders matching or beating SAEs on GPT-2-small and
two Pythia models, and they report the gap widening at scale. By contrast, the Gemma Scope
release reports transcoders strictly worse than MLP-output SAEs across six layers of Gemma 2 2B
\citep{lieberum2024gemmascope}. Tuning of the training recipe on one's own method is a confound
that \citet{dunefsky2024transcoders} declare on their side and that the Gemma Scope release
lists among its candidate explanations for the disagreement.
\status{REPORTED-EMPIRICAL, mutually confounded}

\paragraph{Circuits from Dictionaries}
To connect local attributions to task behavior, \citet{marks2025sparse} insert SAE latents into
the computation graph and estimate the effects of intervening on them. They estimate the indirect effect of every node on a task
metric in parallel by linearized interventions and threshold the effects into a sparse graph.
The graphs are then scored by faithfulness and completeness under mean ablation. The SAE reconstruction error enters as an additional node, retaining the contribution omitted
by the latents. On subject-verb
agreement tasks, circuits of roughly $100$ latent nodes in Pythia-70M explain a majority of task
behavior, while roughly $1{,}500$ neurons are needed to explain half of that behavior. However,
the circuits the authors actually interpret are smaller and sit at faithfulness $0.21$. Removing
the residual-stream error nodes severely disrupts the model, so the interpretable part of the
circuit depends on an uninterpreted remainder. \status{REPORTED-EMPIRICAL} The same latent
circuits support a demonstrated editing application, which \cref{sec:applications} takes up
with the other applications.

\paragraph{Dictionaries Across Layers}
A feature that persists across layers can be relearned separately by every per-layer dictionary.
To avoid this repetition, \citet{lindsey2024crosscoders} propose a crosscoder whose latents read
from and write to several layers. The acausal variant reads from and writes to every covered
layer. Weakly causal and strictly causal variants read at one layer and write to that layer and
the later ones. On an 18-layer model, the acausal crosscoder reaches a lower loss than per-layer
autoencoders at a matched total number of latents. At large compute, however, it needs about
twice the training FLOPs to reach the same loss.
\status{REPORTED-EMPIRICAL, institutional report}

The cross-layer transcoder combines surrogate computation with cross-layer dictionaries. Its
latents read the residual stream at one layer and reconstruct the MLP outputs at that layer and
every later one. A local replacement model built from these transcoders freezes the attention
patterns and the normalization denominators. It carries the approximation residual in an error
node at each token position and layer, and it supports \emph{attribution graphs} over the
latents. On a fixed prompt the replacement model reproduces the original model exactly, because
those error nodes absorb whatever the transcoders fail to capture \citep{ameisen2025circuit}.
The report compares these graphs with per-layer transcoders through replacement and completeness
scores, which \cref{sec:prot-darkmatter} defines and reports. Attention routing remains outside
the explanation because the replacement fixes the attention patterns, a limitation the report
states explicitly.
\status{REPORTED-EMPIRICAL, institutional report}

Representing computation through individual transcoder latents can produce a very fine
decomposition. In the addition example of \citet{lindsey2025molt}, each sum has its own
lookup-table latent. The authors instead propose a sparse mixture of linear transforms for
the computation. Each transform applies a low-rank linear map to the residual stream and is
active on a sparse set of inputs. These transforms have no feature directions of their own
and are intended to be paired with a separate representation decomposition. The authors
describe the method as preliminary and still under development.
\status{REPORTED-EMPIRICAL, institutional report, self-labelled preliminary}

\subsection{Parameter-Space Decomposition}
\label{sec:rec-parameter}

\paragraph{Motivation}
Activation-space replacement still relies on connections between recovered directions. Parameter
decomposition addresses a difficulty in interpreting those connections when directions overlap. The \emph{virtual weight} between
an upstream and a downstream feature is the scalar obtained by reading the weight matrix through
their two directions. The report by \citet{olah2025interference} argues that superposition contaminates
that scalar. Under superposition, other features' directions overlap with those two directions
(clause 1 of \cref{def:superposition}), so the scalar picks up a contribution from every
overlapping feature in proportion to the overlap. An observed weight-space connection between
superposed features then mixes the intended circuit connection with directional interference.

In toy models, \citet{olah2025interference} demonstrate this mixture and match the resulting
weight histograms to those reported for a one-layer language model by
\citet{bricken2023monosemanticity}. For one feature in that language model, the intended
connections and interference overlap in magnitude. A second toy model reproduces this overlap,
leaving no magnitude threshold that separates the two components. The authors label their
report low-confidence.
\status{NUMERICALLY-OBSERVED in toy models; institutional report, self-labelled low confidence}
This evidence motivates decomposing weights directly, although the separation between directional
interference and intended circuit connections remains a question for that decomposition.

\paragraph{Decompositions}
A direct parameter decomposition seeks components that together reproduce the trained weights
and contribute sparsely to individual inputs.
Attribution-based parameter decomposition (APD) seeks components that sum to the original
parameters (faithfulness) and are each of low rank (simplicity), with a few of them sufficing
on any one input (minimality). APD recovers ground-truth mechanisms in toy settings, and those
settings include features in superposition and cross-layer distributed representations
\citep{braun2025apd}.
\status{NUMERICALLY-OBSERVED in toy settings} Stochastic parameter decomposition (SPD) replaces
APD's attribution step with learned stochastic masks over rank-one subcomponents, which reduces
APD's cost and hyperparameter sensitivity. Each subcomponent lies inside one weight matrix, so
recovering a mechanism that spans several ranks or several layers requires grouping
subcomponents. The paper states that the groups are known in its toy settings and leaves an
explicit clustering algorithm to future work \citep{bushnaq2025spd}.
\status{NUMERICALLY-OBSERVED in toy settings; preprint}

Adversarial parameter decomposition (VPD) extends these methods by testing adversarially selected
ablations \citep{bushnaq2026vpd}. The method requires subcomponents marked unimportant to remain
ablatable in any combination, beyond removing all such components together.

In a four-layer language model of roughly $67$ million parameters, VPD decomposes about $28$
million non-embedding parameters into $38{,}912$ rank-one subcomponents. Across runs at $0.5$,
$1$, $2$ and $4$ times the subcomponent count of the main decomposition, the number of live
subcomponents stays flat at roughly $6{,}500$ to $7{,}000$. For the per-layer and cross-layer
transcoders in the same comparison, the number of live latents tracks the dictionary size at
both widths tested (\cref{sec:fail-splitting}). The authors interpret this stability as reduced
splitting and attribute it to the any-combination requirement. For each input and token position,
a subcomponent with importance $g$ can receive a weight mask anywhere in $[g,1]$, so a component
marked unimportant can still be retained. The authors argue that these masks expose narrow
components whose contributions outside their assigned contexts distort the output
\citep[Sections~2.4 and~3.5]{bushnaq2026vpd}.

\begin{samepage}
Beyond the stability of component counts, VPD extends the unit of analysis within attention layers: its subcomponents can span computations across several heads.
The report contrasts this capability with prior analyses that mainly take the head as the
unit of analysis.

\end{samepage}

The any-combination ablation requirement nevertheless remains only partially satisfied in the trained decomposition. After $320$
steps of adversarial optimization, there exist ablations of subcomponents marked unimportant
that change the model's output drastically \citep{bushnaq2026vpd}.
\status{REPORTED-EMPIRICAL, institutional report}

\paragraph{Scope of Parameter Decomposition}
These demonstrations reach about $10^8$ parameters, one to three orders of magnitude below the
Gemma 2 models with open dictionary suites. Their motivating separation of circuit logic from
directional interference has not been tested against ground truth with a known interference
component. The interference report also questions the assumption that mechanisms combine
linearly in weights and identifies APD as sharing that assumption \citep{olah2025interference}.
Parameter decomposition therefore provides a distinct object of analysis whose scalability and
mechanistic interpretation remain under evaluation. Activation-space methods continue to develop
alongside it, including the mixture of linear transforms in \citet{lindsey2025molt}.

\subsection{Superposition Across Layers}
\label{sec:rec-crosslayer}

\paragraph{The Layer as a Unit of Analysis}
The choice of an activation site or weight matrix also determines which part of a feature a
method can examine. In a residual stream, a written direction remains in the running sum until a
later sublayer adds a cancelling component. Per-layer dictionaries can consequently learn the same
persistent direction repeatedly, which motivates the crosscoders in
\cref{sec:rec-replacement} \citep{lindsey2024crosscoders}. The circuit-tracing report illustrates
this duplication with a seven-step chain that cross-layer latents replace at the first layer
\citep{ameisen2025circuit}. Within-layer sparse probing has the same scope restriction, noted by
\citet{gurnee2023haystack}.

Persistence differs from a feature being formed jointly across layers. The crosscoder
report finds latents with substantial decoder norms across most layers and directions that drift
between layers. The report regards this gradual formation as consistent with cross-layer
superposition, while also allowing a feature formed earlier and amplified later
\citep{lindsey2024crosscoders}. \status{REPORTED-EMPIRICAL, institutional report} Cross-layer
dictionaries and transcoders examine these relationships directly; APD and VPD also permit
parameter components spanning several matrices.

\paragraph{Consequences for the Width Bounds}
The bounds in \cref{sec:geo-budget,sec:geo-scherlis,sec:acc-linear} apply to a specified
representation of width $m$. At a single activation site, $m$ remains the observed width and the
bounds concern recovery from that site. A feature distributed across layers requires a separately
specified joint representation before a multi-layer dimension can replace $m$. We did not
identify a published capacity bound formulated for that cross-layer setting; \cref{sec:open}
returns to this question.

\subsection{Comparison of the Families}
\label{sec:rec-comparison}

\Cref{tab:families} compares the recovered objects and the evidence categories in
\cref{sec:def-ladder}. Supervised directions carry a specified meaning but cover the supplied
properties. Unsupervised dictionaries offer a broader activation decomposition while leaving
latent meanings to be evaluated. Combining the two supplies a reference for the specified
properties, as in the causal tests of \citet{makelov2025principled} and the probing analyses of
\citet{gurnee2023haystack}.

Functional replacement and parameter decomposition extend the target from representations to
computation. Their interpretation depends on how faithfully the surrogate or selected parameter
components preserve the computation being explained. The resulting directions, latents, and
components therefore need evaluations matched to their intended claims. We examine the available
references and controls in \cref{sec:protocol}.

\begin{table}[tbp]
\centering\small
\renewcommand{\arraystretch}{1.4}
\caption{Recovery families compared by target, recovered object, and evaluation needs. The
evidence categories in \cref{sec:def-ladder} identify what a test seeks to establish; belonging
to a method family does not automatically establish a category. \Cref{sec:protocol} examines
the required references and controls.}
\label{tab:families}
\begin{tabularx}{\textwidth}{@{}>{\raggedright\arraybackslash}p{0.14\textwidth}
  >{\raggedright\arraybackslash}p{0.19\textwidth}
  >{\raggedright\arraybackslash}p{0.19\textwidth}
  >{\raggedright\arraybackslash}p{0.20\textwidth} >{\RaggedRight\arraybackslash}X@{}}
\toprule
 & Supervised probes and dictionaries & Unsupervised dictionaries & Functional replacement &
Parameter decomposition \\
\midrule
Question answered &
is this named variable accessible, at which decoder class? &
what overcomplete coordinate system makes activations sparse? &
can an interpretable surrogate substitute for the computation? &
which parameter components participate in different computations? \\
Recovered object &
one direction (or offset set) per named variable &
dictionary of $\hat d$ directions with sparse latents &
surrogate sublayer maps; attribution graphs with error nodes &
sparsely used parameter components, built from rank-one subcomponents \\
Evidence sought &
accessibility (L2); causal use (L5) tested separately on templatic tasks &
identification (L3), conditional on a recovery reference or the assumptions of \cref{sec:acc-ident} &
substitutability (L4); mechanistic interpretation (L6) requires further tests &
mechanistic components (L6 target); evaluations in toy settings and one $\sim$$10^8$-parameter model \\
Characteristic artifacts &
the ontology is supplied in advance; dataset idiosyncrasies read as features &
shrinkage, dead latents, absorption, splitting, seed instability (\cref{sec:failures}) &
error nodes carry causal weight; attention routing outside the account; conclusions
surrogate-relative &
unaudited by third parties; interference-separation untested \\
Reported cost &
one supervised fit per variable &
hours on small models; a full open suite consumed over $20\%$ of GPT-3's training compute
\citep{lieberum2024gemmascope} &
surrogate training for the replaced sublayers; graph computation per prompt; cost depends on the replacement recipe &
reported only for models up to $\sim$$10^8$ parameters \\
Sources &
\citet{gurnee2023haystack, makelov2025principled} &
\citet{cunningham2024sae, gao2025scaling, bussmann2025matryoshka, braun2024e2e} &
\citet{dunefsky2024transcoders, marks2025sparse, ameisen2025circuit} &
\citet{braun2025apd, bushnaq2025spd, bushnaq2026vpd} \\
\bottomrule
\end{tabularx}

\end{table}

\takeaway{sec:recovery}{A probe tests accessibility of a specified property; an SAE estimates
candidate directions from sparse reconstruction. Changing sparsity or fidelity changes the
estimated dictionary without establishing feature identity. Transcoders and parameter
decompositions extend analysis to computation, where interpretation depends on the faithfulness
of the replacement or selected components. The activation site also limits the representation
being examined. These choices determine which reference and intervention can evaluate the
recovered object.}

\section{Empirical Protocol, Benchmarks, and Complexity}
\label{sec:protocol}

The methods in \cref{sec:recovery} estimate directions, latent coefficients, surrogate
computations, or parameter components. Evaluating these objects requires a reference for the
property being claimed. Synthetic generators supply known features, and restricted tasks supply
specified properties or counterfactual behavior. General language-model evaluations instead rely
on proxies for feature quality.

A repeatable score can still reflect properties of the evaluation rather than learned feature
structure. Negative controls test this possibility, while comparisons with known features assess
what the score measures. Reconstruction and behavioral evaluations answer different questions
about both recovered features and the residual left by a dictionary. The computational cost and
available data affect how readily these comparisons can be repeated.

\begin{tcolorbox}[summarybox,title={Summary: Empirical Protocol, Benchmarks, and Complexity}]
\roadmap
\begin{itemize}
\item \textbf{Known references} support direct tests of directions and firing patterns in
  synthetic data. Board-game models add trained representations with a specified, partial
  property list.
\item \textbf{Negative controls} test whether learned directions contribute to a score.
  Frozen-random dictionaries approach trained dictionaries on several evaluated proxies, but
  perform worse on task-specific sparse control.
\item \textbf{Metric audits} expose both reseed noise and disagreement with ground truth.
  Reconstruction, interpretation, and intervention scores rank dictionaries differently.
\item \textbf{Behavioral tests} examine effects that explained variance does not measure.
  Published examples separate high-variance directions from causally effective components.
\item \textbf{Residual measurements} quantify omitted activation variance, attribution paths,
  or explained prompts. Predictable residual structure and a fitted error asymptote concern
  different properties of the reconstruction.

\end{itemize}
\end{tcolorbox}

\subsection{Synthetic Ground Truth}
\label{sec:prot-synthetic}

\paragraph{Known Directions and Coefficients}
A direct test of feature recovery requires both the generating directions and their values on
each sample. A synthetic model provides these by fixing the generating directions $W$ and feature distribution $\mathcal D_f$, then sampling feature values and combining the directions with those values as coefficients, with an additive bias if the generator specifies one. The resulting
activations can be used to train a dictionary whose columns are compared with $w_i$ and whose
firing patterns are compared with the support of $f$
\citep{chanin2026synthsaebench,korznikov2026sanity}. Directional alignment and coefficient
recovery test different parts of the claimed representation.

These comparisons establish recovery for the chosen generator. Transferring the conclusion to a
trained network requires checking its feature law and representation, which are unknown.
\Cref{sec:acc-when} shows why this qualification matters: the encoder, feature distribution,
and training loss affect which features become superposed.

\paragraph{Generators with Different Recovery Demands}
Synthetic generators can incorporate properties that complicate recovery from model activations.
\citet{chanin2026synthsaebench} combine hierarchical features, correlations, and Zipfian firing
frequencies. Their benchmark uses $16{,}384$ ground-truth directions in $768$ dimensions and
trains dictionaries of $4{,}096$ latents, leaving dictionary width below the generating feature
count. The authors describe this as a lower-bound test of methods, since the linear
representation hypothesis holds by construction.

The generator reproduces a disconnect between reconstruction and latent quality, together with
an $L_0$-dependent precision-recall trade-off. It also explains why matching-pursuit encoders can
reconstruct well while scoring poorly on feature benchmarks. The encoders use interference
between ground-truth directions to improve reconstruction without recovering those directions.
\status{REPORTED-EMPIRICAL; workshop paper} These results motivate testing recovery even when
reconstruction is accurate.

\citet{korznikov2026sanity} remove the width mismatch by drawing $3{,}200$ directions uniformly
on the sphere in $\R^{100}$ and training $3{,}200$-latent dictionaries at matched sparsity.
BatchTopK and JumpReLU explain about $0.67$ of the variance under a constant firing rate, while
just $3$ directions have a dictionary column within cosine $0.8$. With heavy-tailed firing,
explained variance reaches $0.71$ and $7$ to $9\%$ of directions meet that matching threshold,
almost all among the most frequent features. Across the architectures tested in this setting,
recovery ranges from $7$ to $43\%$. \status{NUMERICALLY-OBSERVED; preprint} The measured gap
between reconstruction and recovery persists despite the matched width and sparsity. The
generator's independently drawn directions omit correlation and hierarchy, which the authors
identify as a limitation.

\paragraph{Matching and Sparsity}
A recovery percentage also depends on how estimated directions are matched to ground truth.
\citet{chanin2026reliable} use Hungarian matching and report mean cosines from $0.56$ to $0.79$
on the hierarchical generator. \citet{korznikov2026sanity} instead count directions whose
maximum cosine exceeds $0.8$, without a one-to-one matching constraint. Their appendix also
reports large architecture differences in the constant-rate setting, with $99.9\%$ recovery
for TopK and $0.03\%$ for Matryoshka. These percentages describe different generators,
architectures, and matching criteria, rather than a common recovery rate.

Matching directions leaves the accuracy of their coefficients to be tested separately.
\citet{chanin2026reliable} compare decoder geometry and latent firing against ground truth and
find that their optima occur at different $L_0$. Thus even a fixed generator can favor different
sparsity settings depending on which part of recovery is evaluated.

\subsection{Model Organisms}
\label{sec:prot-organisms}

\paragraph{Board-game Models}
Board-game models provide trained representations together with a formally specified state.
A transformer trained on chess or Othello move transcripts can encode board properties needed
for predicting legal moves. \citet{karvonen2024boardgame} test dictionary recovery of those
properties through coverage and board reconstruction. Coverage asks whether a thresholded latent
classifies each property. For each property, board reconstruction selects latents with training precision at least
$0.95$, then predicts that property on held-out games whenever any of its selected latents fires.

The best chess dictionaries achieve coverage $0.48$ and board reconstruction $0.85$. Dictionaries
trained on a randomly initialized transformer of the same architecture achieve $0.11$ and $0.01$,
while linear probes on the trained-model activations achieve $0.98$ and $0.98$.
\status{REPORTED-EMPIRICAL} This comparison detects learned board information and shows that the
dictionaries recover less of it than linear probes. The results also depend on the rebuilding
rule, which the authors warn can change the qualitative comparison.

\paragraph{Templatic Tasks with Imposed Counterfactuals}
A task with known counterfactuals supplies a reference for causal control rather than property
classification. On IOI in GPT-2 small, \citet{makelov2025principled} test whether at most $k$
latent edits reproduce the effect of patching an activation from a counterfactual prompt
(\cref{sec:rec-supervised}). The target is the behavioral effect of the patch; matching the
supervised directions is not part of this criterion. Successful control therefore leaves
feature identification as a separate question.

\paragraph{Representation and Surface-form Controls}
Interpreting low coverage requires checking whether the model represents the tested property.
A supplied property list can contain concepts that the model encodes poorly, as well as omit
features outside the list. In the chess study, linear probes fall below $F_1=0.95$ on $6$ of
$15$ strategy concepts, and dictionaries perform worst on those concepts as well
\citep{karvonen2024boardgame}. \status{REPORTED-EMPIRICAL} A probe reference helps separate
weak recovery from a weakly represented target.

A high score needs a complementary control for properties predictable from surface form.
The same study reports board reconstruction of $0.52$ at layer 1, where probing finds little
board information, when pieces on starting squares are included. Masking those squares lowers
the score to $0.01$. Probes on a randomly initialized model also reach $0.89$ for castling
rights, which correlate with move number and transcript syntax
\citep{karvonen2024boardgame}. \status{REPORTED-EMPIRICAL} These controls expose information
available without the intended board representation. We did not identify an equivalent masking
control in the language-model benchmarks discussed next.

\subsection{Benchmarks on Trained Language Models}
\label{sec:prot-benchmarks}

\paragraph{A Common Comparison Protocol}
General language models lack the specified state available in board games, so their benchmarks
compare proxies for recovery quality. SAEBench fixes a training protocol across more than $200$
dictionaries on Gemma-2-2B and Pythia-160M, covering seven architectures, three widths, and six
target sparsity levels \citep{karvonen2025saebench}. Its eight metrics assess concept detection,
interpretability, reconstruction, and disentanglement.

These metrics give different architecture rankings. At $65$k latents and $L_0\in[40,200]$,
Matryoshka BatchTopK leads on $5$ of $8$ metrics despite an intermediate sparsity-fidelity ranking.
Increasing width from $4$k to $65$k improves reconstruction and automated interpretability for
every architecture, but worsens absorption for every architecture except Matryoshka. Spurious
correlation removal also worsens for most architectures, including after controlling intervention
size. \status{REPORTED-EMPIRICAL} The authors retain separate scores rather than combine them
with weights that would conceal these trade-offs. Supervised metrics also behave differently on
the smaller model, limiting transfer of the observed rankings across model scale.

\paragraph{Reliability and Ground-truth Checks}
A ranking supports comparison when the metric measures the intended property and separates the
methods beyond evaluation noise. \citet{chanin2026reliable} audit six SAEBench evaluations for
agreement with ground truth, improvement during training, reseed noise, discrimination between
variants, and internal consistency. Ground-truth checks use the hierarchical generator of
\cref{sec:prot-synthetic}, developed by the same first author.

The reseed measurements reveal markedly different precision across metrics. On a fixed
dictionary, coefficients of variation range from $0.2\%$ for probing-based scores to $23\%$ for
targeted probe perturbation at top-$N=50$. The audit's minimum reliable difference for two
single-seed scores at the $95\%$ level is about $3.93$ times the metric's reseed standard
deviation. In a three-seed panel, $7$ of $34$ metric configurations show a between-variant
variance share distinguishable from chance. Spearman agreement between metrics falls from
$+0.44$ across architectures to $+0.08$ across changes to one hyperparameter.
\status{NUMERICALLY-OBSERVED; preprint} These observations limit the differences that a
single-seed comparison can resolve, especially between similar dictionaries.

Ground-truth and training checks identify an additional problem with targeted probe perturbation
and spurious correlation removal. Both decline during training at top-$N=500$, and the latter
also declines at its canonical top-$N=10$ on two of four dictionaries. At that canonical setting,
it ranks an oracle below $11$ of $35$ trained dictionaries. The oracle contains $4{,}096$ of the
$16{,}384$ ground-truth directions and returns their exact coefficients.
\status{NUMERICALLY-OBSERVED; preprint} This disagreement concerns what the score rewards,
in addition to its sampling noise.

The audit recommends against these two metrics, weakening the interpretation of comparisons
that depend on them. Other SAEBench measurements still favor Matryoshka, including absorption,
which the audit does not test, and a disentanglement score that passes the audit. The combined
evidence supports retaining the individual measurements with their scope and uncertainty,
rather than treating the benchmark ranking as uniformly validated.

\subsection{Baselines and Negative Controls}
\label{sec:prot-baselines}

Comparing trained dictionaries tests relative performance, while controls test whether a score
depends on the representation or directions learned during training. The literature uses several
controls that remove different parts of the procedure. Their interpretation depends on which
part is randomized and which part remains trainable.

\begin{enumerate}
\item \emph{Non-dictionary alternatives.} Neurons, principal components, and raw residual-stream
  channels. \citet{karvonen2025saebench} report that dictionaries score above all three on sparse
  probing and automated interpretability, and note that reading residual channels directly is a
  weak baseline on automated interpretability.
\item \emph{A dictionary trained on a random transformer.} This control removes transformer training while preserving the dictionary-learning procedure. The gap is large where the ground truth is formal, as
  the chess coverage of \cref{sec:prot-organisms} showed \citep{karvonen2024boardgame}. On the
  language-model proxies the targeted probe perturbation score is \emph{higher} for the random
  model. The authors warn against over-reading this, since that score is measured relative to the
  same model's own probe accuracy and the random model's probes start much lower
  \citep{karvonen2025saebench}.
\item \emph{Frozen or constrained SAE parameters on the same trained model.} These controls keep the trained transformer and freeze or constrain different parts of the SAE. \citet{korznikov2026sanity}
  freeze the decoder at random initialization, freeze the encoder at random initialization, or
  constrain the decoder columns to stay within cosine $0.8$ of their initialization, and train
  everything else on real activations. These controls reach $0.87$ against $0.90$ on automated
  interpretability and $0.69$ against $0.72$ on top-1 sparse probing. On RAVEL causal editing they
  reach $0.73$ against $0.72$, and the constrained variant exceeds the trained dictionary at high
  $L_0$. \status{REPORTED-EMPIRICAL; preprint} The top-5 sparse-probing comparison shows a larger training effect,
  where the gaps widen to between $3.0$ and $7.9$ percent of the trained dictionary's score. The
  authors read this as learned latents aggregating better rather than as individual latents being
  better.
\item \emph{A frozen-at-initialization decoder scored causally.} On sparse control of the
  indirect-object-identification circuit, this control performs markedly worse than trained
  dictionaries \citep{makelov2025principled}. \status{REPORTED-EMPIRICAL}
\end{enumerate}

The frozen-random comparisons show that high scores on the tested proxies can arise with little
learned change in dictionary directions. Their interpretation has two qualifications. Biases,
thresholds, and whichever matrix remains unfrozen still receive training, so these controls do
not remove every learned component. Their performance instead limits how much the reported
scores establish about learned directions.

The comparison also depends on the site and task. \citet{korznikov2026sanity} summarize the
quoted pairs across architectures; the sparse-probing and causal-editing pairs match their
BatchTopK results on Gemma-2-2B layer 12 at $L_0=160$. At Gemma layer 19 and Llama-3-8B layer 16,
trained and control causal-editing scores cluster near $0.5$, making layer 12 the informative
comparison for that test. By contrast, the IOI sparse-control test distinguishes the frozen
decoder from trained dictionaries \citep{makelov2025principled}. These results qualify specific
metrics and protocols rather than establish that dictionary learning is universally ineffective.

Strong alternative methods provide a further comparison for applications. \citet{wu2025axbench}
compare dictionaries with prompting, finetuning, probes, and difference-in-means directions on
steering and concept detection, finding the dictionary methods uncompetitive in their evaluated
settings. \status{REPORTED-EMPIRICAL} \citet{kantamneni2025probing} test data scarcity, class
imbalance, label noise, and covariate shift. Dictionary probes underperform logistic regression
on average in every regime, with wins on individual datasets; adding dictionary latents to
baseline ensembles does not consistently improve them. \status{REPORTED-EMPIRICAL} These
comparisons assess application performance. They do not by themselves determine whether a
particular latent identifies a feature, which requires a recovery-specific reference.

\subsection{Metrics and Their Disagreements}
\label{sec:prot-metrics}

The controls above attach meaning to a score by testing how it changes when part of the recovery
procedure is removed. Interpreting the remaining difference also requires specifying the metric
and its normalization. \Cref{tab:metrics} compares the measured properties and documented
disagreements across reconstruction, probing, interpretation, and intervention tests.

\paragraph{Normalization and Sparsity}
Loss recovered compares model loss after replacing the activation at the fitted site with its
dictionary reconstruction against loss under a chosen ablation at that site. The reference uses zero ablation in \citet{karvonen2025saebench} and
\citet{karvonen2024boardgame}, while \citet{makelov2025principled} argue for and use mean
ablation. Scores with these different denominators require separate interpretation.

The sparsity level also changes the representation being tested. Although $L_0$ is often swept
as a sparsity-fidelity trade-off, \citet{chanin2026l0} report that too little activity promotes
mixtures of correlated features and too much activity admits degenerate mixtures. Thus choosing
$L_0$ changes more than the number of active latents.
\status{REPORTED-EMPIRICAL}

\paragraph{Comparison with Supervised References}
A supervised property test can expose differences hidden by reconstruction-based proxies.
On board-game models, Gated dictionaries score above $p$-annealed dictionaries on loss recovered,
while $p$-annealed dictionaries match or exceed Gated dictionaries on board-state coverage.
Dictionaries with $8{,}192$ latents also improve supervised scores over $4{,}096$-latent
dictionaries despite indistinguishable unsupervised proxies \citep{karvonen2024boardgame}.
\status{REPORTED-EMPIRICAL} These comparisons demonstrate why a score for preserved behavior
or activation variance needs a separate test of the properties being recovered.

\paragraph{Reliability and Validity}
Repeating a metric can estimate noise without establishing that the metric tracks feature
recovery. Reconstruction metrics are the most discriminative in the audit, but they measure
activation reconstruction rather than feature identity. The sae-probes variant is the most
reliable remaining metric, while probing-based scores can saturate. On the synthetic panel,
single-feature accuracy is near its ceiling at the canonical setting despite substantial
variation in Hungarian-matched cosine \citep{chanin2026reliable}.

The audit also recommends metrics left untested against synthetic ground truth, including scores
on which \citet{korznikov2026sanity} find frozen-random dictionaries near the trained scores.
Together, the two studies call for both repeatability checks and controls tied to the target
property. Low evaluation noise alone does not validate a score as a measure of recovery.

\begingroup\small
\renewcommand{\arraystretch}{1.25}
\begin{longtable}{@{}>{\raggedright\arraybackslash}p{0.18\textwidth} >{\raggedright\arraybackslash}p{0.21\textwidth} >{\raggedright\arraybackslash}p{0.19\textwidth} >{\raggedright\arraybackslash}p{\dimexpr0.42\textwidth-6\tabcolsep\relax}@{}}
\caption{Evaluation metrics for recovered features. The last column records where a metric has
been shown to disagree with another metric or with a supervised reference, not a general claim
that it is invalid.}\label{tab:metrics}\\
\toprule
Metric & What it measures & What it does not measure & Documented disagreement \\
\midrule
\endfirsthead
\multicolumn{4}{@{}l}{\small\tablename~\thetable\ (continued)}\\
\toprule
Metric & What it measures & What it does not measure & Documented disagreement \\
\midrule
\endhead
\midrule\multicolumn{4}{r@{}}{\small Continued on next page}\\
\endfoot
\bottomrule
\endlastfoot

Reconstruction error; explained variance &
share of activation variance the dictionary returns &
whether the returned directions carry the model's behavior &
most discriminative metric in the audit, yet $593$ against $2$ latents on a direction that is
functionally inert away from the first position \citep{braun2024e2e, chanin2026reliable} \\
$L_0$ &
mean active latents per token &
whether that level matches the data's sparsity &
mixing at too low and degenerate solutions at too high $L_0$ \citep{chanin2026l0};
decoder-geometry and firing-time ground-truth scores peak at different $L_0$
\citep{chanin2026reliable} \\
Loss recovered; KL of spliced outputs &
behavior preserved when the reconstruction replaces the activation &
which part of the behavior the residual carries &
zero-ablation and mean-ablation normalizations are not comparable
\citep{karvonen2025saebench, makelov2025principled} \\
Automated interpretability &
whether a language model predicts a latent's firing from its description &
whether the description names a unit the model uses &
$0.87$ for a frozen-random dictionary against $0.90$ for a trained one
\citep{korznikov2026sanity}; signal-to-noise ratio $1.8$, limiting reliable selection
from a single checkpoint \citep{chanin2026reliable} \\
Sparse probing, SAEBench default &
whether a named concept is readable from a few latents &
whether the model uses the concept &
saturated at the canonical setting, accuracy above $0.99$ while matched cosine spans $0.56$ to
$0.79$; signal-to-noise ratio $4.2$ at top-1, about two-thirds that of the sae-probes variant
\citep{chanin2026reliable}; top-1 near parity with frozen-random dictionaries
\citep{korznikov2026sanity} \\
Sparse probing, sae-probes variant &
the same, over $113$ datasets with cross-validated probe regularization &
whether the model uses the concept &
most reliable metric audited, at signal-to-noise ratio $5.4$ to $6.6$ against $1.8$ for automated
interpretability \citep{chanin2026reliable} \\
Disentanglement scores (spurious correlation removal, targeted probe perturbation) &
probe-accuracy change under ablation of probe-selected latents &
feature isolation, since the same probes select the latents and score the effect &
both decline during training at top-$N=500$, and spurious correlation removal declines at canonical
top-$N=10$ on two of four dictionaries; an oracle with exact coefficients ranks below $11$ of $35$ trained
dictionaries \citep{chanin2026reliable} \\
Causal editing (sufficiency, necessity, sparse control, RAVEL) &
whether reconstruction preserves behavior, removal disrupts it, or bounded latent edits reproduce a counterfactual effect &
whether the edited latents are the model's units &
frozen-random matches trained on RAVEL \citep{korznikov2026sanity} while a frozen-at-init decoder
is markedly worse on circuit-level sparse control \citep{makelov2025principled} \\
Ground-truth recovery (matched cosine, coverage) &
alignment with known directions or classification of known properties &
recovery of features beyond the supplied ground truth &
the cited studies use different generators, dictionary widths, and matching rules, so their
recovery numbers are not a common rate (\cref{sec:prot-synthetic}) \\

\end{longtable}
\endgroup

\subsection{Explained Variance as a Target}
\label{sec:prot-variance}

The disagreement between reconstruction and property tests has a behavioral counterpart.
Squared reconstruction error rewards directions according to their activation variance, while a
behavioral test measures the effect of changing them. A high-variance direction can have a small
behavioral effect, and a lower-variance component can matter more. We examine measurements of
this distinction, beginning with a construction based on output sensitivity.

\paragraph{The Jacobian Lens}
The global-workspace report constructs directions from the model's sensitivity to
activations \citep{gurnee2026workspace}. For each vocabulary token and layer, it averages the
linearized effect on the likelihood of producing that token at the current or later positions
over a corpus of contexts. The resulting Jacobian-lens vectors, called J-lens vectors in the
report, outnumber the residual-stream dimensions and can span the full space.

To isolate a sparse component, the report defines J-space through sparse nonnegative
combinations of J-lens vectors rather than their unrestricted span. Gradient pursuit finds these
combinations without training an autoencoder. A concept vector is separated into its nearest
J-space point and a remainder, allowing interventions on the two components to test their
behavioral effects separately.

The concept vectors used in these tests come from ``Tell me about \{concept\}'' prompts with
the mean of a hundred other concepts subtracted. For this selection, the best $16$ J-lens vectors
capture a median of $6$ to $7\%$ of concept-vector variance in the J-space component. The report
describes this sparsity choice as somewhat arbitrary and uses other choices elsewhere.

The intervention then tests the behavioral effect of that component. In a category-generation
task, exchanging it with the component of a second concept places the second concept in the top
five outputs on $59\%$ of trials. The corresponding rates are $88\%$ when exchanging the
J-lens vectors themselves and $5\%$ when exchanging the remainder.
\status{REPORTED-EMPIRICAL; institutional report} These interventions associate the
smaller-variance component with a larger effect in the tested task.

J-space also depends on a linearization, which leaves effects outside that approximation
unmeasured. The report uses proprietary checkpoints, is self-published, and lacks an identified
replication. These conditions limit generalization of the reported variance and intervention rates.

The report further finds a small fraction of dictionary decoder directions aligned with
J-space; unaligned latents predominantly represent low-level syntactic and bookkeeping
properties. This direction-level comparison does not test whether an SAE reconstruction recovers
the J-space component of an activation. It therefore leaves the causal component missed by a
particular dictionary unresolved.

The gap between variance and behavioral effect also appears in dictionary reconstructions.
\citet{braun2024e2e} find the highest-variance direction
at one site functionally inert under resample ablation after the first position, despite a local
dictionary assigning $593$ latents to it, compared with two for an end-to-end variant.
\status{REPORTED-EMPIRICAL}

Behavioral tests additionally distinguish a recovered direction from a latent that activates
when needed. In a controlled toy setting, \citet{bal2026audit} apply ablation
and steering to correlationally recovered features. They count a feature as causally inert when
the latent associated with its matched dictionary column remains silent whenever the feature is present. Among $22$
features matched above cosine $0.90$, this occurs for $2$ in a well-trained dictionary and up
to $17$ in a degraded dictionary. \status{REPORTED-EMPIRICAL; preprint, toy setting} The latter
comparison directly separates direction alignment from the activity needed for causal control.

These observations show why explained variance and behavioral effects require separate
measurements in the evaluated settings. The bounds in \cref{sec:geometry,sec:computation}
concern representable or computable features, rather than the number of directions selected by
a causal test. Extending such bounds to a specified causal criterion remains an open question
(\cref{sec:open}).

\subsection{Dark Matter}
\label{sec:prot-darkmatter}

\paragraph{Residuals and Their Measurements}
The behavioral components above raise a further question about what a dictionary leaves out.
A dictionary's omitted activation is the residual $e(h)=h-\hat h$. The literature refers to
this residual as SAE dark matter, but reports several quantities under that name. Fraction of
variance unexplained measures activation reconstruction. The complement of a graph replacement
score measures weighted attribution paths through error nodes. A third quantity counts prompts
whose resulting explanation is judged satisfying by the report's authors.

\Cref{tab:darkmatter} records these measurements separately. Activation variance, weighted
paths, and judged prompts have different denominators and test different properties. Their
numerical values consequently do not provide interchangeable estimates of an unexplained
fraction.

\begin{table}[tbp]
\centering\small
\renewcommand{\arraystretch}{1.3}
\caption{Published dark-matter quantities. The three senses are not commensurable, and the row
attributed to \citet{engels2025darkmatter} is the one refereed entry.}
\label{tab:darkmatter}
\begin{tabularx}{\textwidth}{@{}>{\raggedright\arraybackslash}p{0.185\textwidth}
  >{\raggedright\arraybackslash}p{0.185\textwidth}
  >{\raggedright\arraybackslash}p{0.145\textwidth} X@{}}
\toprule
Source & Model & Sense & Reported quantity \\
\midrule
Scaling report \citep{templeton2024scaling} &
Claude 3 Sonnet, middle-layer residual stream &
activation variance &
at least $65\%$ of activation variance explained, at each of the three dictionary widths;
institutional report \\
\citet{engels2025darkmatter} &
Gemma 2 9B, layer 20, open dictionary suite &
activation variance &
fraction of variance unexplained fitted as $0.186 + 2.708\,\hat d^{-0.317}$ in dictionary width
$\hat d$, with the asymptote extrapolated beyond the widths measured; refereed \\
Circuit-tracing report \citep{ameisen2025circuit} &
18-layer model, $10$M-latent cross-layer transcoder &
attribution-graph paths &
replacement score $0.61$ on unpruned graphs, against $0.37$ for per-layer transcoders of the same
size, so $39\%$ of the weighted path influence runs through the residual; institutional report \\
Case-study report \citep{lindsey2025biology} &
Claude 3.5 Haiku &
prompts yielding a satisfying explanation &
attribution graphs give satisfying insight on about a quarter of the prompts tried, which the
authors state as their own judgment and call difficult to quantify precisely; institutional
report \\
\bottomrule
\end{tabularx}

\end{table}

\paragraph{Contributions to the Residual}
The residual can contain structure poorly matched to a sparse, layer-local dictionary as well
as errors in fitting that dictionary. Frequency, layer coverage, and dense structure affect the
first possibility. Synthetic recovery favors frequently firing features under a heavy-tailed
law, leaving rarer directions less well recovered \citep{korznikov2026sanity}. A per-layer
analysis also has access to a feature's representation at that site rather than its full
cross-layer development (\cref{sec:rec-crosslayer}). This limits feature interpretation, but
cross-layer structure alone does not force a local activation-reconstruction error.

Dense structure poses a different mismatch with sparsity. \citet{engels2025darkmatter} model
activations as a sparse sum of feature directions plus a dense component outside any
low-dimensional linear subspace. A sparsity objective can leave part of that component in the
residual even when optimization succeeds. This is a proposed explanation for residual structure,
rather than an identified feature decomposition of trained-model activations.

Fitting introduces additional sources of error. An $\ell_1$ penalty shrinks latent magnitudes,
under-reconstructing even correctly identified directions \citep{rajamanoharan2024gated}.
Dead latents contribute no reconstruction, reducing the utilized dictionary width
\citep{gao2025scaling}. A dictionary of one-dimensional atoms can also approximate a curved
feature with multiple local directions. The counting-task report exhibits a curved
character-count representation \citep{gurnee2026manifolds}. Such a representation can require
more atoms for finer approximation, but curvature alone does not establish a nonzero
asymptotic reconstruction error.

\paragraph{Dictionary-width Scaling and Superposition}
Width scaling tests whether a larger dictionary reduces the residual within the fitted method.
\citet{gao2025scaling} fit reconstruction error with a power law plus a constant and leave the
constant's origin unresolved. Their proposed unstructured activation component is supported by
a Gaussian-data control of matched dimension, where the width exponent is roughly $-0.04$,
compared with $-0.26$ on GPT-2-small activations. \status{NUMERICALLY-OBSERVED}
\citet{engels2025darkmatter} also fit a positive asymptote, with the numerical fit reported in
\cref{tab:darkmatter}. An extrapolated constant describes the fitted scaling law beyond measured
widths; it is not a proved irreducible error for all recovery methods.

The interference bound in \cref{res:budget} concerns a different reconstruction problem.
For $d>m$ unit feature directions, it lower-bounds total squared cross-talk by $d(d-m)/m$.
Its consequence for isotropic linear coefficient reconstruction does not transfer to the SAE
map $h\mapsto z(h)$, which is nonlinear, or to the activation residual $h-Dz(h)-c$.
In particular, this bound establishes neither the existence of the fitted SAE constant nor
its dependence on $d/m$.

The dimensions also refer to different objects. Here $d$ is the unknown latent-feature count,
$m$ is the known activation width at the chosen site, and $\hat d$ is the chosen dictionary
width. A dictionary-width sweep varies $\hat d$, rather than measuring $d/m$. Cross-layer
representations require a separately specified joint space before its dimension can enter a
recovery bound (\cref{sec:rec-crosslayer}). Connecting the empirical asymptote to superposition
therefore requires additional assumptions and an argument for the nonlinear recovery setting.

\paragraph{Predictable Residual Structure}
A residual can be large yet contain structure predictable from the same activation used by the
SAE. \citet{engels2025darkmatter} test this by fitting regressions for the residual vector and,
separately, its squared norm. Vector prediction is scored by mean coordinatewise $R^2$; norm
prediction uses scalar $R^2$. Both regressions include an intercept.

\begin{result}[Linear predictability of the residual \citep{engels2025darkmatter}]
\label{res:darkmatter}
\status{REPORTED-EMPIRICAL}
Let $\tilde e(h)=Bh+b_e$ be the affine least-squares prediction minimizing
$\E\norm{Bh+b_e-e(h)}_2^2$, fitted per dictionary on its activation distribution.
\citet{engels2025darkmatter} report held-out mean coordinatewise $R^2$ from $0.15$ to $0.72$
for this error-vector prediction on open dictionary suites. The score decreases as dictionary
width or the active latent count $L_0$ grows. A separate intercept-including regression of
$\norm{e(h)}_2^2$ on $h$ reaches scalar $R^2$ from $0.70$ to $0.95$.
The remainder $e(h)-\tilde e(h)$ differs from $\tilde e(h)$ in two ways. Dictionaries trained
on the remainder reconstruct it worse, with fraction of variance unexplained $0.59$ against
$0.54$ for dictionaries trained on $\tilde e(h)$, and its norm is harder to predict from the
activation. In the reported replacement tests, each component raises cross-entropy in
proportion to its norm. Per-token error norms are predictable across dictionary scales,
indicating that larger dictionaries largely retain the difficult contexts of smaller ones.
\end{result}

\doesnot{Mean coordinatewise $R^2$ averages scores with different coordinate variances; it is
not automatically the fraction of total error-vector variance explained. The fitted prediction
identifies predictable residual structure, not interpretable features. The split is relative to
an affine predictor, so its remainder need not contain exclusively nonlinear feature structure.
The main width and sparsity sweeps use Gemma 2 2B layer 12 and Gemma 2 9B layer 20, with
additional layers examined in the source appendix. The other entries of \cref{tab:darkmatter}
use different settings and dictionary classes.}

The source informally summarizes error-vector predictability as about half, but the reported
coordinatewise statistic should be retained when comparing that claim with total variance.
At fixed $L_0$, the predictable component decreases with dictionary width while the remainder
stays roughly constant in the reported measurements. Predictability across scales also shows
that residual errors concentrate on similar contexts. These findings motivate evaluating the
residual's behavioral effect when dictionaries replace activations.

\paragraph{Residual Paths and Behavioral Coverage}
Circuit tracing measures that omission through error nodes rather than activation variance.
The circuit-tracing report compares a 10-million-latent cross-layer transcoder with per-layer transcoders
of the same total size on an 18-layer model \citep{ameisen2025circuit}. The replacement score
counts paths from embeddings to logits, weighted by strength, that pass through latents without
an error node. It rises from $0.37$ to $0.61$, leaving $39\%$ of weighted path influence through
the residual in the cross-layer model. The edge-based completeness score changes less, from
$0.78$ to $0.80$, while average path length decreases from $3.7$ to $2.3$.
\status{REPORTED-EMPIRICAL, institutional report} Shorter paths have fewer opportunities to
encounter error nodes, which offers an interpretation of the larger change in replacement score.

These graph scores measure a different omission from the norm of $e(h)$. The intervention
results in \cref{sec:prot-variance} likewise compare variance with effects on selected outputs,
without establishing a conversion between the two. A reconstruction report therefore needs its
activation metric and behavioral or graph metric stated separately.

\subsection{Computational and Data Cost}
\label{sec:prot-cost}

Repeated evaluations and control dictionaries add costs beyond training the candidate method.
\Cref{tab:cost} records the published settings and resources. A full benchmark pass takes
hours for one dictionary, and its audit spent more compute on evaluation than on dictionary
training. Repeated seeds can be necessary to resolve differences near the noise levels in
\cref{sec:prot-benchmarks}. Board-game evaluations take minutes, but their coverage remains
restricted to the specified properties. These costs make the evaluation target and required
precision relevant to the design of a comparison.

\begin{table}[tbp]
\centering\small
\renewcommand{\arraystretch}{1.3}
\caption{Reported compute and data for training and evaluating recovered dictionaries. Numbers are
as published and are not normalized to common hardware or a common token count.}
\label{tab:cost}
\begin{tabularx}{\textwidth}{@{}>{\raggedright\arraybackslash}p{0.24\textwidth}
  >{\raggedright\arraybackslash}p{0.30\textwidth} X@{}}
\toprule
Activity & Setting & Reported cost \\
\midrule
Training an open dictionary suite &
more than $400$ JumpReLU dictionaries over all layers of Gemma 2 2B and 9B plus selected layers of
27B &
over $20\%$ of GPT-3's training compute \citep{lieberum2024gemmascope} \\
Training one dictionary &
Gemma-2-2B residual stream, one site &
$500$M tokens per dictionary in both protocols
\citep{korznikov2026sanity, karvonen2025saebench} \\
Training one dictionary, model organism &
chess or Othello, $300$M tokens &
under $24$ hours on one A100; over $500$ dictionaries released
\citep{karvonen2024boardgame} \\
Evaluating one dictionary &
eight-metric suite, $16$k latents &
about $110$ minutes per dictionary plus $152$ minutes of one-time setup on one RTX 3090
\citep{karvonen2025saebench} \\
Evaluating one dictionary, model organism &
coverage and board reconstruction &
under $5$ minutes \citep{karvonen2024boardgame} \\
Auditing the benchmark &
snapshot panels, three seeds, plus a synthetic validity panel &
roughly $500$ to $650$ H100-hours of evaluation and about $84$ H100-hours of dictionary training
\citep{chanin2026reliable} \\
\bottomrule
\end{tabularx}

\end{table}

\subsection{The Structure of the Evidence Base}
\label{sec:prot-evidence}

Reproducibility depends on the model, dictionary, data, and evaluation procedure being available
for the reported comparison. Several measurements here come from self-published institutional
reports on proprietary checkpoints, including graph replacement, judged explanations,
production-model variance explained, and the J-space intervention results. Their methods and
reported observations can be examined, while access to the underlying checkpoints limits
independent repetition. Open-model studies include both refereed work and public preprints,
among them the benchmark audit, frozen-random controls, and \cref{res:darkmatter}.

Later analyses can also change the interpretation of an earlier decomposition. In the sequence of
crosscoder studies, apparent model-exclusive latents were traced partly to the objective,
without an estimated fraction. The revised objective still isolates the expected behavioral
difference \citep{lindsey2024crosscoders,mishrasharma2025diffing}. A separate study shows that
latents at one width decompose into latents at another \citep{leask2025canonical}, qualifying
the use of individual latents as canonical units in interpreting earlier dictionary reports
\citep{bricken2023monosemanticity,templeton2024scaling}. These results concern the interpretation
of recovered objects, which \cref{sec:failures} examines in detail.

Public resources support repeating such comparisons under shared conditions. Gemma Scope
releases dictionaries across model layers \citep{lieberum2024gemmascope}; SAEBench fixes a
training protocol and releases its compared dictionaries \citep{karvonen2025saebench}.
The metric audit releases its panels and code \citep{chanin2026reliable}, and synthetic
benchmarks release generators with known ground truth \citep{chanin2026synthsaebench}.
These resources allow reported differences to be checked against common data, controls,
and matching rules.

Even a repeatable score can support a narrower conclusion than an interpretation attached to it.
The frozen-random comparisons demonstrate this gap for specific proxy evaluations. We next
examine documented failures in the recovered dictionaries themselves, including cases that
evaluation scores leave undetected (\cref{sec:failures}).

\takeaway{sec:protocol}{High reconstruction and interpretation scores can coexist with poor
recovery of known features, and some proxy scores remain high with frozen-random directions.
Task-specific sparse control can distinguish trained dictionaries from those controls.
Activation variance also differs from behavioral importance: a small component can carry a
large intervention effect. Dictionary residuals retain predictable structure, while their
fitted width asymptotes have no established derivation from the linear interference bound.}

\section{Failure Modes, Illusions, and Negative Results}
\label{sec:failures}

The evaluations in \cref{sec:protocol} separate reconstruction, interpretation, and causal use,
but a dictionary can score well on one of these tests while misrepresenting the model's features.
Such errors can arise from reconstruction and sparsity objectives that split or merge features.
Variation across training runs adds a different problem, as similar information can be assigned
different directions. The recovered latents can also omit structure that affects behavior.
These omissions and changes in the recovered units limit what a description or intervention
establishes about a latent.

These failures concern five assumptions of sparse dictionary learning: linear feature format,
one-dimensional atoms, sparse coefficients, layer-local representation, and identifiability
(\cref{sec:rel-foundation}).
\Cref{tab:presuppositions} connects each presupposition to its documented failures or remaining
evidence gap. Layer-locality is treated with cross-layer recovery in \cref{sec:rec-crosslayer};
the following subsections examine the other failures and distinguish them from evidence against
superposition itself.

\begin{tcolorbox}[summarybox,title={Summary: Failure Modes, Illusions, and Negative Results}]
\roadmap
\begin{itemize}
\item Sparse reconstruction can favor splitting, absorption, and merging even when the generating
  features are known (\cref{sec:fail-splitting}).
\item Dictionary width, objective, and initialization affect which latents reproduce. Some studies
  recover stable subspaces despite unstable directions (\cref{sec:fail-canonical}).
\item Top-activating examples can support confident descriptions of random directions, and automated
  scores can miss failures of recovery (\cref{sec:fail-illusions}).
\item The reconstruction residual contains predictable structure and can carry behavior omitted from
  an interpretable circuit (\cref{sec:fail-darkmatter}).
\item Curved and dense representations challenge one-dimensional sparse dictionaries. These
  mismatches restrict recovery assumptions without refuting superposition
  (\cref{sec:fail-ontology,sec:fail-scope}).
\end{itemize}
\end{tcolorbox}

\begin{table}[tbp]
\centering\small
\renewcommand{\arraystretch}{1.35}
\caption{Presuppositions used in dictionary learning, the failures they permit, and the evidence
reported for each. A listed failure need not have been established in every setting.}
\label{tab:presuppositions}
\begin{tabularx}{\textwidth}{@{}>{\raggedright\arraybackslash}p{0.19\textwidth}
  >{\raggedright\arraybackslash}p{0.25\textwidth} X
  >{\raggedright\arraybackslash}p{0.15\textwidth}@{}}
\toprule
Presupposition (\cref{sec:rel-foundation}) & Failure when it does not hold & Evidence &
Documented in \\
\midrule
Features are linear directions (\cref{def:lrh}) &
a concept carried by a curved function of the activations has no direction to recover &
the surveyed circular subspaces of \citet{engels2025notall} are
linear; these examples do not establish a failure of linear subspace representation &
\cref{sec:fail-ontology} \\
Atoms are one-dimensional &
a group of atoms tiles the feature instead of spanning it &
two-dimensional circular day, month, and year subspaces \citep{engels2025notall}; a counting task
in which the model manipulates a curved low-dimensional manifold, reported by \citet{gurnee2026manifolds}; five architectures on Llama-3.1-8B plateau far
beyond each manifold's ambient dimension \citep{bhalla2026manifolds} &
\cref{sec:fail-ontology} \\
Coefficients are sparse &
a sparse dictionary can leave dense structure in its residual &
the residual is modeled as unlearned features, a dense term, and a term the autoencoder introduces
\citep{engels2025darkmatter}; the width scaling of reconstruction error needs a constant term
\citep{gao2025scaling}; separately, removing residual-stream error nodes severely disrupts
a circuit model, without isolating the dense component \citep{marks2025sparse} &
\cref{sec:fail-darkmatter} \\
Atoms are layer-local &
a per-layer dictionary decomposes a projection of an object that is not layer-local &
The crosscoder report finds latents whose decoder norm is substantial across most layers,
and states that this is consistent with cross-layer superposition without establishing it
\citep{lindsey2024crosscoders} &
\cref{sec:rec-crosslayer} \\
The dictionary is identifiable &
absorption, splitting, hedging, and seed-dependent atoms &
\cref{res:absorption}; over-splitting preferred with two true features
\citep{makelov2025principled}; stitching and meta-SAEs \citep{leask2025canonical} &
\cref{sec:fail-splitting,sec:fail-canonical} \\
\bottomrule
\end{tabularx}

\end{table}

\subsection{Feature Splitting, Absorption, and Hedging}
\label{sec:fail-splitting}

\paragraph{Splitting.}
A wider dictionary can replace one latent with several more selective latents.
\citet{cunningham2024sae}, for example, interpret context-specific variants of an apostrophe
latent as finer monosemantic units. Such a refinement could reflect structure in the model, but
it could also improve the dictionary objective without recovering additional features.
\citet{makelov2025principled} distinguish these explanations using a uniform mixture of two
isotropic Gaussians in $\R^{100}$, where the generating feature count is fixed at two.

On this distribution, a randomly constructed dictionary with $1000$ latents has a lower combined
reconstruction error and sparsity penalty than the best two-latent dictionary. The advantage
holds for every penalty weight up to about $7$ on the reported grid. The construction is
independent of the sample, placing the comparison in the infinite-data limit.
\status{NUMERICALLY-OBSERVED} The objective can thus prefer a split representation without
multi-scale structure in the generating features.

The same paper reports splitting in GPT-2 small. A binary position attribute in the
indirect-object-identification task is distributed across at least ten latents, which activate
on small, mostly disjoint, semantically opaque prompt sets. \status{REPORTED-EMPIRICAL}
These observations motivate a causal test of whether the split latents have distinct roles.
We did not identify a study that establishes such a separation for the reported split.

\paragraph{Absorption.}
Splitting distributes a concept across latents, while absorption makes a concept's latent silent
on part of the concept's support. Its top-activating examples can remain precise even though
other inputs containing the concept receive no activation. \citet{chanin2025absorption} derive
this behavior for a feature hierarchy in which a child feature activates on a subset of the
inputs containing its parent. The following construction holds reconstruction fixed while
varying the sparsity penalty.

\begin{result}[Absorption lowers the sparsity penalty and leaves the reconstruction unchanged
\citep{chanin2025absorption}]
\label{res:absorption}
\status{PROVED within the stated family}
Let $w_1, w_2 \in \R^m$ be orthonormal feature directions and let feature $2$ fire only when
feature $1$ fires, both with value $1$, so that $h$ takes the values $0$, $w_1$ and $w_1 + w_2$
with probabilities $p_0$, $p_{10}$ and $p_{11}$. Take a two-latent ReLU dictionary without biases,
with encoder rows $u_i$, latent activations $z_i = \mathrm{ReLU}(u_i^\top h)$, and dictionary
columns $D_i$ given by
\[
  u_1 = w_1 - \delta\, w_2, \quad u_2 = w_2,
  \qquad
  D_1 = w_1, \quad D_2 = w_2 + \delta\, w_1,
  \qquad \delta \in [0,1] .
\]
For every $\delta$ the reconstruction equals $h$ on all three values. The expected penalty
$\E\norm{z}_1$ is $p_{11}(2 - \delta) + p_{10}$, which decreases in $\delta$ at rate $p_{11}$. At
$\delta = 1$ latent $1$ does not fire when feature $2$ is active, and $D_2$ carries the whole
reconstruction there.
\end{result}

\doesnot{The statement covers a single one-parameter family with orthonormal feature directions, unit
feature magnitudes, unnormalized dictionary columns, and no biases, so it does not establish that a global minimizer of the objective
absorbs. The absorption rates reported below are measured against supervised probe directions
rather than against the model's own units.}

At $\delta=0$, the encoder rows and dictionary columns equal the feature directions. Increasing
$\delta$ transfers the parent's contribution on joint activations to the child's decoder column.
The family ends at an absorbing solution, giving a mechanism for missed activations despite
unchanged reconstruction.

The trained-model example in \citet{chanin2025absorption} concerns first-letter spelling in
Gemma-2-2B. A latent for words beginning with S reaches $F_1=0.81$ against a logistic probe but
stays silent on \texttt{\_short}. A token-aligned latent for \emph{short} activates there with
probe cosine $0.12$. Zero-ablation of this latent at the \texttt{\_short} token reduces the correct-letter logit
minus the mean incorrect-letter logit by more than $6$, while every other latent's effect has
magnitude below $0.5$. Projecting the probe direction out of that latent removes the effect.
\status{REPORTED-EMPIRICAL} This intervention supports absorption for one latent in one dictionary.

The broader dictionary sweep reports mean absorption rates up to about $0.35$, increasing with
width and sparsity. These rates depend on three hand-set thresholds and a requirement that one
absorbing latent dominate, without a reported false-positive estimate. The trained-model tests
cover first-letter spelling, and the metric becomes silent at and above layer $18$ of
Gemma-2-2B, where letter information has moved to the final token. The reported rates consequently
describe this task and measurement procedure.

\paragraph{Width and Sparsity Level.}
Narrow dictionaries introduce a complementary error. A dictionary with fewer latents than
underlying features can reduce reconstruction error by merging correlated components into a
single latent, a behavior called hedging \citep{chanin2025hedging}. This merging strengthens as
the dictionary narrows and arises from the reconstruction term.
\status{REPORTED-EMPIRICAL; preprint} The sparsity level also affects merging.
\citet{chanin2026l0} report mixtures of correlated features at low $L_0$ and degenerate mixtures
at high $L_0$, and conclude that many commonly used dictionaries set $L_0$ too low.
\status{REPORTED-EMPIRICAL}

Calling a merged latent a recovery error requires specifying the recovery target. The model
itself merges correlated features in the setting of \cref{sec:geo-shapes}. A dictionary that
reproduces this assignment can disagree with the generating feature list while describing the
model correctly. Distinguishing these cases requires evidence about the model's representation
in addition to the dictionary's reconstruction.

The empirical width comparisons also depend on the evaluation target. The absorption and editing
trade-offs in SAEBench are detailed in \cref{sec:prot-benchmarks}
\citep{karvonen2025saebench}. That comparison fixes training tokens rather than compute and
evaluates each dictionary once. In board-game models, wider dictionaries improve supervised
board-state scores while remaining indistinguishable on unsupervised proxies
\citep{karvonen2024boardgame}; \citet{gao2025scaling} report general improvement on their chosen
metrics. \status{REPORTED-EMPIRICAL} Width therefore changes several properties that a single
score need not rank together. Matryoshka training addresses this dependence by imposing nested
prefixes (\cref{sec:rec-dictionaries}), but the nesting supplies a hierarchy in advance rather
than establishing that the model uses that hierarchy.

\subsection{Non-Canonicity and Seed Instability}
\label{sec:fail-canonical}

\paragraph{Dictionary Width.}
The width trade-offs raise a further question about latent identity. If different widths
recover the same underlying units, their differences should reflect completeness or resolution
rather than a change in what counts as a unit. \citet{leask2025canonical} define canonical units
as unique, complete, and atomic, then test completeness by transplanting latents from a larger
dictionary into a smaller dictionary fitted to the same activations.

Transplanted latents improve reconstruction when their decoder columns have low maximum cosine
similarity to the smaller dictionary's columns. Adding all such latents reduces mean squared
error by about $10\%$, providing evidence that the smaller dictionary omits information.
Atomicity is tested with a meta-SAE containing $2{,}304$ latents, fitted to the $49{,}152$ decoder
columns of a GPT-2 dictionary. It explains $55.47\%$ of their variance with about four active
meta-latents per column. The authors interpret these decompositions as compositions of concepts.
\status{REPORTED-EMPIRICAL}

Compositional latents are compatible with the sparsity objective. A conjunction represented by
one active latent can incur less penalty than a representation activating a latent for each
component (\cref{sec:rec-dictionaries}). However, the reported tests leave the relation to the
model's own units unresolved. The transplant procedure thresholds a continuous cosine
similarity distribution, and the meta-SAE reconstructs decoder weights without testing
activations or interventions.

\paragraph{Training Seeds.}
A fixed width permits a separate test of whether latent identity depends on initialization.
As a control, \citet{leask2025canonical} train two width-$3072$ dictionaries with independent
seeds. They find that $94\%$ of the first dictionary's latents have a decoder column with cosine
similarity at least $0.7$ to a column in the second. \status{REPORTED-EMPIRICAL}
This aggregate agreement leaves open which latents reproduce and what information the unstable
latents carry.

\citet{gerasimov2026unstable} address that distinction by estimating the probability that a
similar latent reappears in an independent training run. Stable latents carry most of the
reconstruction-relevant and prediction-relevant signal in their experiments. Unstable latents
have weak marginal effects and predominantly low-frequency surface-form triggers, yet concentrate
in reproducible lower-rank subspaces. The authors interpret this concentration as ambiguity
among bases for shared information. A synthetic model supports that explanation by recovering a
known low-rank subspace without identifying individual latents across seeds.
\status{REPORTED-EMPIRICAL and NUMERICALLY-OBSERVED; preprint}

Related evidence comes from \citet{bhalla2026manifolds}, whose optimal-transport comparisons find
weak overall alignment of decoder directions but stronger alignment within individual concept
manifolds. The training objective also changes stability. Pure end-to-end dictionaries learn
different features across seeds, while local and end-to-end-plus-downstream variants are
seed-stable in the experiments of \citet{braun2024e2e}. \status{REPORTED-EMPIRICAL}
These findings support reporting direction and subspace agreement separately when both are
measured, without prescribing subspaces as the recovery target for every task.

Agreement also depends on how the runs are initialized. \citet{brzozowski2026archetypes}
distinguish endpoint agreement, which they call stability, from convergence of independently
initialized runs, which they call stabilization. The reported stability of archetypal
dictionaries follows from a shared deterministic initialization that makes their distance zero
before training. Under independent initialization, the archetypal constraint gives no
stabilization advantage in their setting. The study also identifies a preprocessing-dependent
cosine geometry that complicates endpoint comparisons.
\status{REPORTED-EMPIRICAL; preprint}

Together, these studies make latent reproducibility conditional on the objective, initialization,
and level of comparison. Some support a stable subspace despite unstable directions; others
support stable directions under particular objectives. The distinction matters for
interpretation because a description of one unstable latent need not transfer to another run.
It also parallels the distinction between accessibility and identification in
\cref{sec:acc-ident}, where L2 does not imply L3.

\subsection{Interpretability Illusions}
\label{sec:fail-illusions}

\paragraph{Top-Activating Examples.}
Reproducibility does not establish that a latent's description matches its role. A description
inferred from top-activating examples can reflect regularities in the examples even for a random
direction. \citet{bolukbasi2021illusion} demonstrate this problem with final-layer BERT
embeddings of sentences from four corpora. One coordinate supports three confident but mutually
incompatible descriptions across three corpora, and annotators find an average of $2.5$ distinct
patterns per coordinate across the four corpora.

The random-direction control produces a similar pattern rate. Annotators find patterns in $80\%$
of top-activating sets for $25$ coordinates and $82\%$ for $33$ random directions, compared with
$14\%$ for $29$ random sentence sets. The paper reports no confidence intervals.
\status{REPORTED-EMPIRICAL} In this setting, a convincing description does not distinguish a
model coordinate from a random direction.

The authors attribute these descriptions to separable corpus regions, local coherence in the
embedding space, and annotator propensity. Testing a description on other corpora addresses its
dependence on the examples used to construct it. Such a test can reject a corpus-specific
description, while the model's use of the direction remains a separate causal question.

\paragraph{Automated Scores.}
Automated evaluation inherits a related limitation when the score depends weakly on the learned
directions. The frozen-random comparisons in \cref{sec:prot-baselines} show near-parity on
several scores at the reported sites \citep{korznikov2026sanity}. Such parity limits what a high
score establishes about decoder learning, even when the score measures a reproducible property
of the dictionary. The site-specific RAVEL saturation and the exceptions on sparse probing and
circuit control remain essential to that interpretation.

Metric reliability addresses a different concern. The audit in \cref{sec:prot-benchmarks}
identifies seed noise, training-time reversals, and rankings inconsistent with synthetic ground
truth \citep{chanin2026reliable}. Its synthetic validity checks use the companion generator of
\citet{chanin2026synthsaebench}. By contrast, recommendations for automated interpretability
and RAVEL rest on reliability and discrimination between variants, since their natural-language
concepts were not tested against that ground truth. The frozen-random results show why these properties
alone do not establish validity as tests of feature recovery. Architecture comparisons inherit both the measured noise and
this uncertainty about what their scores reward.

\subsection{Dark Matter and Accumulated Replacement Error}
\label{sec:fail-darkmatter}

\paragraph{Structure in the Residual.}
A dictionary's errors also concern the activation structure its latents omit.
\Cref{res:darkmatter} reports linear predictability of the error vector and a separate,
stronger prediction of its squared norm \citep{engels2025darkmatter}. The error-vector result
uses mean coordinatewise $R^2$; it does not establish that most of every residual is linearly
predictable. Predictable error and repeated high-error contexts across dictionary scales show
that the residual contains systematic omissions. The relevant question for a circuit is how
much behavior depends on those omissions.

\paragraph{Accumulation Across Layers.}
Circuit methods retain reconstruction errors as explicit nodes, allowing the effect of their
removal to be tested (\cref{sec:rec-replacement}). \citet{marks2025sparse} report that removing
residual-stream error nodes severely disrupts their model and limits its maximum performance.
Removing attention and MLP error nodes is less disruptive.
\status{REPORTED-EMPIRICAL} The recovered circuit consequently relies on an uninterpreted
remainder whose behavioral role is not summarized by reconstruction accuracy alone.

At frontier scale, the local replacement model uses error nodes to match the original
model on a fixed prompt by construction \citep{ameisen2025circuit}. This agreement retains the
transcoders' unexplained computation in the surrogate, with attention patterns frozen as described
in \cref{sec:rec-replacement}. The corresponding attribution-path and prompt-level measurements
are distinguished in \cref{sec:prot-darkmatter} \citep{lindsey2025biology}.
\status{REPORTED-EMPIRICAL, institutional report} Successful replacement can thus support L4
while leaving L6 unresolved, because exact prompt-level agreement includes computation assigned
to error nodes rather than recovered latents.

\subsection{Ontology Mismatch}
\label{sec:fail-ontology}

\paragraph{One-Dimensional Atoms and Curved Features.}
Some omitted structure reflects a mismatch between the dictionary's atoms and the representation
being described. One-dimensional atoms can cover a curved low-dimensional representation with
local detectors instead of recovering a compact spanning group. \citet{engels2025notall}
recover circular day, month, and year subspaces in GPT-2-small, and circular day and month
subspaces in Mistral 7B. In Mistral 7B and Llama 3 8B, patching the circular subspace reproduces
most of the whole-layer patching effect. The authors interpret this result as sufficiency when
the remaining layer is ablated, rather than evidence that the model computes with the circle.
\status{REPORTED-EMPIRICAL}

The interpretation depends on the model and test. GPT-2-small has clean circles but answers
$8$ of $49$ day prompts and $10$ of $144$ month prompts correctly, and its circles receive no
causal test. A competing explanation constructs a similar circle by compressing twelve correlated
one-dimensional features \citep{prieto2026correlations}, leaving the underlying feature
dimensionality open (\cref{sec:rel-foundation}).

Beyond identifying curved geometry, \citet{gurnee2026manifolds} study its computational role
in a counting task in which the model manipulates a low-dimensional curved manifold that dictionary
atoms discretize.
\status{REPORTED-EMPIRICAL, institutional report}

To assess recovery beyond individual atoms, \citet{bhalla2026manifolds} distinguish subspace capture, where a compact group of atoms spans
the manifold, from local coverage that is disjoint (shattering) or redundant (dilution).
Their Llama-3.1-8B experiments cover five architectures, two expansion factors, and three sparsity
levels. Explained variance continues to increase as more atoms are allowed and plateaus far
beyond the manifold's ambient dimension, supporting dilution in these experiments. Individual
atoms behave like tuning curves; on the years manifold, most tested dictionaries learn both
ones-digit and decade-selective atoms. \status{REPORTED-EMPIRICAL; preprint}

A description of an individual atom in this setting can identify a local region without
characterizing the full manifold. Recovering the relevant group therefore becomes a separate
problem. \citet{engels2025notall} cluster decoder columns by cosine similarity, while
\citet{bhalla2026manifolds} compare decoder cosine similarity with statistics of which latents
fire together, including conditional co-activation probabilities and pairwise Ising couplings
fitted to binary firing indicators. The latter two statistics separate the colors, days, and
temperature groups where decoder cosine similarity fails in Llama-3.1-8B. A comparison on shared
ground truth would be needed to rank these grouping methods across those settings.

The Ising affinity has a specific theoretical limit. Its conditional-independence interpretation
assumes a pairwise Ising distribution with positive probability for every binary firing pattern.
A fixed-cardinality TopK activation pattern violates this support condition, so fitted couplings
do not certify conditional independence of the atoms.
The paper's recovery theorem assumes a spanning set of atoms that its measurements do not find
and uses a matching-pursuit decoder instead of the trained encoder. The reported principal-component
plots of template sweeps at one layer consequently motivate the manifold interpretation without
establishing the theorem's premises for those dictionaries.

\paragraph{Dense Structure.}
Dense structure challenges the sparsity assumption rather than the dimensionality of an atom.
A latent representing a frequently active feature incurs the sparsity penalty on many inputs,
which can make leaving that structure in the residual preferable under the objective.
\citet{engels2025darkmatter} model the residual as unlearned features, a dense term, and a term
introduced by the autoencoder. Their separation assumes that the introduced term is not linearly
predictable from the input; violations can inflate the estimated dense component.

A related indication comes from the constant term fitted to the width scaling of reconstruction
error by \citet{gao2025scaling}, with its Gaussian control discussed in
\cref{sec:prot-darkmatter}. The constant is an extrapolated fit rather than a measured
irreducible error. These findings motivate distinguishing sparse structure from other activation
components. The uniform sparse-recovery bounds in \cref{sec:access}
assume sparse feature vectors and do not directly bound a representation containing an
additional dense component.

\subsection{Scope of the Negative Results}
\label{sec:fail-scope}

\paragraph{Artifacts of the Objective.}
The preceding failures concern an estimator or an evaluation procedure, whereas superposition
is a property of the assignment of features to directions (\cref{def:superposition}).
\Cref{res:absorption} makes the distinction explicit by producing absorption with two
orthonormal feature directions, without superposition. Its hierarchy and sparsity penalty
suffice for the effect. The splitting and hedging analyses likewise use stipulated generating
features to expose behavior favored by the objective. Applying their conclusions to a trained
model requires establishing how those generating features relate to the model's representation.

\paragraph{Different Randomization Controls.}
The controls in \cref{sec:prot-baselines} expose failures of particular recovery scores
\citep{karvonen2024boardgame,makelov2025principled,karvonen2025saebench}.
A score insensitive to learned dictionary directions leaves the trained model's representation
unresolved. The control qualifies the score's interpretation; it does not determine whether
the model itself uses shared feature directions.

\paragraph{Evidence About Superposition.}
Dense and curved structure restrict assumptions used in the theory without contradicting
conclusions proved under those assumptions. Incidental polysemanticity
(\cref{sec:acc-incidental}) and corpus-dependent descriptions \citep{bolukbasi2021illusion}
also provide explanations for some observations that motivate the superposition hypothesis.
Testing that hypothesis requires the representation measurements described in
\cref{sec:protocol} and the targets developed in \cref{sec:open}. The applications in
\cref{sec:applications} ask a related but distinct question about whether recovered latents
support reliable detection, intervention, or mechanistic prediction despite these limitations.

\takeaway{sec:failures}{A sparsity penalty can make a concept latent fall silent precisely
where a more specific latent reconstructs it. Wider dictionaries can split a fixed generating
feature, and repeated training can recover stable subspaces with unstable individual directions.
These failures concern the recovered units. Residual error creates a separate limitation: a
circuit can preserve the model's output while relying on computation carried by uninterpreted
error nodes.}

\section{Interventions and Applications}
\label{sec:applications}

The failures in \cref{sec:failures} make a recovered latent's usefulness an empirical question.
A latent may detect a concept despite an incomplete description, while an intervention on the
same latent may alter additional behaviors. Applications therefore require tests of the intended
use, alongside the recovery evaluations in \cref{sec:protocol}.

Monitoring tests whether a variable can be read from activations; interventions test the effects
of changing them; circuit analysis seeks to predict those effects from a recovered computation.
A successful intervention supports a causal claim at the level specified in \cref{sec:def-ladder},
but does not by itself establish that features outnumber dimensions. The discussion below separates
these application results from the representation evidence relevant to superposition.

\begin{tcolorbox}[summarybox,title={Summary: Interventions and Applications}]
\roadmap
\begin{itemize}
\item Supervised directions provide strong baselines for detecting named concepts, while
  dictionaries also propose concepts that were not specified in advance
  (\cref{sec:app-monitoring}).
\item Steering comparisons depend on latent selection, intervention magnitude, capability loss,
  and off-target effects (\cref{sec:app-steering}).
\item Controlled editing can remove a spurious cue; circuit graphs seek to predict intervention
  outcomes but retain unexplained computation (\cref{sec:app-editing,sec:app-circuits}).
\item Robustness and fine-tuning studies connect feature overlap to behavioral changes, while
  model diffing also exposes artifacts of the recovery objective
  (\cref{sec:app-robustness,sec:app-diffing}).
\item Studies beyond language compare recovered concepts with biological, physical, or task-level
  references, with mixed evidence for useful interventions (\cref{sec:app-architectures}).
\end{itemize}
\end{tcolorbox}

\subsection{Monitoring and Concept Detection}
\label{sec:app-monitoring}

\paragraph{Named Concepts and Unsupervised Candidates.}
A monitor reports whether a concept is present in the current activations. A concept specified
in advance permits a supervised direction fitted by the methods of \cref{sec:rec-supervised}.
An unsupervised dictionary offers a different benefit by proposing candidate concepts before a
monitoring target is selected. Detection tests establish L1 for the monitored variable, and
support L2 when the decoder class and tolerance $\varepsilon$ are specified.

\paragraph{Frontier-Scale Demonstrations.}
Using this unsupervised approach, \citet{templeton2024scaling} train dictionaries in the
middle-layer residual stream of Claude 3 Sonnet at three widths up to $33.5$ million latents.
The resulting dictionaries include latents described as responding to code vulnerabilities,
bias, sycophancy, deception, power-seeking, and dangerous or criminal content. Clamping selected
latents changes outputs consistently with their proposed interpretations, supporting L5 for
the individual interventions tested. \status{REPORTED-EMPIRICAL, institutional report}

These selected cases demonstrate candidate discovery rather than measured performance on a
safety-monitoring task. The steering factors are tuned by hand using qualitative indicators,
and the examples are selected successes. The authors expect their recovery to remain orders
of magnitude short of the model's feature inventory. A monitoring comparison consequently
requires a separate evaluation of the candidates' detection performance.

\paragraph{Comparisons on Named Concepts.}
For concepts fixed before evaluation, \citet{wu2025axbench} compare dictionary detectors with
supervised alternatives across $500$ concepts, evaluating two residual
sites each in Gemma-2-2B and Gemma-2-9B. Difference-in-means directions, linear probes, and a
rank-one representation-finetuning method give the strongest detection performance, with no
statistically significant difference among them. Five supervised methods fitted on small
synthetic datasets outperform autoencoder latents. \status{REPORTED-EMPIRICAL}
The Gemma-2-2B comparison uses a base-model dictionary on the instruction-tuned model because a
matching dictionary was unavailable, which qualifies the comparison at that model size.

A concept basis might be especially useful when supervision is limited or unreliable.
\citet{kantamneni2025probing} test data scarcity, class imbalance, label noise, and covariate
shift. Autoencoder probes win on some individual datasets but fall below logistic regression on
average in each setting. Ensembles combining autoencoder probes with baseline methods also fail
to consistently improve on ensembles of those baselines alone.
\status{REPORTED-EMPIRICAL} These results favor supervised alternatives for the named concepts
tested, while leaving the benefit of discovering previously unspecified concepts unmeasured.

\paragraph{Activation Versus Direction Agreement.}
A useful detector also needs the selected latent to activate when its concept is present.
\citet{bal2026audit} test this requirement by ablating and steering latents matched to known
feature directions in a synthetic setting (\cref{sec:prot-variance}). Some matched latents
remain silent whenever the feature is present, both in a well-trained autoencoder and in a
degraded autoencoder. Decoder-column agreement alone consequently does not establish detection.
\status{NUMERICALLY-OBSERVED in a synthetic setting}

The same harness tests a published GPT-2-small dictionary against probe directions with a lower
cosine threshold of $0.5$. It finds $7$ of $83$ concepts above the threshold, with $1$ of those
$7$ causally inert. The author explicitly separates this rate from the synthetic measurements.
\status{REPORTED-EMPIRICAL on a trained GPT-2 dictionary; single-author preprint}
These comparisons distinguish geometric matching from the firing-pattern test needed for a
monitor.

\subsection{Steering}
\label{sec:app-steering}

\paragraph{Target Behavior and Preserved Behavior.}
Steering changes activations by adding a multiple of a recovered direction or clamping a latent
during the forward pass. Its intended effect is a change in target behavior with acceptable
changes elsewhere. Evaluating that goal requires the target response over intervention
magnitudes, effects on behaviors intended to remain unchanged, and general capability loss.
A target score at one magnitude leaves the trade-offs among these quantities unresolved.

\paragraph{Magnitude and Capability Loss.}
\citet{wu2025axbench} compare steering methods by sweeping intervention magnitude and scoring
fluency, instruction following, and concept presence. The benchmark combines the three judge
scores as a harmonic mean on a $0$ to $2$ scale, selecting a steering factor separately for each
concept and method. This procedure measures a capability trade-off but does not explicitly test
off-target concepts.

Prompting obtains the highest mean score, $0.894$, followed by finetuning methods. The rank-one
representation-finetuning method scores $0.543$, difference-in-means scores $0.239$, and
autoencoder latents score $0.165$. Across the four sites, difference-in-means wins $58.7\%$ of
per-concept comparisons with autoencoder latents, while the rank-one method wins $81.8\%$;
ties count as half a win. \status{REPORTED-EMPIRICAL}

The magnitude sweeps explain why a single intervention score is incomplete. Increasing the
steering factor monotonically reduces instruction following for every tested method. Concept
presence rises with magnitude at later layers but rises and then falls at earlier layers.
\status{REPORTED-EMPIRICAL} The selected operating point thus depends on both the intervention
site and the weighting of capability against concept presence.

Latent selection also changes the comparison. \citet{jorgensen2026steering} replace the
benchmark's automated latent labels with supervised labeling and selection, after which
autoencoder steering approaches the benchmark's LoRA reference while remaining below prompting.
\status{REPORTED-EMPIRICAL; preprint} This counterexample limits a broad ranking of dictionaries
against supervised methods: the reported performance belongs to the dictionary together with
its selection procedure. Evidence that downstream circuit connections improve selection is
discussed in \cref{sec:app-circuits}.

\paragraph{Off-Target Effects.}
Explicit tests of preserved behavior are available outside language models.
\citet{lehnschioler2026eeg} rank dictionary latents by alignment with a target clinical concept
direction. They clamp an increasing fraction of the dictionary to the corresponding mean latent
values in a target group, such as normal recordings when steering abnormal recordings toward
normal. The modified latents are decoded into embeddings and scored by target and off-target
probes fitted on clean decoded embeddings and held fixed during the sweep. At each clamping fraction, the probes are evaluated by the area under the receiver operating
characteristic curve (AUROC). The statistic integrates the gap between their scores over that
fraction, then subtracts the expected gap under random clamping. This comparison separates selectively steerable concepts
from encoded but entangled concepts and from concepts not encoded under the test. The entangled
cases include collapsed global performance and age-pathology coupling that prevents either
variable from being suppressed independently.
\status{REPORTED-EMPIRICAL; preprint}

Diffusion-model erasure illustrates a failure after successful localization.
\citet{cassano2026unlearning} apply published autoencoder-based erasure pipelines and examine
the modified activations. At the median negative multiplier, $44.7\%$ to $100\%$ of activations
lie outside the unmodified norm distribution, depending on the autoencoder. A vision-language
judge finds artifacts in $43.1\%$ to $61\%$ of generated images, against $2\%$ and $1\%$ for
the two unmodified generators. Weaker multipliers leave some target concept unerased.
\status{REPORTED-EMPIRICAL}

The authors interpret this damage as movement outside the activation distribution on which the
model was trained. Using the latents as detectors instead, then replacing flagged patch
embeddings, reduces SAeUron's artifact rate from $57\%$ to $16.3\%$. The comparison locates the
failure in multiplier-based control rather than concept localization. The repair uses the
dictionary to identify patches; it does not establish selective generation control through the
latent itself.

\paragraph{A Large Effect From a Single Latent.}
A substantial target effect is also possible without an accompanying selectivity measurement.
The November 2025 circuits update \citepalias{anthropic2025november} reports that Claude 3.5 Haiku answers $129$ two-choice
questions with $100\%$ accuracy. Adding harmful intent reduces accuracy to $48.1\%$, and
negative steering of one refusal latent restores it to $93\%$.
\status{REPORTED-EMPIRICAL, institutional report}
This result supports an L5 claim for the latent on the selected prompt set.

The restored quantity is accuracy on those questions, rather than a direct measure of refusal
removal. The preliminary update reports no broad off-target evaluation or systematic capability
comparison, leaving selectivity beyond the chosen questions unmeasured in the report.
A large selected-case effect remains compatible with lower average
performance in a benchmark that fixes its concept list and sweeps intervention magnitude.

\subsection{Unlearning and Targeted Editing}
\label{sec:app-editing}

\paragraph{Removing a Cue While Preserving a Task.}
Targeted editing aims to remove a capability or spurious cue while preserving other behavior.
Overlapping feature directions can make this difficult because changing the component along one
direction also changes projections onto overlapping directions. A recovered dictionary offers
candidate units for separating the unwanted cue from the task-relevant information. The
application needs a control for indiscriminate ablation and a comparison with alternative
representations or editing units.

\begin{result}[SHIFT: editing a classifier through recovered latents \citep{marks2025sparse}]
\label{res:shift}
\status{REPORTED-EMPIRICAL}
A linear head is trained on mean-pooled residual activations for a profession-classification task,
on a split in which gender predicts the label perfectly. A causal graph over autoencoder latents is
computed for the head's output ($67$ latents in Pythia-70M, $46$ in Gemma-2-2B). A human annotator
reads each latent and zero-ablates those judged irrelevant to profession ($55$ and $43$
respectively). On a balanced test set, worst-group accuracy rises from $24.4$ to $76.0$ in
Pythia-70M and from $18.2$ to $50.0$ in Gemma-2-2B. Ablating the same number of randomly chosen
latents gives $24.4$ and $18.0$, against the unedited baselines of $24.4$ and $18.2$. Retraining
the head on the edited representation gives $89.0$ and $92.9$, against an oracle trained on
balanced data at $91.9$ and $93.1$. A skyline, their term for a baseline that is allowed to use the
balanced set, reaches $62.9$ and $56.7$ when it zero-ablates latents and $41.5$ and $5.6$ when it
mean-ablates neurons. Concept bottleneck probing, a baseline that classifies from the affinities
between the activation and $20$ hand-chosen keyword directions, reaches $67.7$ and $86.7$.
\end{result}

\doesnot{The demonstration covers one dataset, one binary task, and one spurious attribute that is
perfectly predictive by construction, in Pythia-70M and Gemma-2-2B. The annotator knows the task
and supplies the concept of the spurious signal, so the demonstration removes the need for
disambiguating labels and not the need for supervision. The Gemma head is trained at the layer that
generalizes worst, which maximizes the available headroom, and the Gemma result without retraining
is much weaker than the Pythia result. Latents firing mainly on the beginning-of-sequence token
were excluded by hand from the annotation and the ablation. The neuron comparison uses a
thresholded-attribution neuron circuit rather than the best available neuron-based explanation, and
no supervised direction fitted for gender is compared against.}

The controls in \cref{res:shift} distinguish targeted editing from a generic effect of removing
latents. Random ablation preserves the poor baseline performance, while selected ablation
improves worst-group accuracy. The skyline comparison holds access to balanced data fixed and
finds higher scores for zero-ablation of latents than mean-ablation of neurons. This comparison
changes both the unit and the ablation convention, as qualified above.

The alternative classifier also prevents a uniform method ranking. Concept bottleneck probing
performs better than the edited representation on Gemma-2-2B but worse on Pythia-70M. It uses
hand-chosen keyword directions to build a new classifier without editing the representation.
Together, these controls support the particular SHIFT intervention while keeping the comparison
with other ways of using task supervision model-dependent.

Target removal additionally requires measuring damage to preserved behavior. The diffusion
erasure results in \cref{sec:app-steering} show that successful localization can accompany
substantial image artifacts \citep{cassano2026unlearning}. Forward-pass interventions also
leave trained weights intact, limiting any unlearning claim against access to those weights.
At the benchmark level, SAEBench's unlearning score remains outside the reliability audit
because it requires chat models \citep{chanin2026reliable}.

\subsection{Circuit Discovery and Mechanistic Prediction}
\label{sec:app-circuits}

\paragraph{Predicting Interventions.}
Circuit discovery aims to predict how an intervention changes the model's behavior from the
interactions among recovered features. SHIFT already uses a causal graph to select edits
(\cref{res:shift}); a further test asks whether a graph improves predictions beyond the
activation-level descriptions of its latents. This application uses the functional replacements
of \cref{sec:rec-replacement}, while their construction and recovery limitations remain those
examined in \cref{sec:recovery,sec:protocol,sec:failures}.

The May 2026 circuits update reports one such prediction comparison \citepalias{anthropic2026may}. Cross-layer
transcoder latents with similar top-activating examples and top unembeddings can have different
causal effects. A language model selects the latent that steers a prompt from groups of similar
candidates, using either activation descriptions or additional downstream connections. This
comparison tests whether the connections help distinguish interventions that their activation
descriptions leave ambiguous.

Across ten groups containing three to five candidates, the correct latent's normalized rank is
$0.457$ from top-activating examples, $0.446$ with top unembeddings added, and $0.381$ with
downstream latents added instead. Providing all three kinds of information gives $0.355$.
A perfect ranking scores $0$; the normalization implies an expected score of $0.5$ for uniform
random ranking. \status{REPORTED-EMPIRICAL, institutional report}
Downstream connections improve prediction in this preliminary comparison, but substantial
ranking error remains; the authors describe latent selection as far from solved.

\paragraph{Discovery Across Behaviors.}
Scaling circuit discovery introduces a separate evaluation problem concerning the fraction of
behaviors successfully explained. \citet{marks2025sparse} automatically construct graphs for
thousands of unsupervised clusters of Pythia-70M behavior. Two reported clusters that initially
appear to share one mechanism separate into distinct mechanisms under circuit analysis.
\status{REPORTED-EMPIRICAL} The authors leave evaluation of the full set of clusters and
circuits open, so these cases do not quantify the pipeline's yield.

The attribution-graph case studies provide a subjective estimate of yield.
The authors report satisfying insight on about a quarter of the prompts they tried, and state
that even successful cases capture a small fraction of the model's mechanisms
\citep{lindsey2025biology}. Their definition of a feature also remains under revision as the
tools develop. \status{REPORTED-EMPIRICAL, institutional report}
This selected-prompt judgment is distinct from a benchmark measurement of intervention prediction.

\paragraph{Limits of the Recovered Computation.}
The prediction target is restricted by the replacement model. Frozen attention patterns leave
attention selection outside the explanation \citep{ameisen2025circuit}, and error nodes retain
unexplained computation (\cref{sec:fail-darkmatter}). Attribution graphs are also computed for
individual prompts. The preliminary, low-confidence interference-weights note argues
in toy models that weights between superposed features combine circuit logic with overlaps
between feature directions \citep{olah2025interference}. The note conjectures that this mixing
obstructs recovery of global weight-level circuits from prompt-specific graphs.
\status{CONJECTURE, institutional report}
A standalone faithfulness audit of cross-layer transcoders or attribution graphs comparable to
the autoencoder audits was absent from the literature we identified
(\cref{sec:protocol,sec:open}).

\subsection{Robustness}
\label{sec:app-robustness}

Beyond its role in interpreting circuit weights, feature overlap is also a proposed target
for improving robustness in the interference studies of \cref{sec:rel-explanation}. Their strongest control increases width to a known class count, permitting
orthogonal directions in a synthetic task \citep{stevinson2026interference}. This changes the
available capacity rather than removing overlap from a fixed-width trained model.
Adversarial-training studies instead measure changes in different feature-count proxies, with
different observed trends \citep{gorton2025adversarial,bereska2025lossy}.

These interventions leave the practical question unresolved: whether overlap can be reduced at
fixed capacity while improving robustness and preserving task performance. Establishing that
connection requires a representation measurement alongside the attack outcome. In real tasks,
the additional requirement that a perturbation preserve the ground-truth label is also harder
to verify than in the synthetic control.

\subsection{Model Diffing and Fine-Tuning Analysis}
\label{sec:app-diffing}

\paragraph{Comparing Checkpoints.}
To characterize changes in a representation after training, model diffing compares recovered
features across checkpoints, often a base model and a finetuned descendant. The report by \citet{lindsey2024crosscoders} jointly trains a crosscoder on activations from the compared checkpoints and identifies model-exclusive
latents by a large decoder norm in one model and a small norm in the other. This comparison depends on the crosscoder's objective as well
as the models. A follow-up report shows that summing the two decoder norms separately encourages
exclusivity, whereas computing their norm jointly produces no exclusive latents in its experiment
\citep{mishrasharma2025diffing}.

The exclusive latents also activate about an order of magnitude more often than shared latents,
and many are polysemantic. A toy model with known shared and exclusive factors reproduces this
density gap at $1024$ latents for $700$ factors, while the gap disappears at $4096$ latents for
$450$ factors. In a small dictionary, a shared latent incurs the sparsity penalty twice and
reduces reconstruction error for both models. Its contribution to the objective is consequently
twice that of a comparable exclusive latent, favoring shared latents unless an exclusive latent
activates more frequently.

The report reduces this density difference by designating a small set of shared latents with a
lower sparsity penalty. It leaves the near-identical numbers of exclusive latents in the two
models unexplained. \status{REPORTED-EMPIRICAL, institutional report}
The objective changes and synthetic control show how a property attributed to model differences
can arise partly from dictionary training, linking diffing to the recovery failures in
\cref{sec:fail-splitting}.

\paragraph{Fine-Tuning and Unintended Behavior.}
Fine-tuning analysis also asks why a narrow training task can change behavior outside that task.
\citet{minegishi2026misalignment} derive a gradient-based explanation in which amplification of
a target feature also amplifies overlapping feature directions. Their prediction associates
misalignment-inducing data with features geometrically closer to harmful features than those in
data that does not induce misalignment.

The reported measurements follow this pattern across three Gemma-2 sizes, Llama-3.1 8B, and
gpt-oss 20B. In Gemma-2 2B, removing the half of a mixed secure/insecure training set closest to toxic
features reduces the reported count of misaligned outputs from $87$ to $57$. Evaluation uses
eight diagnostic prompts and a language-model judge, counting outputs with alignment below
$30$ after excluding those with coherence below $50$, but the evaluation description does not
specify the total number of sampled completions. This improvement exceeds removal of the same number of samples at
random and is comparable to filtering with a language-model judge.
\status{REPORTED-EMPIRICAL; preprint}
The geometric filter provides a tested use of feature overlap, although the remaining responses
and comparable judge-based control limit claims about its effectiveness as a remedy. Its
explanation depends on relations between directions (\cref{sec:geo-risk}), adding a use of
geometry alongside applications that depend on individual latent identities.

\subsection{Architectures and Domains Beyond Language}
\label{sec:app-architectures}

Feature recovery has also been applied to vision, biology, clinical prediction, and physical
modeling. These domains provide external concepts or measurements against which recovered
latents can be tested. \Cref{tab:crossarch} lists the architectures, decomposition targets, and
reported findings. The studies support transferring recovery procedures across domains, but use
different controls and do not establish a common application ranking.

\paragraph{External Concept and Causal References.}
Single-cell foundation models illustrate the difference between concept annotation and external
causal validation. Many recovered latents match biological database entries, and almost all are
absent from the leading singular vectors, which the author interprets as evidence of
superposition. Under genome-scale perturbation, however, $3$ of $48$ transcription factors show
target-specific latent responses \citep{kendiukhov2026singlecell}. Database agreement therefore
exceeds the demonstrated correspondence to regulatory effects in this study.

A physical model offers a different external reference. In a continuum-dynamics emulator,
recovered feature recruitment recurs intermittently and piecewise across shear-flow setups but
does not align with standard physical decompositions \citep{rosenfeld2026physics}. This
comparison tests whether the recovered descriptions transfer across simulations and correspond
to established physical variables. The clinical and diffusion interventions in
\cref{sec:app-steering} add tests of preserved behavior to these external comparisons.

\paragraph{Architecture-Specific Evidence.}
A time-series study reports little benefit from an expanded dictionary in the post-GELU
feed-forward activations of a forecasting transformer. Large parts of the dictionary remain
inactive, widening changes downstream performance by $0.214\%$ on average, and interventions
on dominant latents barely change the forecast. The author concludes that superposition is
unnecessary for competitive forecasting in this setting \citep{yildirim2026timeseries}.
The single-author preprint concerns one model and one site; its result motivates testing the
representation in each architecture rather than assuming that language-model observations transfer.

The available evidence is narrower for graph and recurrent architectures. The workshop study by
\citet{pertl2026gnn} uses class-conditional centroids and probe directions on synthetic graph
tasks with unambiguous concepts. Across three message-passing architectures, direction overlap
varies with width in a phase pattern and decreases with sharper pooling. These measurements
concern prescribed concepts, without dictionary recovery or intervention evidence for L3 or L5.
The recurrent-model study in \cref{sec:comp-temporal} sweeps width on a delay task
\citep{sharma2026temporal}; our search identified no study at scale in a selective state-space
language model.

The application evidence thus leaves both unmeasured properties of existing interventions and
architectures with little direct evaluation. \Cref{sec:open} develops tests for the missing
comparisons, including matched alternatives, magnitude sweeps, preserved behavior, and external
references for the representation itself.

\begin{table}[tbp]
\centering\small
\renewcommand{\arraystretch}{1.35}
\caption{Work on superposition and feature recovery outside language models that we identified.
The studies use different dictionaries, evaluation targets, and ground truth, so the rows are not
comparable with one another.}
\label{tab:crossarch}
\begin{tabularx}{\textwidth}{@{}>{\raggedright\arraybackslash}p{0.20\textwidth}
  >{\raggedright\arraybackslash}p{0.19\textwidth} X
  >{\raggedright\arraybackslash}p{0.13\textwidth}@{}}
\toprule
Domain and architecture & Object decomposed & Reported finding & Status \\
\midrule
Vision and image generation (InceptionV1, DINOv3, SDXL) &
activations, block-sparse featurizers &
concepts recovered as low-dimensional blocks rather than single directions, selected by a
description-length comparison against dictionaries of directions; generation steered within a
block \citep{fel2026blocksparse} &
preprint \\
Diffusion image models (SD 1.5, SDXL Turbo) &
activations, autoencoder latents &
latents detect and localize the target concept; multiplier interventions push activations off
distribution and produce artifacts in $43.1\%$ to $61\%$ of images
\citep{cassano2026unlearning} &
preprint \\
EEG transformers (SleepFM, REVE, LaBraM) &
embeddings, TopK autoencoder &
three steering regimes separated by a target-versus-off-target probe statistic; interventions that
collapse global performance; age and pathology inseparable \citep{lehnschioler2026eeg} &
preprint \\
Single-cell foundation models (Geneformer V2-316M, scGPT) &
residual stream, all layers &
$82{,}525$ and $24{,}527$ latents, $99.8\%$ invisible to a singular-value decomposition; $29\%$ to
$59\%$ annotate to biological databases; $3$ of $48$ transcription factors show target-specific
responses under CRISPRi \citep{kendiukhov2026singlecell} &
preprint \\
Continuum-dynamics emulator (Walrus) &
one layer, over $20{,}000$ latents &
feature recruitment recurs across shear-flow setups only piecewise, is intermittent, and does not
map onto standard physical decompositions \citep{rosenfeld2026physics} &
workshop \\
Time-series forecasting transformer (PatchTST) &
post-GELU feed-forward activations &
widening the dictionary changes performance by $0.214\%$ on average; interventions on dominant
latents barely move the forecast; the author concludes superposition is not necessary here
\citep{yildirim2026timeseries} &
preprint \\
Message-passing networks (GCN, GIN, GAT) &
class-conditional centroids and probe directions, synthetic graph tasks &
overlap between feature directions varies with width in a phase pattern; sharper pooling reduces
sharing; no dictionary and no intervention \citep{pertl2026gnn} &
workshop \\
Recurrent and linear-recurrence state-space models &
hidden state at width $m = 2$, sweep to $m = 100$, delay task &
temporal superposition, with the loss decomposition and geometry of \cref{sec:comp-temporal}
\citep{sharma2026temporal} &
refereed \\
\bottomrule
\end{tabularx}

\end{table}

\takeaway{sec:applications}{Supervised directions outperform dictionary latents on the
named-concept benchmarks reviewed here, while improved latent selection changes steering
performance. SHIFT shows that selected latent ablations can remove a spurious cue in a controlled
classifier. Steering can also damage preserved behavior after successful concept localization.
Biological and physical references expose further gaps between interpretable descriptions and
the effects attributed to them.}

\section{Open Questions}
\label{sec:open}

The applications in \cref{sec:applications} identify useful recovered features, but also expose
gaps in intervention controls, mechanistic prediction, and coverage across architectures.
The connection to theory depends on which features are claimed to be represented, where in the
network they are encoded, and which recovery or intervention test supports the claim. The theoretical results leave related questions about
representations across layers, nonlinear readouts, and the computations reached by training.

We assess the questions already posed by the field against subsequent work to distinguish
resolved problems from those that remain open (\cref{sec:open-ledger}). Some remaining questions
call for additional assumptions or bounds (\cref{sec:open-theory}), while others require measurements
of representations and behavior in trained networks (\cref{sec:open-empirical}). For each question, the discussion identifies the result that would
support the proposed explanation and an outcome that would limit or redirect it.

\begin{tcolorbox}[summarybox,title={Summary: Open Questions}]
\roadmap
\begin{itemize}
\item Large dictionaries are now feasible, but their width, objective, and latent labels do not
  yet provide a reliable inventory of a trained model's features (\cref{sec:open-ledger}).
\item Capacity across layers needs an explicit model of the available state, while nonlinear
  readout bounds need matched input and error criteria (\cref{sec:open-theory}).
\item Training guarantees remain separate from hand-built computation, and recovery beyond
  individual directions requires an identifiable target (\cref{sec:open-theory}).
\item A trained-network feature inventory would support direct recovery tests and a measurement
  of packing degree for the declared features (\cref{sec:open-empirical}).
\item Replacement-model audits and measurements of causally relevant reconstruction would test
  the connection between recovery scores and mechanistic explanation
  (\cref{sec:open-empirical}).
\end{itemize}
\end{tcolorbox}

\subsection{Status of the Field's Stated Open Problems}
\label{sec:open-ledger}

The questions posed in the toy-model report by \citet{elhage2022toy} and the broader
interpretability agenda of \citet{sharkey2025open} now divide between demonstrated
computational feasibility and unresolved feature recovery. \Cref{tab:ledger} records the
subsequent results and the scope of its selected questions.

The six negative findings concern how many dictionary latents are justified, whether the fitted directions recover
model features or mixtures, whether sparsity penalties and latent labels support that
interpretation, and whether optimizing the dictionary objective favors the model's own
directions. The evidence limits the
intended interpretations in the analyzed settings; a negative finding here does not assert
that every future recovery method fails. In particular, the results on objective optima
\citep{cui2025limits,tang2026unified} and width-dependent units \citep{leask2025canonical}
explain failures that the recovery and interpretation tests in \cref{sec:failures} expose.

The three reformulated questions need a more explicit target before an answer can be evaluated.
A recovery-qualified test of superposition requires a feature list, tolerance, and decoder
class, as the different readouts in \cref{ex:antipodal} illustrate. Counting features further
requires distinguishing representation, joint access, computation, and chosen dictionary width
(\cref{sec:def-capacity}). Asking for a faithful decomposition likewise requires specifying
its task and evidence, rather than assigning every method the same target
(\cref{tab:families,sec:def-ladder}).

The two positive answers are existence results. A toy model admits a closed-form optimum
\citep{scherlis2022capacity}, and dictionary learning has been run at large scale on language
models \citep{gao2025scaling,lieberum2024gemmascope}. These achievements establish feasibility
without settling recovery quality. The partly answered and open questions concern how capacity,
training, and scaling determine the representation, and whether measurements in trained
networks support the proposed explanations. We develop these questions after the ledger.

\begingroup\small
\renewcommand{\arraystretch}{1.25}
\begin{longtable}{@{}>{\raggedright\arraybackslash}p{0.215\textwidth}
  >{\raggedright\arraybackslash}p{0.072\textwidth}
  >{\raggedright\arraybackslash}p{0.112\textwidth}
  >{\raggedright\arraybackslash}p{\dimexpr0.601\textwidth-6\tabcolsep\relax}@{}}
\caption{What became of the open questions stated by \citet{elhage2022toy} and, for the
superposition-relevant part of a much longer list, by \citet{sharkey2025open}. Of fifty questions assessed, seventeen are displayed: all six with negative findings, all three
reformulated questions, both answered questions, three partly answered questions, and three open
questions. The thirty-three omitted questions comprise twenty-eight partly answered and five open
questions. Verdicts summarize the evidence in the analyzed settings.}\label{tab:ledger}\\
\toprule
Question & Posed in & Verdict & Evidence and Remaining Scope \\
\midrule
\endfirsthead
\multicolumn{4}{@{}l}{Table \thetable\ (continued)} \\
\toprule
Question & Posed in & Verdict & Evidence and Remaining Scope \\
\midrule
\endhead
\midrule
\multicolumn{4}{r@{}}{Continued on next page} \\
\endfoot
\bottomrule
\endlastfoot
How many latents should a dictionary have? &
2022 report & unique width unestablished &
The studies leave dictionary width without a uniquely supported choice. Latents fitted at one
width decompose into latents fitted at another
\citep{leask2025canonical} (\cref{sec:fail-canonical}). \\
Does dictionary learning on real activations return the model's own directions? &
2022 report & answered negatively &
Recovery varies with the generator, matching rule, and threshold rather than consistently
returning the stipulated directions: $3$ of $3{,}200$ directions at explained variance $0.67$ under one generator and
matching rule, against almost complete recovery under another \citep{korznikov2026sanity}
(\cref{sec:prot-synthetic}); chess coverage $0.48$ against $0.98$ for probes on the same
activations \citep{karvonen2024boardgame} (\cref{sec:prot-organisms}). \\
Are the recovered atoms the model's features or mixtures of them? &
2025 agenda & mixtures can occur &
The analyzed optima can favor mixtures, with reconstruction succeeding while recovery fails
outside the stated extreme-sparsity conditions \citep{cui2025limits, tang2026unified} (\cref{sec:acc-ident}). \\
Is the sparsity objective the right training signal for recovery? &
2025 agenda & answered negatively &
The objective also favors artifacts. Absorption follows from the penalty (\cref{res:absorption}), and
splitting appears with no multi-scale structure present \citep{makelov2025principled}
(\cref{sec:fail-splitting}). \\
Has a latent's label been shown to be a property of the model? &
2025 agenda & answered negatively &
Confident descriptions attach to $82\%$ of random directions against $80\%$ of real coordinates
\citep{bolukbasi2021illusion}, and a frozen random decoder matches trained dictionaries on
automated interpretability \citep{korznikov2026sanity} (\cref{sec:fail-illusions}). \\
\midrule
Is there a test certifying from activations alone that a layer holds more features than
dimensions? &
2022 report & refor\-mulated &
A recovery-qualified verdict depends on the feature list, tolerance, and decoder class
(\cref{def:superposition}, \cref{ex:antipodal}); the packing degree at a specified layer tests whether the stated feature list contains more
recoverable features than that layer has dimensions (\cref{def:packing}). \\
What is the right decomposition of a network, and what makes it faithful? &
2025 agenda & refor\-mulated &
The reported dictionaries lack demonstrated completeness or atomicity \citep{leask2025canonical}. A dictionary reconstructs activations, whereas a replacement model approximates computation
(\cref{tab:families}); the six evidential commitments distinguish what each type of test can
establish (\cref{sec:def-ladder}). \\
How many features does a model have? &
2025 agenda & refor\-mulated &
The relevant counts distinguish representable features, jointly readable features \citep{garg2026howmany}, computable features
\citep{adler2024complexity}, and the dictionary width $\hat d$, which is a training choice
(\cref{sec:def-capacity}, \cref{tab:capacity}). \\
\midrule
Is there a toy model whose optimum and phase boundaries can be written down? &
2022 report & answered &
The quadratic toy model has a closed-form optimum, with a fourth-moment threshold for the
fractional regime \citep{scherlis2022capacity} (\cref{res:scherlis}); the boundaries of the
original ReLU model are still measured (\cref{sec:acc-when}). \\
Can dictionary learning run on a frontier model's activations? &
2022 report & answered &
$16$ million latents on GPT-4 \citep{gao2025scaling}, $34$ million on Claude 3 Sonnet
\citep{templeton2024scaling}, and every layer of Gemma 2 2B and 9B at over $20\%$ of GPT-3's training compute
\citep{lieberum2024gemmascope} (\cref{sec:rec-dictionaries}, \cref{tab:cost}). \\
What pressure drives a post-hoc dictionary toward the model's own features? &
2022 report & objective insufficient in analyzed settings &
The analyzed standard objectives can reach their optima without recovering the directions
\citep{cui2025limits, tang2026unified}, and a decoder frozen at random initialization matches
trained dictionaries on three standard scores \citep{korznikov2026sanity} (\cref{sec:acc-ident}). \\
\midrule
How many features can a width-$m$ layer carry? &
2022 report & partly answered &
$m = \widetilde\Theta_\varepsilon(k^2)$ for one linear map against $O(k \log(d/k))$ under suitable restricted-isometry conditions for sparse
optimization, so the count is indexed by decoder class \citep{garg2026howmany} (\cref{res:garg},
\cref{tab:access}). These rates concern a specified activation site and recovery criterion. \\
Which functions can a network compute on superposed features, and at what width? &
2022 report & partly answered &
A universal architecture realizing the specified Boolean families needs
$\Omega(d'\log d')$ parameter bits; constant depth and bounded precision give the
$O(m^2/\log m)$ output-count limit on family expressivity
\citep{adler2024complexity} (\cref{res:separation}). The error-tolerant result retains its
selected input set and robust function subset; the upper constructions are assembled by hand
(\cref{sec:comp-gap}). \\
Is superposition caused by capacity pressure or by incidental effects? &
2025 agenda & partly answered &
Both, in their own settings: sharing is optimal above a fourth-moment threshold
\citep{scherlis2022capacity}, and polysemantic units also form at $m \ge d$
\citep{lecomte2024incidental}. Their proportions in a trained model remain unmeasured
(\cref{sec:acc-when}, \cref{sec:acc-incidental}). \\
\midrule
Does the superposed fraction of features vanish as width grows? &
2022 report & open &
A general condition on the importance and sparsity curves remains to be derived. Open-weight models fit the $1/m$ law
of the strong-superposition regime \citep{liu2025scaling}, while the loss increase from ablating one
context neuron falls from $8\%$ at 70M parameters to $0.2\%$ at 6.9B \citep{gurnee2023haystack}
(\cref{sec:acc-scaling}). \\
Can superposition be eliminated rather than reduced? &
2022 report & open &
In the reported synthetic control at $m = d$, attacks fail in all $1000$ attempts \citep{stevinson2026interference}, but $m = d$
requires knowing how many features the model represents, and no method surveyed here reduces
superposition at fixed capability (\cref{sec:app-robustness}). \\
Is superposition a true description of trained models, or a pragmatically useful one? &
2025 agenda & open &
A packing degree against a stated feature list, tolerance, and decoder class is unreported
for the trained networks surveyed here (\cref{sec:open-empirical}).
\end{longtable}
\endgroup

\subsection{Theoretical Questions}
\label{sec:open-theory}

\paragraph{Which State Determines Capacity Across Layers?}
Single-site bounds concern recovery from the activation vector at one location in the network.
For a multi-layer analysis, the observation model must distinguish joint access to activations
from several layers from access only to the state passed through successive transformations.
This distinction also matters for the crosscoder report, which finds latents with substantial
decoder norm across many layers but also considers sequential production and amplification
as an explanation \citep{lindsey2024crosscoders}. These
observations leave shared representation and repeated manifestations of a feature unresolved
(\cref{sec:rec-crosslayer}).

The missing capacity bound concerns this available state and its transformations, rather than
the sum of layer widths. The construction would need
to account for dependencies between layers before applying a bound such as
\cref{eq:budget,res:scherlis,res:garg}. If the observed cross-layer latents instead track
production followed by amplification, single-site analyses may describe the relevant
bottlenecks. Simply summing layer widths would not distinguish these alternatives.

\paragraph{What Width Suffices for a Linear Map Followed by a Pointwise Nonlinearity?}
The access results in \cref{tab:access} leave sharp rates for
$\Psi_{\mathrm{lin}+\sigma}$ unresolved. Uniform linear recovery has width
$m=\widetilde\Theta_\varepsilon(k^2)$ in the setting of \cref{res:garg}, whereas sparse
optimization admits $m=O(k\log(d/k))$ under appropriate dictionary assumptions. These results
use specified recovery criteria and do not order all decoder families into nested sets.
A linear map followed by a pointwise nonlinearity is relevant to individual network operations,
but a multi-layer downstream computation need not belong to that class
\citep{rajamanoharan2024gated}.

To determine the width required by this decoder class, upper and lower bounds must use the same
admissible inputs, nonlinearity, error criterion, and encoder constraints. A quadratic dependence on
$k$ would retain the linear rate in that setting; a compressed-sensing-type rate would show a
larger improvement under matched assumptions. Either outcome would concern that recovery
problem rather than every distinction between linear and nonlinear superposition. A second
comparison would replace uniform sparse inputs with a stated correlated feature law, since
correlations already change optimal geometry \citep{prieto2026correlations}.

\paragraph{Does Training Produce Computation in Superposition?}
The hand-built networks of \cref{sec:computation} establish computational possibilities, while
a guarantee that gradient training reaches those constructions remains absent from the
surveyed results. The trained examples address narrower settings. One reverse-engineered
representation appears under a fourth-power loss \citep{ferreira2026l4}; another study finds a
routing motif in two-layer MLPs trained on sparse Boolean formulas
\citep{adler2025combinatorial}.

A convergence guarantee for $d'$ computed features in $m\ll d'$ neurons must connect the
architecture and objective to the solution reached from a specified initialization. The guarantee would need to establish that the trained network computes the target features,
rather than merely allowing an externally chosen decoder to read them out, since a random MLP already permits readout of pairwise ANDs in the setting of
\cref{sec:comp-gap}. Empirically, a loss-exponent-by-task sweep with per-feature error
measurements could separate loss-dependent behavior from task-dependent behavior
(\cref{sec:comp-dispute}). Computation under squared error in some tasks would limit an
explanation based on the loss exponent alone.

\paragraph{Which Objects Are Identifiable Beyond Individual Directions?}
Recovery guarantees in \cref{sec:acc-ident} require explicit generative assumptions, while
trained-model studies also describe circular subspaces \citep{engels2025notall}, curved
manifolds \citep{gurnee2026manifolds}, and hierarchies affected by absorption
(\cref{sec:fail-splitting}). These observations motivate identifying the appropriate recovery
target before asking whether repeated dictionary fits recover the same individual directions.

A result for a specified class of subspaces, manifolds, or hierarchical features would state
the symmetries under which recovery is equivalent and the distributional assumptions that
permit identification. Subspace stability in some experiments motivates this target without
making subspaces preferable in every setting (\cref{sec:fail-canonical}). A non-identifiability
construction with the same observed distribution and distinct candidate objects would instead
show that additional assumptions are needed.

The observation model is another part of this question. \citet{liu2026extracting} recover
nondegenerate nonlinear ridge directions through chosen queries, while a dictionary fitted to
activations usually receives passive samples. The source also discusses computational hardness
in passive settings. A useful extension would identify restricted response functions and
sampling distributions that admit recovery with a stated sample size, or derive a lower bound
showing why query access remains necessary under the proposed assumptions.

\paragraph{What Capacity Can Be Defined for Causally Relevant Features?}
The representation and computation results in
\cref{res:budget,res:separation} do not supply a count defined by a causal intervention test.
The global-workspace report motivates such a distinction by finding that a small
median share of a concept vector's variance carries much of the measured output effect
\citep{gurnee2026workspace}. The measurement concerns a particular intervention and output
criterion; it does not directly turn a causal effect into a bound on dimension.

A capacity question would first define distinct candidate features, allowed interventions,
effect tolerance, and behaviors intended to remain unchanged. A subsequent bound could ask
how many features satisfy that criterion at width $m$, just as
$R^2_i\ge 1-\varepsilon$ fixes a recovery criterion in \cref{def:superposition}.
If arbitrary rescalings or repeated directions satisfy the causal criterion, a count requires
further normalization or equivalence conditions. This alternative would expose a deficient
counting criterion rather than a failure of the existing representation bounds.

\subsection{Empirical Questions}
\label{sec:open-empirical}

\paragraph{Can a Trained Network Have a Known Feature Inventory?}
Synthetic generators supply known features under a stipulated encoder and distribution
\citep{chanin2026synthsaebench,korznikov2026sanity}, whereas board-game models supply a trained
network with a partial list of relevant properties \citep{karvonen2024boardgame}. The remaining
gap is a trained network with a nontrivial, independently justified feature inventory and tests
that distinguish those features from surface-form cues.

A benchmark approaching this goal would establish which listed variables the model represents,
then score dictionary recovery against both their directions and activation patterns.
Surface-form masking, as in \cref{sec:prot-organisms}, would test leakage; model randomization
would test dependence on learned structure. A frozen-random dictionary would separately test
whether a recovery score depends on learned decoder directions. If random controls scored as well as the trained methods, that result would call the recovery
procedure or its measurement into question without by itself disproving the proposed feature
inventory.

\paragraph{Can Packing Degree Be Measured in a Trained Layer?}
A feature inventory would permit the recovery-qualified counting test that existing
intermediate-layer studies leave open. \citet{gurnee2023haystack} explicitly note that
establishing superposition by counting requires exhibiting more features than neurons.
The language-model-head measurements of \citet{liu2025scaling} concern row norms and
interference distributions, rather than that direct count at an intermediate layer.

A measurement would fix the feature list, site, tolerance, decoder class, and evaluation
distribution, then estimate or bound $\kappa_\varepsilon$ (\cref{def:packing}). Exhibiting
$\kappa_\varepsilon>1$ would support superposition for those features under that criterion.
Establishing $\kappa_\varepsilon\le1$ would require an upper bound over the permitted decoders,
rather than a failed recovery run, and would leave omitted features outside the conclusion.
The count also separates overcomplete recovery from polysemanticity alone
(\cref{sec:acc-incidental}), but does not establish which training pressure caused the
representation.

\paragraph{How Faithful Are Replacement Models?}
The sparse-autoencoder audits test dependencies on random directions, evaluation seeds, and
latent activation or intervention behavior
\citep{korznikov2026sanity,chanin2026reliable,bal2026audit}. Comparable independent tests of
cross-layer transcoders and attribution graphs remain missing from the literature we identified.
Their replacement and prompt-insight measurements come from institutional reports
\citep{ameisen2025circuit,lindsey2025biology}, whose distinct evaluation targets are described
in \cref{sec:prot-darkmatter,sec:app-circuits}.

The unresolved issue is whether learned transcoder directions improve intervention predictions
beyond a random replacement with the same reconstruction and evaluation conventions.
The comparison would keep the attention and error-node conventions fixed
(\cref{sec:rec-replacement,sec:fail-darkmatter}). Better held-out intervention predictions
would support the graph's tested explanatory use. Parity with random replacements, or
predictions that depend chiefly on error nodes, would limit what the replacement score
establishes about the recovered computation.

\paragraph{Does Reconstruction Preserve Causally Relevant Components?}
A reconstruction objective weights activation differences without directly scoring their
behavioral effects. The evidence in \cref{sec:prot-variance} motivates a direct comparison,
including replication of the global-workspace analysis on open models
\citep{gurnee2026workspace}. The missing measurement is how much of the causally effective
component a dictionary preserves when its reconstruction replaces the activation.

The unresolved relation is whether dictionaries that recover more activation variance also
preserve more of the component responsible for the tested behavior. Existing intervention
results compare components of concept vectors, whereas this question concerns replacing an
activation with its dictionary reconstruction. Measurements across dictionaries and sites would
establish where reconstruction accuracy tracks that behavioral criterion.

\paragraph{How Do Recovery and Intervention Transfer Across Architectures?}
The architecture coverage in \cref{tab:crossarch} remains uneven. A message-passing-network
workshop study measures overlap without training a dictionary or intervening
\citep{pertl2026gnn}, while the recurrent study develops its geometric example at $m=2$ and
sweeps width to $m=100$ on a delay task \citep{sharma2026temporal}. Our search identified no
corresponding study at scale in a selective state-space language model.

A new study would specify a task-relevant feature list and test recovery, interventions, and
preserved behavior before interpreting dictionary width as evidence for superposition.
A packing-degree measurement would add a direct recovery-qualified count. The time-series
preprint in \cref{sec:app-architectures} motivates allowing a different outcome: wider
dictionaries and dominant-latent interventions may yield little task benefit
\citep{yildirim2026timeseries}. Such a finding would restrict the usefulness of the tested
recovery and intervention procedure in that model, while a conclusion about absent
superposition would require the corresponding counting evidence.

\takeaway{sec:open}{The remaining gaps concern feature recovery across layers, the training processes that produce
useful computation, and the conditions under which activation reconstruction preserves causal effects.
A known feature inventory in a trained model would connect several of these questions by
supporting recovery tests and a direct count of readable features. Replacement-model
interventions would further test whether recovered computations predict behavior.}

\section{Concluding Remarks}
\label{sec:conclusion}

In this survey, we connected the theoretical conditions for representing and recovering
superposed features with the methods and evidence used to study trained networks. Feature
geometry constrains interference, while feature statistics, error tolerance, and decoder
choice determine what can be recovered. The toy-model results show how training objectives
favor different representations, and the computational results distinguish storing features
from computing specified functions of them. These conclusions depend on their stated input
laws and recovery criteria; the number of represented features alone does not determine which values a linear decoder can
recover or which functions the network computes.

In trained networks, however, the feature values and directions must also be estimated from
activations or weights. Our comparison of recovery methods therefore distinguishes the evidence
supplied by reconstruction, feature matching, descriptions, and interventions. The documented
failures explain why these tests answer different questions: sparse objectives can alter the
recovered units, random directions can receive convincing descriptions, and activation components
left out of a reconstruction can affect the model's behavior. Applications nevertheless demonstrate useful detection and intervention in particular
settings, with their conclusions determined by the task, alternatives, and measured effects
on behavior intended to remain unchanged.

To turn these tested uses into an explanation of the model's learned representation and
computation, recovery methods need explicit feature targets, recovery criteria, and evidence
that the recovered computation predicts the effects of interventions. On the theoretical side,
bounds must state which layer activations a decoder can observe, while training guarantees must
connect the optimization procedure to computation in superposition. Progress on these theoretical and empirical
questions would clarify when superposition explains a trained network and when a feature
recovery method provides a useful but incomplete description of its behavior.

\clearpage
\bibliographystyle{plainnat}
\bibliography{main}

\end{document}